%% file: root.tex
\documentclass[lettersize,journal]{IEEEtran}
\usepackage{amsmath,amsfonts}
\DeclareMathOperator*{\argmax}{arg\,max}
\DeclareMathOperator*{\argmin}{arg\,min}
\DeclareMathOperator*{\argminC}{\arg\min}   
\DeclareMathOperator*{\argmaxC}{\arg\max}   
\usepackage[utf8]{inputenc}
\usepackage{multirow}
\usepackage{csquotes}
\usepackage{array}
\usepackage{algorithm2e}
\usepackage{enumitem}
\usepackage{adjustbox}
\usepackage{booktabs}
\usepackage{rotating}

\usepackage[font=small]{caption}
\usepackage{subcaption}
\usepackage{textcomp}
\usepackage{stfloats}
\usepackage{url}
\usepackage{cite}
\usepackage{verbatim}
\usepackage{graphicx}
\usepackage{collectbox}
\usepackage{xcolor}
\RestyleAlgo{ruled}

\usepackage[colorlinks=true]{hyperref}

\begin{document}

\title{Learning Multimodal Latent Dynamics for Human-Robot Interaction}

\author{Vignesh Prasad$^1$, Lea Heitlinger$^2$, Dorothea Koert$^{3,4}$\\Ruth Stock-Homburg$^2$, Jan Peters$^{4,5,6,7}$, Georgia Chalvatzaki$^{1,7}$\vspace{-2em}
\thanks{
\hspace{-1em}$^1$ Interactive Robot Perception and Learning group (PEARL), Department of Computer Science, TU Darmstadt, Germany.\\
$^2$ Chair for Marketing and Human Resource Management, Department of Law and Economics, TU Darmstadt, Germany.\\
$^3$ Interactive AI Algorithms \& Cognitive Models for Human-AI Interaction (IKIDA), Department of Computer Science, TU Darmstadt, Germany.\\
$^4$ Centre for Cognitive Science, TU Darmstadt, Germany.\\
$^5$ Institute for Intelligent Autonomous Systems (IAS), Department of Computer Science, TU Darmstadt, Germany.\\
$^6$ Systems AI for Robot Learning, German Research Center for AI (DFKI).\\
$^7$ Hessian Center for
Artificial Intelligence (Hessian.AI), Darmstadt, Germany\\
Contact: \tt{\href{mailto:vignesh.prasad@tu-darmstadt.de}{vignesh.prasad@tu-darmstadt.de}}\\
Website: \url{https://sites.google.com/view/mild-hri}}
}

\markboth{Transactions on Robotics,~Vol.~X, No.~X, X~20XX}%
{Prasad \MakeLowercase{\textit{et al.}}: Learning Multimodal Latent Dynamics for Human-Robot Interaction}

\IEEEpubid{0000--0000/00\$00.00~\copyright~2025 IEEE}


\maketitle
\begin{abstract}
This article presents a method for learning well-coordinated Human-Robot Interaction (HRI) from Human-Human Interactions (HHI). We devise a hybrid approach using Hidden Markov Models (HMMs) as the latent space priors for a Variational Autoencoder to model a joint distribution over the interacting agents. We leverage the interaction dynamics learned from HHI to learn HRI and incorporate the conditional generation of robot motions from human observations into the training, thereby predicting more accurate robot trajectories. The generated robot motions are further adapted with Inverse Kinematics to ensure the desired physical proximity with a human, combining the ease of joint space learning and accurate task space reachability. For contact-rich interactions, we modulate the robot’s stiffness using HMM segmentation for a compliant interaction. We verify the effectiveness of our approach deployed on a Humanoid robot via a user study. Our method generalizes well to various humans despite being trained on data from just two humans. We find that users perceive our method as more human-like, timely, and accurate and rank our method with a higher degree of preference over other baselines. We additionally show the ability of our approach to generate successful interactions in a more complex scenario of Bimanual Robot-to-Human Handovers.

\end{abstract}

\begin{IEEEkeywords}
Physical Human-Robot Interaction, Learning from Demonstration, Humanoid Robots, Probability and Statistical Methods
\end{IEEEkeywords}

\section{Introduction}

\begin{figure}[h!]
    \centering
    \includegraphics[width=0.9\linewidth]{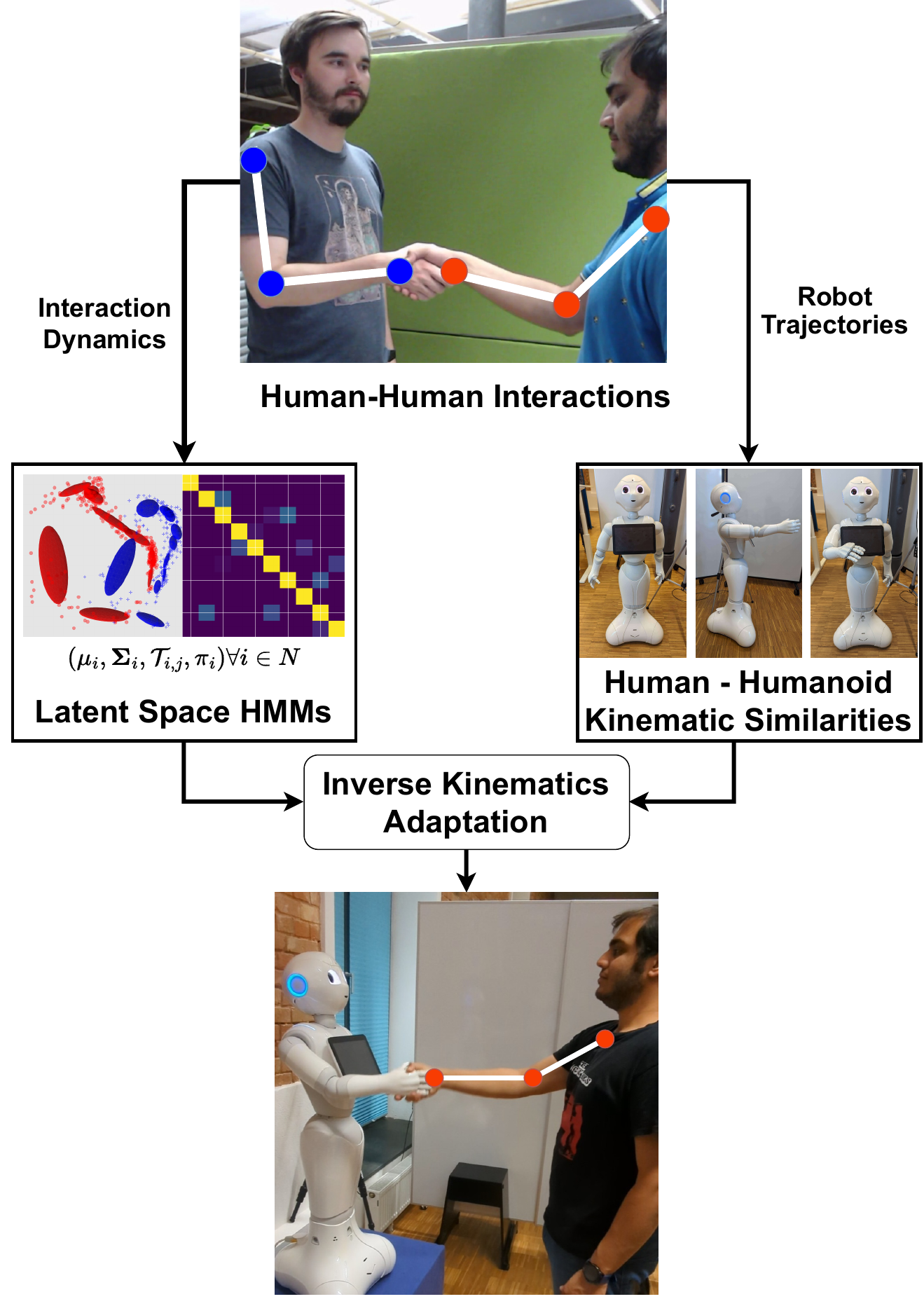}
    \caption{We explore learning coordinated HRI behaviors using Hidden Markov Models (HMMs) to learn the interaction dynamics over a representation space spanned by a Variational Autoencoder (VAE). The VAEs are trained with the HMMs as a prior to better incorporate the understanding of the dynamics. Moreover, we learn directly from Human-Human interactions by leveraging the kinematic similarities of Humans and Humanoid robots. During testing, the predicted trajectories are adapted with Inverse Kinematics to mitigate errors in the spatial accuracy arising either from prediction errors or from shortcomings in transferring the human's actions to the robot.}
    \label{fig:teaser}
    \vspace{-1em}
\end{figure}


\IEEEPARstart{E}{nsuring} a synchronized and accurate reaction is an important aspect in Human-Robot Interaction (HRI)~\cite{ajoudani2018progress}. To do so, a human and a robot need to spatiotemporally coordinate their movements to reach a common target or perform a common task, that can be denoted as a \textit{joint action}~\cite{sebanz2006joint}. For realizing such coordinated jointly performed actions, spatial and temporal adaptation of motions in the shared physical space to accurately perform the action are key factors~\cite{sebanz2009prediction}. Such interpersonal coordination can enable a connection to one's interaction partner~\cite{marsh2009social}. Therefore, well-executed and well-coordinated interactive behaviors can improve the perception of a social robot. 

\IEEEpubidadjcol
The paradigm of Learning from Demonstrations (LfD) is promising for HRI by learning joint distributions over human and robot trajectories in a modular, multimodal manner~\cite{evrard2009teaching,calinon2009learning,pignat2017learning,ewerton2015learning,koert2018online}. 
However, LfD approaches scale poorly with higher dimensions, which can be circumvented by incorporating Deep Learning for learning latent-space dynamics. 
Such Deep State-Space Models have shown good performance in capturing temporal dependencies of latent space trajectories, either using some kind of a forward propagation model~\cite{becker2019recurrent,dai2016recurrent,chung2015recurrent,butepage2020imitating}, learning the parameters for a dynamics model with LfD~\cite{karl2016deep,chen2016dynamic} or fitting parameterized LfD models to the underlying latent trajectories~\cite{dermy2018prediction,nagano2019hvgh}. \IEEEpubidadjcol In this paper, we further explore this direction of latent space dynamics in the context of Human-Robot Interaction.

Furthermore, when learning trajectories in the joint configuration space of a robot, minor deviations in the joint space can cause a perceivable deviation in the robot's task space. While learning task space trajectories can mitigate problems from errors in the joint space, to do so, the demonstrated trajectories need to fall within the reachability of the robot. When this reachability assumption is violated, the task space proximity in HRI scenarios may not be effectively learned. Moreover, learning purely task space trajectories doesn't necessarily ensure human-like joint configurations, which is especially relevant for Humanoid robots. To this end, the combination of joint space learning and task space reachability can be achieved using Inverse Kinematics to supplement the learned behaviors \cite{gomez2020adaptation,prasad2021learning,capdepuy2015improving}. 

In this paper, as shown in Figure~\ref{fig:teaser}, we are interested in learning and adapting Deep LfD policies for spatiotemporally coherent interactive behaviors. We learn the interaction dynamics from HHI demonstrations using latent space Hidden Markov Models (HMMs). We show the efficacy of the learnt dynamics in real-world HRI scenarios. We do so in a manner that ensures spatially accurate physical proximity to the human partner and additionally enable compliant robot motions in contact-rich interactions like a handshake, thereby improving the perceived quality of the HRI behaviors.

\subsection{Related Work}
\label{sec:related}

\subsubsection{Learning HRI from Demonstrations}
\label{sec:lfd-hri}

Early approaches for learning modular HRI policies modeled the interaction as a joint distribution with a Gaussian Mixture Model (GMM) learned over demonstrated trajectories of a human and a robot in a collaborative task~\cite{calinon2009learning}. The correlations between the human and the robot degrees of freedom (DoFs) can then be leveraged to generate the robot's trajectory given observations of the human for learning both proactive and reactive controllers~\cite{rozo2016learning,calinon2016tutorial,pignat2017learning}. Segmenting HRI demonstrations has also been shown using Graphical Models with Markov chain Monte Carlo~\cite{ShuIJCAI16,ShuICRA17}.

Along the lines of leveraging Gaussian approximations for LfD, Movement Primitives~\cite{paraschos2013probabilistic}, which learn a distribution over underlying linear regression weight vectors, were extended for HRI by similarly learning a joint distribution over the weights of interacting agents~\cite{amor2014interaction,maeda2014learning,campbell2017bayesian}. 
Movement Primitive approaches can further be 
combined with GMMs for learning multiple task sequences seamlessly~\cite{ewerton2015learning,maeda2017active,koert2018online,oikonomou2022reproduction,lioutikov2017learning}. 
One drawback of the aforementioned primitive-based approaches is that they do not perform well on out-of-distribution data~\cite{zhou2019learning}. Additionally, for tasks that are loosely coupled in time, Movement Primitives can underperform~\cite{paraschos2018using}. These drawbacks therefore make them an unviable option for being able to generalize spatiotemporally in interactive behaviors. For a more extensive overview of Movement Primitive approaches for robot learning, we refer the reader to~\cite{tavassoli2023learning}. 



Vogt et al.~\cite{vogt2017system} are similar in theory to our approach where they learn the interaction dynamics of Human-Human Interactions using an HMM over a low dimensional representation space of the human skeletons. In our approach, we show further performance improvements by incorporating human-conditioned reactive motion generation into the training pipeline.
Additionally, Vogt et al.~\cite{vogt2017system} train Interaction Meshes~\cite{ho2010spatial} for transferring the learned trajectories to a robot whose prediction is optimized during runtime to adapt to the human user's movements. As we work with Humanoid Robots, we follow a simpler approach by leveraging the structural similarity between a human and a humanoid. This allows us to map the joint motions directly~\cite{fritsche2015first} and adapt the motions to the interaction partner using Inverse Kinematics~\cite{prasad2021learning}. Through our user study, we find that our approach, while being simplistic, provides acceptable interactions with various human partners.

To make the learned distribution more robust, LfD methods are often learned via kinesthetic teaching which can be tedious for HRI tasks where one would need a variety of human partners. The need for extensive training data can potentially be circumvented by learning how humans adapt to a robot's trajectory \cite{campbell2019learning}. A more general way, especially in the context of Humanoid robots, is by learning from human-human demonstrations by leveraging the kinematic similarities between humans and robots \cite{fritsche2015first}. While such approaches of motion retargeting leave some room for error due to minor differences between human and robot geometries, adapting the trajectories in the robot's task space via Inverse Kinematics can improve the accuracy of such interactive behaviors \cite{prasad2021learning,vinayavekhin2017human}. 

\subsubsection{Integrating LfD with Deep Learning}
\label{ssec:deep-lfd}
Techniques at the intersection of LfD coupled with the use of Neural Network representations for higher dimensional data have grown in popularity for learning latent trajectory dynamics from demonstrations. Typically, an autoencoding approach, like VAEs, is used to encode latent trajectories over which a latent dynamics model is trained. In their simplest form, the latent dynamics can be modeled either with linear Gaussian models~\cite{karl2016deep} or Kalman filters~\cite{becker2019recurrent}. Other approaches learn stable dynamical systems, like Dynamic Movement Primitives~\cite{schaal2006dynamic} over VAE latent spaces~\cite{bitzer2009latent,chen2015efficient,chen2016dynamic,chaveroche2018variational,colome2014dimensionality}. Instead of learning a feedforward dynamics model, Dermy et al.~\cite{dermy2018prediction} model the entire trajectory's dynamics at once using Probabilistic Movement Primitives~\cite{paraschos2013probabilistic} achieving better results than~\cite{chaveroche2018variational}.

Nagano et al.~\cite{nagano2019hvgh} demonstrated the use of Hidden Semi-Markov Models (HSMMs) as latent priors in a VAE for temporal action segmentation of motions involving a single human. They model each latent dimension independently, which is not favorable when learning interaction dynamics. To extend such an approach for learning HRI, one needs to consider the interdependence between dimensions. In our previous work, \enquote{MILD}~\cite{prasad2022mild}, we extend this idea of using HSMMs as VAE priors for interactive tasks by exploiting the full rank of HSMM covariance matrices thereby capturing the dependencies between both agents by learning a joint distribution over the latent trajectories of interacting partners.

\begin{figure*}[h!]
    \centering
    \includegraphics[width=0.7\textwidth]{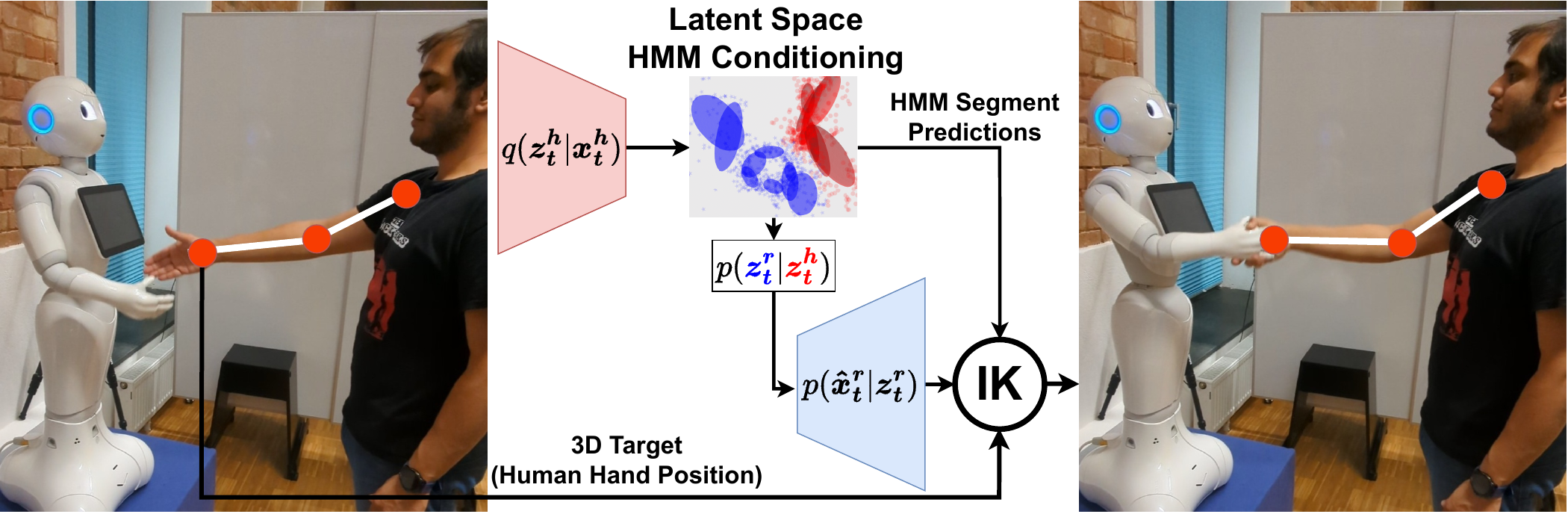}
    \caption{Overview of our approach. We train VAEs to reconstruct the observations of the interactions agents $(\boldsymbol{x}^h_{1:t},\boldsymbol{x}^r_{1:t})$ with an HMM prior to learn a joint distribution over the latent space trajectories $p(\boldsymbol{z}^h_{1:t},\boldsymbol{z}^r_{1:t})$ of the interacting agents. During test time, the observed agent's latent trajectory conditions the HMM to infer the robot agent's latent trajectory $p(\boldsymbol{z}^r_t|\boldsymbol{z}^h_{1:t})$ which is decoded to generate the robot agent's joint trajectory $\boldsymbol{\hat{x}}^r_t$. To ensure the proximity of a robot's hand to the human's hand during test time, we additionally adapt the predicted trajectory using Inverse Kinematics to fulfill the contact-based nature of the interaction.}
    \label{fig:vaeHMM-test}
    \vspace{-1em}
\end{figure*}

\subsubsection{Interaction Modeling with Recurrent Neural Networks}
\label{ssec:int-rnn}
When large datasets are available, Recurrent latent space models are powerful tools in approximating latent dynamics with some form of a forward propagating distribution~\cite{chung2015recurrent,han2021disentangled,krishnan2017structured,liu2020powering,fabius2015variational}. Given their power of modeling temporal sequences, they yield themselves naturally for learning interactive/collaborative tasks in HRI~\cite{oguz2019ontology,zhao2018collaborative,semeraro2023human}. To simplify the various scenarios of interaction tasks Oguz et al.~\cite{oguz2019ontology} develop an ontology to categorize interaction scenarios and train an LSTM network for each case in a simple Imitation Learning paradigm. Similarly, Zhao et al.~\cite{zhao2018collaborative} explore using such LSTM-based policies for learning HRI in a simple collaboration scenario. Rather than simply regressing robot actions to human inputs, B\"utepage et al.~\cite{butepage2020imitating} explore the idea of learning shared latent dynamics in HRI in a systematic way. B\"utepage et al.~\cite{butepage2020imitating} initially learn latent embeddings of the human and robot motions individually using a VAE. Once the embeddings are learned, the latent interaction dynamics model of Human-Human Interactions (HHI) is learned using LSTMs. The learned HHI dynamics are used in conjunction with the learned robot embeddings to subsequently learn the robot dynamics from HRI demonstrations. However, we find in our experiments that the autoregressive nature of their approach, unfortunately, leads to a divergence in the performance since they only train it using ground truth demonstrations and not in an autoregressive manner. In contrast, rather than having an implicit shared representation, such as with an LSTM, the LfD approaches 
can be explicitly conditioned on the interaction partner's observations~\cite{amor2014interaction,evrard2009teaching} which we find leads to improved predictive performance.

\subsubsection{Unsupervised Skill Discovery}

While skill discovery is not the main focus of our approach, there are some parallels between our approach and works on unsupervised skill discovery which we discuss below. In general, such approaches aim to partition a latent space into different skills which can then be sequenced for learning a variety of different tasks either from demonstrations or using Reinforcement Learning (an overview can be found in Sec. III-B in~\cite{mu2023boosting}). The works closest to our approach~\cite{tanneberg2021skid,nasiriany2023learning,rao2021learning} explore this key idea of learning how to segment demonstrations into underlying recurring segments or \enquote{skills} by decomposing a latent representation of the trajectories and subsequently learning the temporal relations between these skills to compose the observed trajectories. Similar approaches are further explored in literature using different skill representations such as the Options framework~\cite{shankar2020learning} or Autoregressive HMMs~\cite{niekum2012learning,kroemer2015towards}. While the main focus of such approaches is the decomposition of demonstrations to learn re-usable atomic \enquote{skills}, our main aim is to explore how to do so in HRI settings via a joint distribution over the actions of interacting partners and subsequently how the learned skills/segments can be inferred during test time based on the dynamic observations of a human, rather than reproducing a given task or based on a static goal.

\subsection{Objectives and Contributions}
In our previous work~\cite{prasad2022mild} on learning Multimodal Interactive Latent Dynamics, \enquote{MILD}, we explored how Deep LfD can be used for learning spatiotemporally coherent interaction dynamics. Rather than using an uninformed, stationary prior for learning HRI~\cite{butepage2020imitating}, we showed the use of Hidden Semi-Markov Models 
to model the latent space prior of a VAE as a joint distribution over the trajectories of both interacting agents. During test time, the robot trajectories can be conditionally generated based on the observed human trajectories in the latent space, showing that such an approach can capture the latent interaction dynamics suitably well. 
In this article, we extend MILD~\cite{prasad2022mild} by systematically improving the predictive abilities of our model by incorporating reactive motion generation into the training process. We do so by exploring two different sampling approaches to incorporate the conditional distribution of the HMM into the training, rather than using the HMM just as a VAE prior as done in~\cite{prasad2022mild}. Furthermore, to deploy the proposed approach on the Humanoid Social Robot Pepper~\cite{pandey2018mass}, we leverage the similarity between the kinematic structures of Pepper and a Human to effectively learn HRI behaviors from Human-Human Interactions. We further show how the predictions from the latent space HMMs can help supplement the robot controller by modulating the target motions via Inverse Kinematics for a more accurate interaction and modulating the joint stiffnesses for ensuring compliant motions during a contact-rich interaction like a handshake (Figure~\ref{fig:vaeHMM-test}).

Specifically, our contributions beyond MILD~\cite{prasad2022mild} are summarized below:

\begin{enumerate}[label=(\roman*)]
    \item We systematically explore how to incorporate human-conditioned reactive motion generation of the robot trajectories 
    in the training to improve the prediction accuracy of our model.
    \item We integrate Inverse Kinematics in our pipeline to spatially adapt the predicted robot motions to the interaction partner, thereby ensuring suitable physical proximity and improved spatial accuracy of the learned behaviors. 
    \item For contact-rich interactions like handshaking, we perform stiffness modulation using the HMM's segment predictions to enable realistic and compliant robot motions. 
    \item We validate our approach via a user study where participants interact with a Pepper robot controlled by our approach. We find that our method ranks highly compared to other baselines and provides a perceivably more human-like, natural, timely, and accurate interaction, thereby showcasing its effectiveness. We further show to applicability of our approach on a more complex task of Bimanual Robot-to-Human Handovers.
\end{enumerate}

The rest of the paper is organized as follows. In Section~\ref{sec:Foundations}, we explain the foundations needed to understand our work. 
We then explain our approach in Section~\ref{sec:approach}. We highlight our experiments and results in Section~\ref{sec:experiments} and present some concluding remarks and directions for future work in Section~\ref{sec:conclusion}.

\section{Foundations}
\label{sec:Foundations}
We first explain Variational Autoencoders (VAEs) (Section~\ref{ssec:vae}) which is the main backbone of our approach, followed by Hidden Markov Models (HMMs) (Section~\ref{ssec:HMM}), which are key for learning the interaction dynamics, and finally, we give an introduction to Inverse Kinematics (IK) (Sec~\ref{ssec:ik}) which is essential for improving the overall acceptance of our approach.

\subsection{Variational Autoencoders}
\label{ssec:vae}

Variational Autoencoders (VAEs)~\cite{kingma2013auto,rezende2014stochastic} are a type of neural network architecture that learns the identity function in an unsupervised, probabilistic way. The inputs \enquote{$\boldsymbol{x}$} are encoded into lower dimensional latent space embeddings \enquote{$\boldsymbol{z}$} that a decoder uses to reconstruct the original input. A prior distribution is enforced over the latent space during the learning, which is typically a standard normal distribution $p(\boldsymbol{z})=\mathcal{N}(\boldsymbol{z}; \boldsymbol{0, I})$. The goal is to estimate the true posterior $p(\boldsymbol{z|x})$, using a neural network $q(\boldsymbol{z|x})$ and is trained by minimizing the Kullback-Leibler (KL) divergence between them

\begin{equation}
    KL(q(\boldsymbol{z|x})||p(\boldsymbol{z|x})) = \mathbb{E}_q[\log\frac{ q(\boldsymbol{z|x})}{p(\boldsymbol{x,z})}] + \log p(\boldsymbol{x})
\end{equation}
which  can be re-written as 
\begin{equation}
    \label{eq:evidence_log}
    \log p(\boldsymbol{x}) = KL(q(\boldsymbol{z|x})||p(\boldsymbol{z|x})) + \mathbb{E}_q[\log\frac{p(\boldsymbol{x,z})}{q(\boldsymbol{z|x})}].
\end{equation}

The KL divergence is always non-negative, therefore the second term in Eq.~\ref{eq:evidence_log} acts as a lower bound. Maximizing it would effectively maximize the log-likelihood of the data distribution or evidence, and is hence called the Evidence Lower Bound (ELBO), which can be written as
\begin{equation}
    \label{eq:elbo}
    \mathbb{E}_q[\log\frac{p(\boldsymbol{x,z})}{q(\boldsymbol{z|x})}] = \mathbb{E}_q[\log p(\boldsymbol{x|z})] - \beta KL(q(\boldsymbol{z|x})||p(\boldsymbol{z})).
\end{equation}
The first term aims to reconstruct the input via samples decoded from the posterior. The second term is the KL divergence between the prior and the posterior, which regularizes the learning. To prevent over-regularization, the KL divergence term is also weighted down with a factor $\beta$. Further information can be found in~\cite{kingma2013auto,rezende2014stochastic,higgins2016beta}.

\subsection{Hidden Markov Models}
\label{ssec:HMM}
A Hidden Markov Model (HMM) is used to model a sequence of observations $\boldsymbol{z}_{1:T}$ (robot joint angles, human skeletons, etc.) as a sequence of underlying hidden states such that they can \enquote{emit} the observations with a given probability. In mathematical terms, an HMM is characterized by a set of hidden states $i\in\{1, 2 \dots N\}$, each of which denotes a probability distribution, in our case a Gaussian with mean $\boldsymbol{\mu}_i$ and covariance $\boldsymbol{\Sigma}_i$, which characterize the emission probabilities of observations $\mathcal{N}(\boldsymbol{z}_t;\boldsymbol{\mu}_i, \boldsymbol{\Sigma}_i)$. An initial state distribution $\pi_i$ denotes the initial probabilities of being in each state, and the state transition probabilities $\mathcal{T}_{i,j}$ describe the probability of the model going from the $i^{th}$ state to the $j^{th}$ state. 
The sequential progression via the probability of each hidden state given an observed sequence $\boldsymbol{z}_{1:t}$ is denoted by the forward variable of the HMM $\alpha_i(\boldsymbol{z}_t)$
\begin{equation}
\begin{gathered}
\label{eq:hmm-h}
    \alpha_i(\boldsymbol{z}_t) = \frac{\hat{\alpha}_i(\boldsymbol{z}_t)}{\sum\nolimits_{j=1}^N\hat{\alpha}_j(\boldsymbol{z}_t)}\\
\hat{\alpha}_i(\boldsymbol{z}_t) = \mathcal{N}(\boldsymbol{z}_t;\boldsymbol{\mu}_i, \boldsymbol{\Sigma}_i)\sum_{j=1}^N \alpha_j(\boldsymbol{z}_{t-1})\mathcal{T}_{j,i} \\
\hat{\alpha}_i(\boldsymbol{z}_0) = \pi_i \hspace{0.2em} \mathcal{N}(\boldsymbol{z}_0;\boldsymbol{\mu}_i, \boldsymbol{\Sigma}_i)
\end{gathered}
\end{equation}
where $\hat{\alpha}_i(\boldsymbol{z}_t)$ represents the non-normalized forward variable and $\pi_i$ is the initial state distribution. 

The HMM is trained using Expectation-Maximization over the parameters $(\pi_i, \boldsymbol{\mu}_i, \boldsymbol{\Sigma}_i, \mathcal{T}_{j,i})$ with the trajectory data. We refer the reader to~\cite{calinon2016tutorial,pignat2017learning} for further details on HMMs in the context of robot learning.

In HRI scenarios, to encode the joint distribution between the human and the robot, we concatenate the Degrees of Freedom (DoFs) of both the human and the robot~\cite{calinon2009learning,evrard2009teaching}, allowing the distribution to be decomposed as
\begin{equation}
\label{eq:gmr}
    \boldsymbol{\mu}_i = \begin{bmatrix}
\boldsymbol{\mu}^h_i\\
\boldsymbol{\mu}^r_i
\end{bmatrix}; \boldsymbol{\Sigma}_i = \begin{bmatrix}
\boldsymbol{\Sigma}^{hh}_i & \boldsymbol{\Sigma}^{hr}_i\\
\boldsymbol{\Sigma}^{rh}_i & \boldsymbol{\Sigma}^{rr}_i
\end{bmatrix}
\end{equation}
where the superscript indicates the different agents ($h$-human, $r$-robot).
Once the distributions are learned, given some observations of the human agent $\boldsymbol{z}^h_t$, the robot's trajectory can be conditionally generated using Gaussian Mixture Regression as 
\begin{align}
\label{eq:gmr-conditioning-K-i}
   \boldsymbol{K}_i &= \boldsymbol{\Sigma}^{rh}_i(\boldsymbol{\Sigma}^{hh}_i)^{-1}\\
\label{eq:gmr-conditioning-mu-i}
    \boldsymbol{\hat{\mu}}^r_i &= \boldsymbol{\mu}^r_i + \boldsymbol{K}_i(\boldsymbol{z}^h_t - \boldsymbol{\mu}^h_i) \\
    \label{eq:gmr-conditioning-sigma-i}
    \boldsymbol{\hat{\Sigma}}^r_i &= \boldsymbol{\Sigma}^{rr}_i - \boldsymbol{K}_i\boldsymbol{\Sigma}^{hr}_i + \boldsymbol{\hat{\mu}}^r_i(\boldsymbol{\hat{\mu}}^r_i)^T\\
    \label{eq:gmr-conditioning-mu-t}
    \boldsymbol{\hat{\mu}}^r_t &= \sum_{i=1}^N \alpha_i(\boldsymbol{z}^h_t) \boldsymbol{\hat{\mu}}^r_i\\
    \label{eq:gmr-conditioning-sigma-t}
    \boldsymbol{\hat{\Sigma}}^r_t &= \sum_{i=1}^N \alpha_i(\boldsymbol{z}^h_t) \boldsymbol{\hat{\Sigma}}^r_i - \boldsymbol{\hat{\mu}}^r_t(\boldsymbol{\hat{\mu}}^r_t)^T\\
    \label{eq:gmr-conditioning-pz2z1}
    p(\boldsymbol{z}^r_t|\boldsymbol{z}^h_t) &= \mathcal{N}(\boldsymbol{z}^r_t | \boldsymbol{\hat{\mu}}^r_t, \boldsymbol{\hat{\Sigma}}^r_t)
\end{align}

where $\alpha_i(\boldsymbol{z}^h_t)$ is the forward variable calculated using the marginal distribution of the observed human agent. 

\subsection{Inverse Kinematics}
\label{ssec:ik}
Given a joint angle configuration $\boldsymbol{y}$, the end effector position can be calculated as $\boldsymbol{x}_{ee} = f(\boldsymbol{y})$ where $f(\boldsymbol{y})$ denotes the forward kinematics of the robot, which is based on the robot's geometry and calculates the end effector position through the hierarchy of intermediate transformations of each of the individual joints. For example, given the geometry of one's arm sizes, and the angles of the shoulder, elbow, and wrist joints, the position of the hand can be calculated through the position of the shoulder to the elbow to the hand. 

Given a list of $n$ joints $\boldsymbol{y} = \{y_1, y_2 \ldots y_n\}$ and the corresponding relative transformations that the joints represent in the robot geometry $\boldsymbol{T}^0_1(y_1), \boldsymbol{T}^1_2(y_2) \ldots \boldsymbol{T}^{n-1}_n(y_n)$ where $\boldsymbol{T}^i_j \in SE(3)$ is the relative transformation between links $i$ and $j$, the end effector pose $\boldsymbol{T}_{ee}$ can be calculated as
\begin{equation}
    \boldsymbol{T}_{ee} = \begin{bmatrix}
\boldsymbol{R}_{ee} & \boldsymbol{x}_{ee}\\
\boldsymbol{0} & 1
\end{bmatrix} = \boldsymbol{T}^0_n = \prod_{i=1}^n\boldsymbol{T}^{i-1}_i(y_i)
\end{equation}
where $\boldsymbol{x}_{ee}, \boldsymbol{R}_{ee}$ are the end effector position and rotation.

As the name suggests, Inverse Kinematics (IK) is the reverse process of estimating the robot's joint angles from a given end effector pose
, which can be solved via optimization. 
We aim to estimate $\boldsymbol{y} = f^{-1}(\boldsymbol{x}_{ee})$, which can be solved by finding the optimal joint configuration that minimizes the distance between the expected and predicted positions
\begin{equation}
\label{eq:ik-plain}
\boldsymbol{y}^* = \argmin_{\boldsymbol{y}} \hspace{1em} \lVert f(\boldsymbol{y}) -  \boldsymbol{x}_{ee}\rVert^2.
\end{equation}

\section{Multimodal Interactive Latent Dynamics}
\label{sec:approach}

\begin{figure*}[h!]
    \centering
    \begin{subfigure}[b]{\textwidth}
         \centering
         \includegraphics[width=\textwidth]{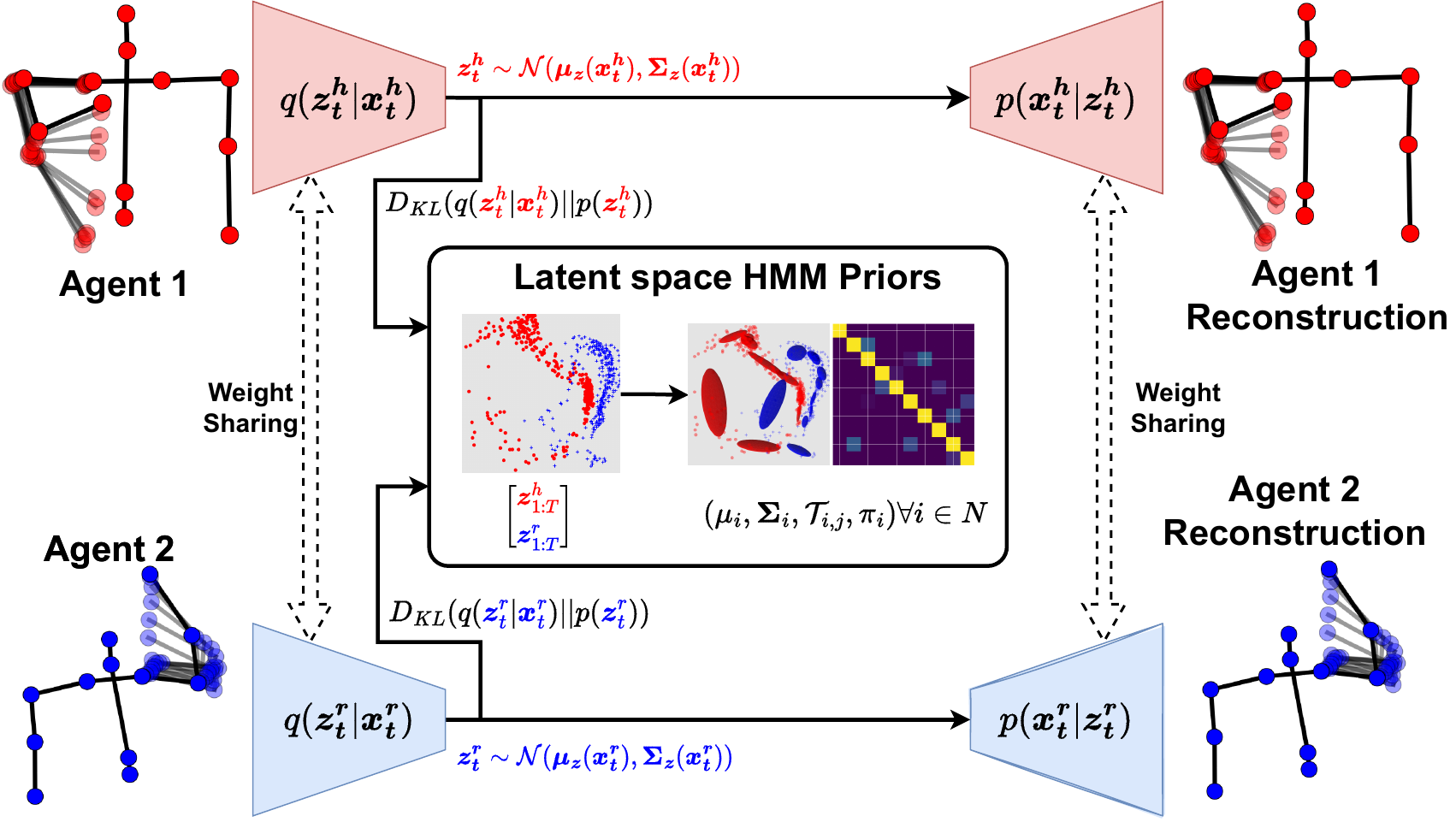}
         \caption{Learning Interaction Models from Human-Human Interactions.}
         \label{fig:training-a}
     \end{subfigure}
     
     \vspace{1em}
     
     \begin{subfigure}[b]{\textwidth}
         \centering
         \includegraphics[width=\textwidth]{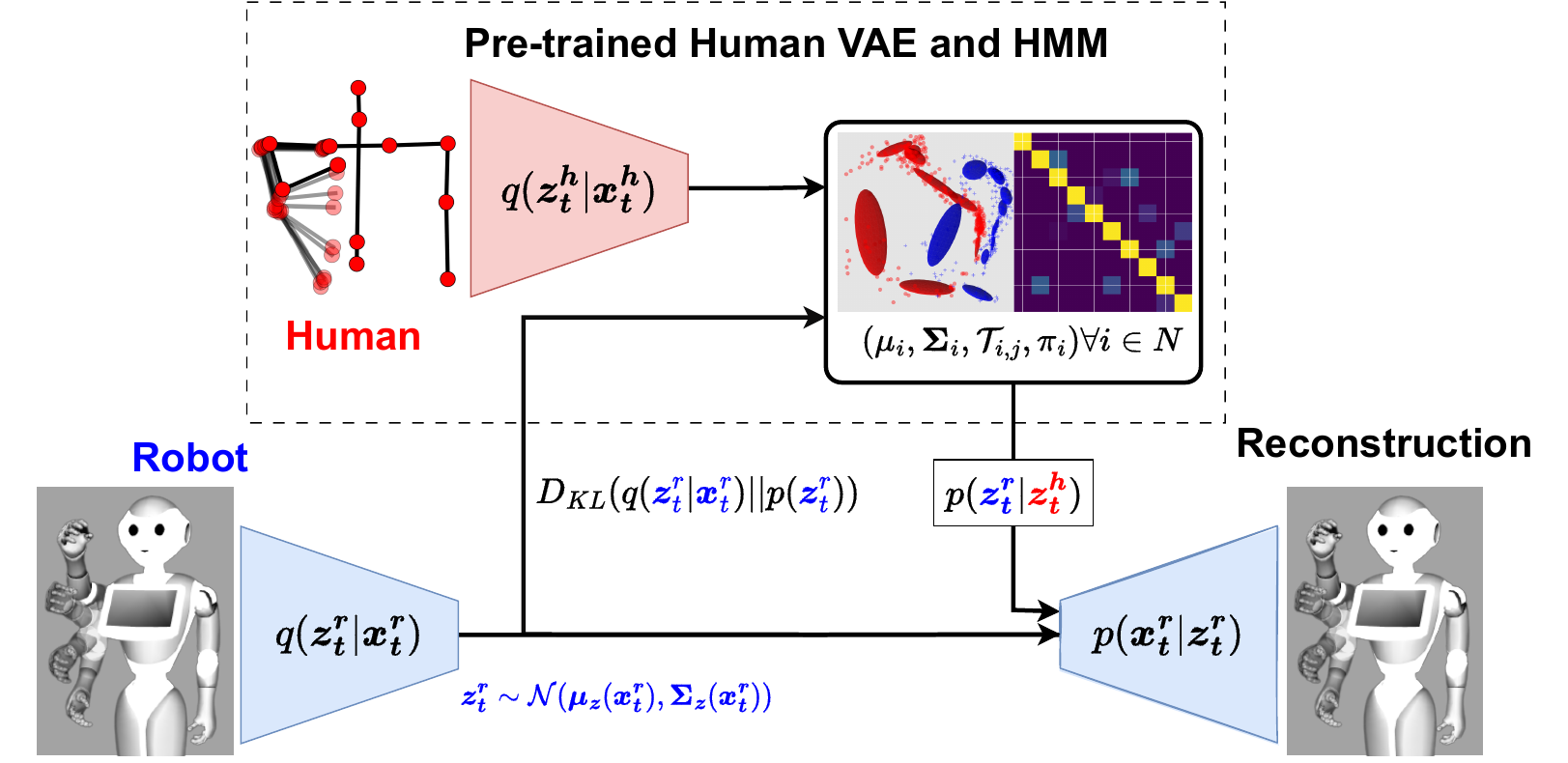}
         \caption{Using Human-Human Interaction models for Human-Robot Interaction models.}
         \label{fig:training-b}
     \end{subfigure}
    \caption{Overview of our training approach. We model the interaction dynamics in the latent space of a VAE using HMMs to model a joint distribution over the latent trajectories of both agents. The use of HMMs for the prior, as opposed to a unimodal Gaussian, enforces a multimodal modularized latent space with the HMM forward variable incorporating the learned transitions between the multiple modes. We first learn the interaction dynamics from Human-Human demonstrations by training both the VAEs and HMMs (Figure~\ref{fig:training-a}). We subsequently use the HMM prior learned from Human-Human demonstrations for learning the latent dynamics for HRI. We do so by regularizing the robot VAEs and, additionally, by training the robot decoder to reconstruct samples from the HMM's conditional distribution after observing the human partner (Figure~\ref{fig:training-b}). This further incorporates the interdependence between the human and the robot during training, thereby improving the predicted robot trajectories.}
    \label{fig:vaeHMM-train}
\end{figure*}

In this section, we introduce our overall approach, which can be seen in Figure~\ref{fig:vaeHMM-train}. We start by explaining the key idea of our previous work MILD~\cite{prasad2022mild}, which enhances the representation learning abilities of VAEs with HMMs as the prior distribution for modeling the interaction dynamics (Section~\ref{ssec:vae-HMM}). 
After introducing our previous work, we go through the key improvements on top of MILD, namely incorporating the human-conditioned HMM distribution for reactive motion generation into the training process in Section~\ref{ssec:cond-kl} followed by the adaptation of the learned trajectories via Inverse Kinematics for improving the generated robot motions as well as the incorporation of stiffness control in Section~\ref{ssec:ik-adaptation}. For HRI scenarios, we use the superscripts $h$ and $r$ to denote the human and the robot variables respectively. For consistency, in the HHI scenarios, we use $h$ and $r$ to denote the variables of the first and second human partners respectively.

\subsection{Learning Interaction Dynamics using HMMs as VAE priors}
\label{ssec:vae-HMM}
Typically in VAEs, the prior $p(\boldsymbol{z})$ is modeled as a stationary distribution. When it comes to learning trajectory dynamics, 
having meaningful priors can help learn temporally coherent latent spaces~\cite{chen2016dynamic}. 
To this end, we explore using HMMs to learn latent-space dynamics in a modular manner by breaking down the trajectories into multiple phases and learning the sequencing between them. 
Since the HMMs learn a joint distribution over the latent trajectories of both interacting agents, we can conditionally generate the motion of the robot after observing the human agent. We do so by using the HMM distribution as the prior for the VAE. We then update the HMM at the end of each epoch by running expectation-maximization on the VAE embeddings, alternating between training the VAE and then subsequently updating the HMM.

The VAE prior at each time-step is calculated as the respective marginal distribution of the most likely HMM component at that timestep, given by the HMM forward variable (Eq.~\ref{eq:hmm-h}). However, the recurrent nature of estimating the forward variable, and consequently its gradients, leads to numerical instabilities from backpropagation through time. Hence, we use an approximation in the form of an unobserved forward variable by setting the likelihood term in Eq.~\ref{eq:hmm-h} to unity. This unobserved forward variable provides a good approximation of the sequential progression of the hidden states based on the learned transitions, which can be written as 
\begin{equation}
\begin{gathered}
    \label{eq:vaeHMM-prior}
    \bar \alpha_i^t = \frac{\hat{\alpha}_i^t}{\sum_{j=1}^N\hat{\alpha}_j^t} \hspace{1em} \hat{\alpha}_i^t = \sum_{j=1}^N \bar \alpha_j^{t-1}\mathcal{T}_{j,i} \hspace{1em} \hat{\alpha}_i^0 = \pi_i \\
    i^*_t = \argmax_i \bar\alpha_i^t \\
    KL_t^h = KL(q(\boldsymbol{z}^h_t|\boldsymbol{x}^h_t)||\mathcal{N}(\boldsymbol{z}^h_t;\boldsymbol{\mu}^h_{i^*_t}, \boldsymbol{\Sigma}^{hh}_{i^*_t}))\\
    KL_t^r = KL(q(\boldsymbol{z}^r_t|\boldsymbol{x}^r_t)||\mathcal{N}(\boldsymbol{z}^r_t;\boldsymbol{\mu}^r_{i^*_t}, \boldsymbol{\Sigma}^{rr}_{i^*_t}))
\end{gathered}
\end{equation}
where $\mathcal{N}(\boldsymbol{z}^h_t;\boldsymbol{\mu}^h_{i^*_t}, \boldsymbol{\Sigma}^{hh}_{i^*_t}))$ and $\mathcal{N}(\boldsymbol{z}^r_t;\boldsymbol{\mu}^r_{i^*_t}, \boldsymbol{\Sigma}^{rr}_{i^*_t})$ are the marginal distributions of the most likely HMM component $i^*_t$ at the given timestep $t$ for each agent. The ELBO can then be reformulated as

\begin{equation}
\begin{aligned}
    \label{eq:vaeHMM-ELBO}
    \mathbb{E}_q[\log\frac{p(\boldsymbol{x}^h_t,\boldsymbol{x}^r_t,\boldsymbol{z}^h_t,\boldsymbol{z}^r_t)}{q(\boldsymbol{z}^h_t,\boldsymbol{z}^r_t|\boldsymbol{x}^h_t,\boldsymbol{x}^r_t)}] &= \mathbb{E}_q[\log p(\boldsymbol{x}^h_t|\boldsymbol{z}^h_t)] \\  &+\mathbb{E}_q[\log p(\boldsymbol{x}^r_t|\boldsymbol{z}^r_t)] \\
    &- \beta (KL_t^h + KL_t^r).
\end{aligned}
\end{equation}

The overall training procedure, shown in Algorithm~\ref{alg:vae-HMM-HHI} and Figure~\ref{fig:training-a}, is as follows. 
We train the VAE to reconstruct the input trajectory and use the marginal distributions of each agent from the HMM (Eq.~\ref{eq:vaeHMM-prior}) to regularize the latent posterior distribution. During a given epoch in the VAE training, the HMM parameters are fixed. After an epoch, the new HMM parameters are estimated using the learned encodings $\boldsymbol{z}^h_{1:T}, \boldsymbol{z}^r_{1:T}$ of the training trajectories. The HMMs are then fixed as the VAE prior for the next epoch. Currently, we learn a separate HMM per interaction. We defer learning the HMM selection via activity recognition to future work. For further details on training HMMs with expectation-maximization, or the use of HMMs in robot learning, we refer the reader to~\cite{dempster1977maximum,calinon2016tutorial}.

\begin{algorithm}[h!]
\caption{Learning Latent Dynamics from Human-Human Interactions}  \label{alg:vae-HMM-HHI}
\small
    \KwData{A set of trajectories with action labels $\boldsymbol{X} =\{\boldsymbol{X}^h_{1:T}, \boldsymbol{X}^r_{1:T}, c\}$ for $|\mathcal{C}|$ interactions}
    \KwResult{VAE weights and $|\mathcal{C}|$ HMM parameters}
    
    Initialize VAE weights randomly\\

    \For {$c \in [1,|\mathcal{C}|]$}{
        \For {$i \in [1,N]$}{
            $\boldsymbol{\mu}^c_i \gets \boldsymbol{0}$ \\
            $\boldsymbol{\Sigma}^c_i \gets \boldsymbol{I}$\\
        }
    }
    \While{not converged}{
        \For {$\boldsymbol{x}^h_{1:T}, \boldsymbol{x}^r_{1:T}, c \in \boldsymbol{X}$}{
            Compute VAE Posterior $q(\boldsymbol{z}^h_t|\boldsymbol{x}^h_t)$ and $q(\boldsymbol{z}^r_t|\boldsymbol{x}^r_t)$\\
            Reconstruct posterior samples\\
            Maximize ELBO (Eq.~\ref{eq:vaeHMM-ELBO}) to update VAE weights \\
        }
    \For {$c \in [1,|\mathcal{C}|]$}{
        $\boldsymbol{X}^c \gets$ set of demonstrations of Interaction $c$\\
        $\boldsymbol{Z}^c \gets \emptyset$ \\
        \For {$\boldsymbol{x}^h_{1:T}, \boldsymbol{x}^r_{1:T}, c \in \boldsymbol{X}^c$}{
            $\boldsymbol{z}^h_{1:T}\sim q(\cdot|\boldsymbol{x}^h_{1:T}) ; \boldsymbol{z}^r_{1:T}\sim q(\cdot|\boldsymbol{x}^r_{1:T})$\\
            $\boldsymbol{Z}^c \gets \boldsymbol{Z}^c \cup \begin{bmatrix}
                \boldsymbol{z}^h_{1:T}\\
                \boldsymbol{z}^r_{1:T}
                \end{bmatrix}$ \\
          }
          Train the $c^{th}$ HMM with $\boldsymbol{Z}^c$
    }}
\end{algorithm}

\subsection{Conditional Training of HRI Dynamics from HHI}
\label{ssec:cond-kl}

Based on the initial idea of~\cite{prasad2022mild} presented above in Section~\ref{ssec:vae-HMM}, in this section, we first explore how the HMMs learned from the Human-Human demonstrations can be used to regularize learning the robot motions, after which we highlight the incorporation of the conditional generation of the robot motions. The HMMs learned from Human-Human demonstrations capture the overall latent interaction dynamics between two agents. These HMMs can, therefore, be used as an informative prior to learn robot motions to perform the given interaction. Hence, we use the marginal distribution of the second agent from the HMMs trained on the HHI demonstrations as the latent space prior for the robot VAE.
Although we regularize the VAEs with the HMM marginals (Eq.~\ref{eq:vaeHMM-prior}), during test time the decoder would see samples from the conditional distribution of the HMM after observing the human agent (Eq.~\ref{eq:gmr-conditioning-K-i}-\ref{eq:gmr-conditioning-pz2z1}) which would be finitely divergent from what the decoder would be trained to reconstruct in a normal autoencoding approach.

When the output spaces of both interaction partners are similar, such as in the HHI scenarios, sharing the weights of the VAEs additionally enables the decoder to learn to reconstruct the target distribution. However, given the difference in output spaces of the human and the robot in HRI scenarios, we instead train the robot decoder to additionally reconstruct samples from conditional distribution (Figure~\ref{fig:training-b}).


Moreover, the VAE provides a confidence estimate of the posterior probability, which can additionally be incorporated into the conditional distribution $p(\boldsymbol{z}_t^r | \boldsymbol{z}_t^h)$ as 
\begin{align}
    \label{eq:conditioning-withcov-K-i}
    \boldsymbol{K}_i &= {\color{orange}\boldsymbol{\Sigma}^{rh}_i}({\color{orange}\boldsymbol{\Sigma}^{hh}_i} + {\color{magenta}\boldsymbol{\Sigma}_{\boldsymbol{z}}(\boldsymbol{x}^h_t)})^{-1} \\
    \label{eq:conditioning-withcov-mu-i}
    \boldsymbol{\hat{\mu}}^r_i &= {\color{orange}\boldsymbol{\mu}^r_i} + \boldsymbol{K}_i({\color{magenta}\boldsymbol{\mu}_{\boldsymbol{z}}(\boldsymbol{x}^h_t)} - {\color{orange}\boldsymbol{\mu}^h_i})\\
    \label{eq:conditioning-withcov-sigma-i}
    \boldsymbol{\hat{\Sigma}}^r_i &= {\color{orange}\boldsymbol{\Sigma}^{rr}_i} - \boldsymbol{K}_i{\color{orange}\boldsymbol{\Sigma}^{hr}_i} + \boldsymbol{\hat{\mu}}^r_i(\boldsymbol{\hat{\mu}}^r_i)^T\\
    \label{eq:conditioning-withcov-mu-t}
    \boldsymbol{\hat{\mu}}^r_t &= \sum_{i=1}^N {\color{orange}\bar\alpha_i^t} \hspace{0.2em} \boldsymbol{\hat{\mu}}^r_i\\
    \label{eq:conditioning-withcov-sigma-t}
    \boldsymbol{\hat{\Sigma}}^r_t &= \left[\sum_{i=1}^N {\color{orange}\bar\alpha_i^t} \hspace{0.2em} \boldsymbol{\hat{\Sigma}}^r_i\right]  - \boldsymbol{\hat{\mu}}^r_t(\boldsymbol{\hat{\mu}}^r_t)^T\\
    \label{eq:conditioning-withcov-pz2z1}
    p(\boldsymbol{z}_t^r | {\color{magenta}q_t^h}) &= \mathcal{N}(\boldsymbol{z}^r_t;\boldsymbol{\hat{\mu}}^r_t, \boldsymbol{\hat{\Sigma}}^r_t)
\end{align}
where the terms in {\color{magenta} magenta} are from the human VAE posterior $q_t^h = q(\boldsymbol{z}^h_t|\boldsymbol{x}^h_t)$ and the terms in {\color{orange} orange} are from the HMM. We then reconstruct samples from $p(\boldsymbol{z}_t^r | {\color{magenta}q_t^h})$ thereby allowing the robot VAE's decoder to be trained with samples drawn from a distribution similar to what it would encounter during test time, instead of just using samples from the posterior. 
Our modified ELBO at each timestep for training the robot VAE in the HRI scenario (Algorithm~\ref{alg:vae-HMM}) can be written as

\begin{equation}
\label{eq:new-elbo}
\begin{aligned}
    \mathbb{E}_q[\log\frac{p(\boldsymbol{x}^h_t,\boldsymbol{x}^r_t,\boldsymbol{z}^r_t)}{q(\boldsymbol{z}^r_t|\boldsymbol{x}^h_t,\boldsymbol{x}^r_t)}] &= \mathbb{E}_{q_r}\log p(\boldsymbol{x}^r_t|\boldsymbol{z}^r_t) \\ 
    & - \beta KL_t^r\\
    & + \mathbb{E}_{\boldsymbol{z}^r_t\sim p(\boldsymbol{z}^r_t|q^h_t)}\log p(\boldsymbol{x}^r_t|\boldsymbol{z}^r_t)
\end{aligned}
\end{equation}

where the first term is the reconstruction term, $KL_t^r$ is the regularization term calculated according to Eq.~\ref{eq:vaeHMM-prior}, and the third term denotes the reconstruction of the conditional samples drawn from $p(\boldsymbol{z}^r_t|q_t^h)$ (Eq.~\ref{eq:conditioning-withcov-pz2z1}). Both the reconstruction term and the conditional reconstruction are estimated in a Monte Carlo fashion by averaging the loss over multiple samples drawn from the corresponding latent distributions. 

\begin{algorithm}[h!]
\caption{Learning HRI Dynamics}  \label{alg:vae-HMM}
\small
    \KwData{A set of trajectories with action labels $\boldsymbol{X} =\{\boldsymbol{X}^h_{1:T}, \boldsymbol{X}^r_{1:T}, c\}$ for $|\mathcal{C}|$ actions, Human VAE and HMMs}
    \KwResult{Robot VAE weights}
    
    Initialize VAE weights randomly\\

    \While{not converged}{
        \For {$\boldsymbol{x}^h_{1:T}, \boldsymbol{x}^r_{1:T}, c \in \boldsymbol{X}$}{
            Compute VAE Posterior $q(\boldsymbol{z}^r_t|\boldsymbol{x}^r_t)$\\
            Compute Latent Conditional $p(\boldsymbol{z}_t^r | q_t^h)$ (Eq.~\ref{eq:conditioning-withcov-K-i}-\ref{eq:conditioning-withcov-pz2z1})\\
            Reconstruct posterior and conditional samples\\
            Maximize ELBO (Eq.~\ref{eq:new-elbo}) to update VAE weights \\
        }      
    }
\end{algorithm}

During testing, we first encode the observations of the human agent $\boldsymbol{x}^h_{t}$ 
and then condition the HMM $p(\boldsymbol{z}_t^r | q_t^h)$ to generate the latent trajectory of the second agent $\hat{\boldsymbol{z}}^r_{t}$ using Eq.~\ref{eq:conditioning-withcov-K-i} -~\ref{eq:conditioning-withcov-pz2z1}, which is then decoded to obtain the actions of the second agent $p(\hat{\boldsymbol{x}}^r_{t}|\boldsymbol{z}^r_{t})$.

\subsection{Inverse Kinematics Adaptation and Stiffness Modulation}
\label{ssec:ik-adaptation}
In typical LfD approaches, robot motions are learned via kinesthetic teaching \cite{paraschos2013probabilistic,schaal2006dynamic}. In HRI scenarios, kinesthetic teaching gives good results \cite{evrard2009teaching,amor2014interaction,campbell2017bayesian,butepage2020imitating} but can become tedious when trying to generalize to multiple human interaction partners. One way to circumvent this could be to execute randomized open-loop trajectories for the robot and learn how the human adapts to the robot motions \cite{campbell2019learning}. 

\begin{figure*}[h!]
    \centering\hfill

    \includegraphics[width=0.1875\textwidth]{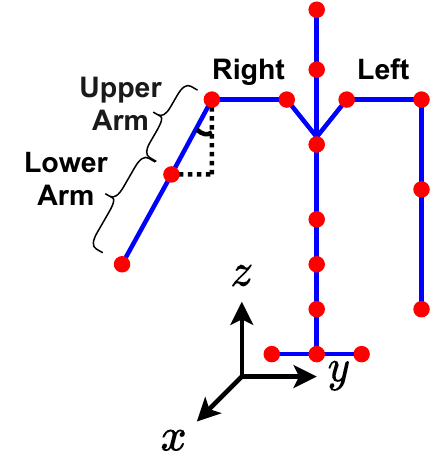} \hfill
    \includegraphics[width=0.12375\textwidth]{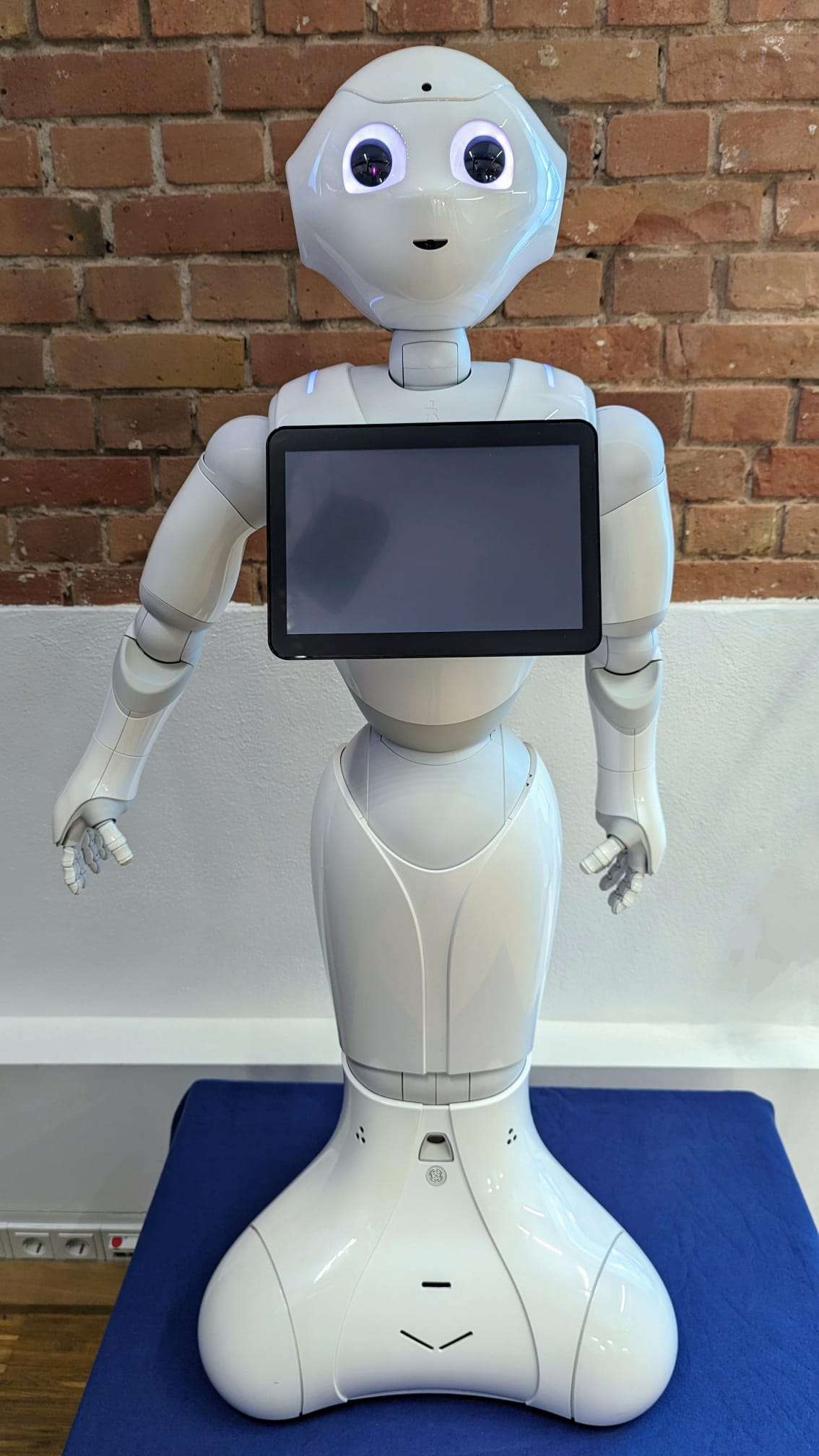} \hfill
    \includegraphics[width=0.165\textwidth]{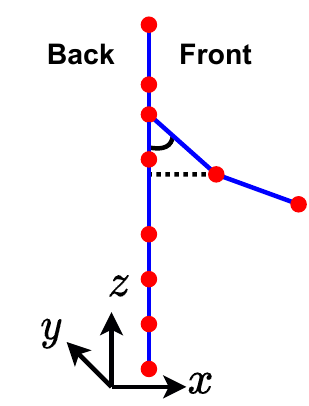} \hfill
    \includegraphics[width=0.12375\textwidth]{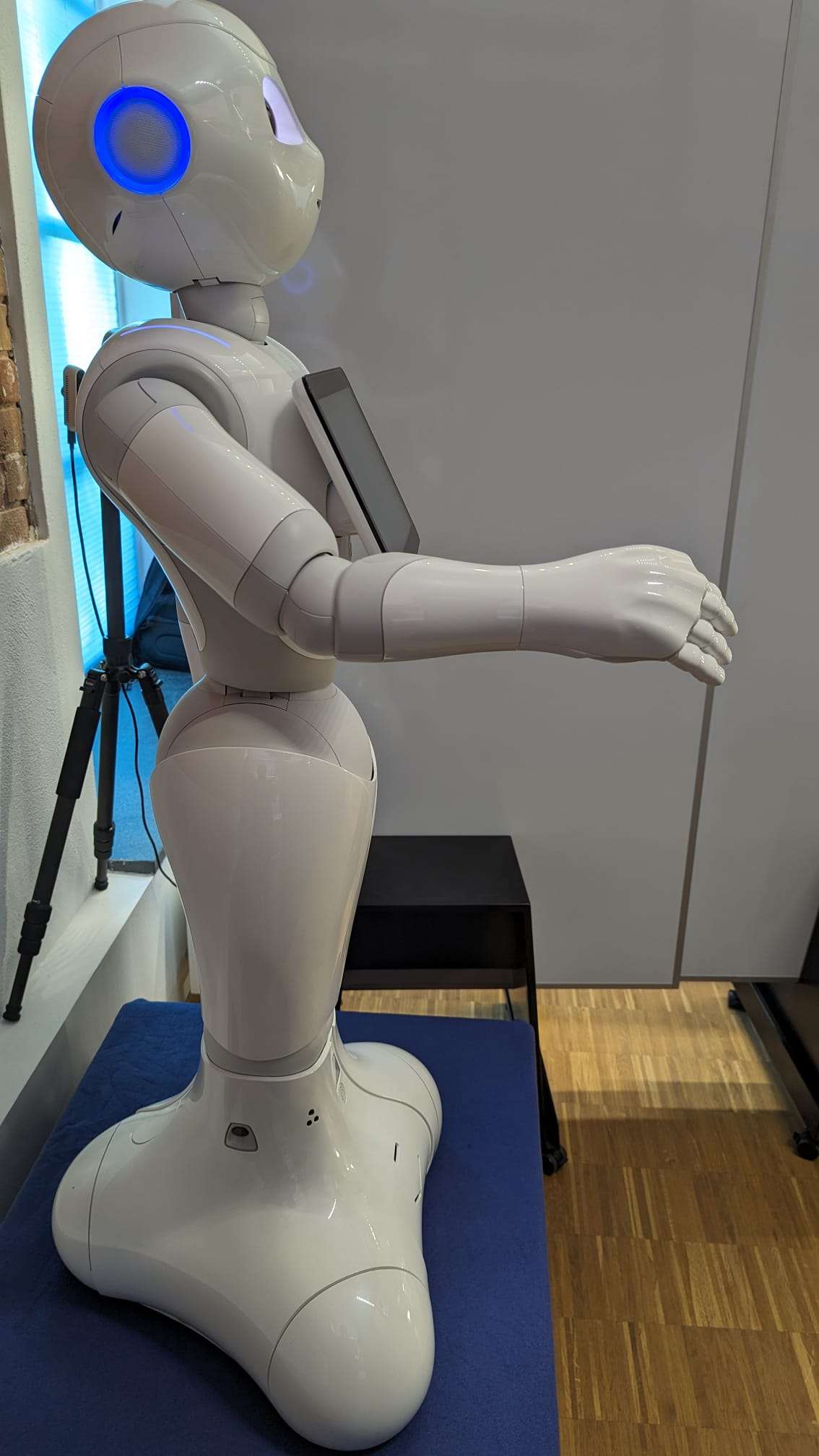} \hfill
    \includegraphics[width=0.135\textwidth]{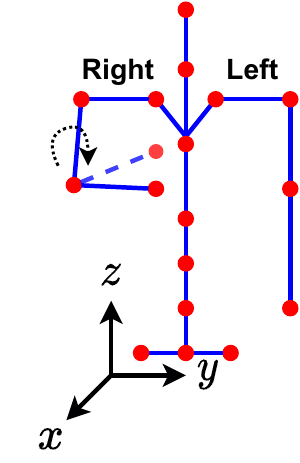} \hfill
    \includegraphics[width=0.12375\textwidth]{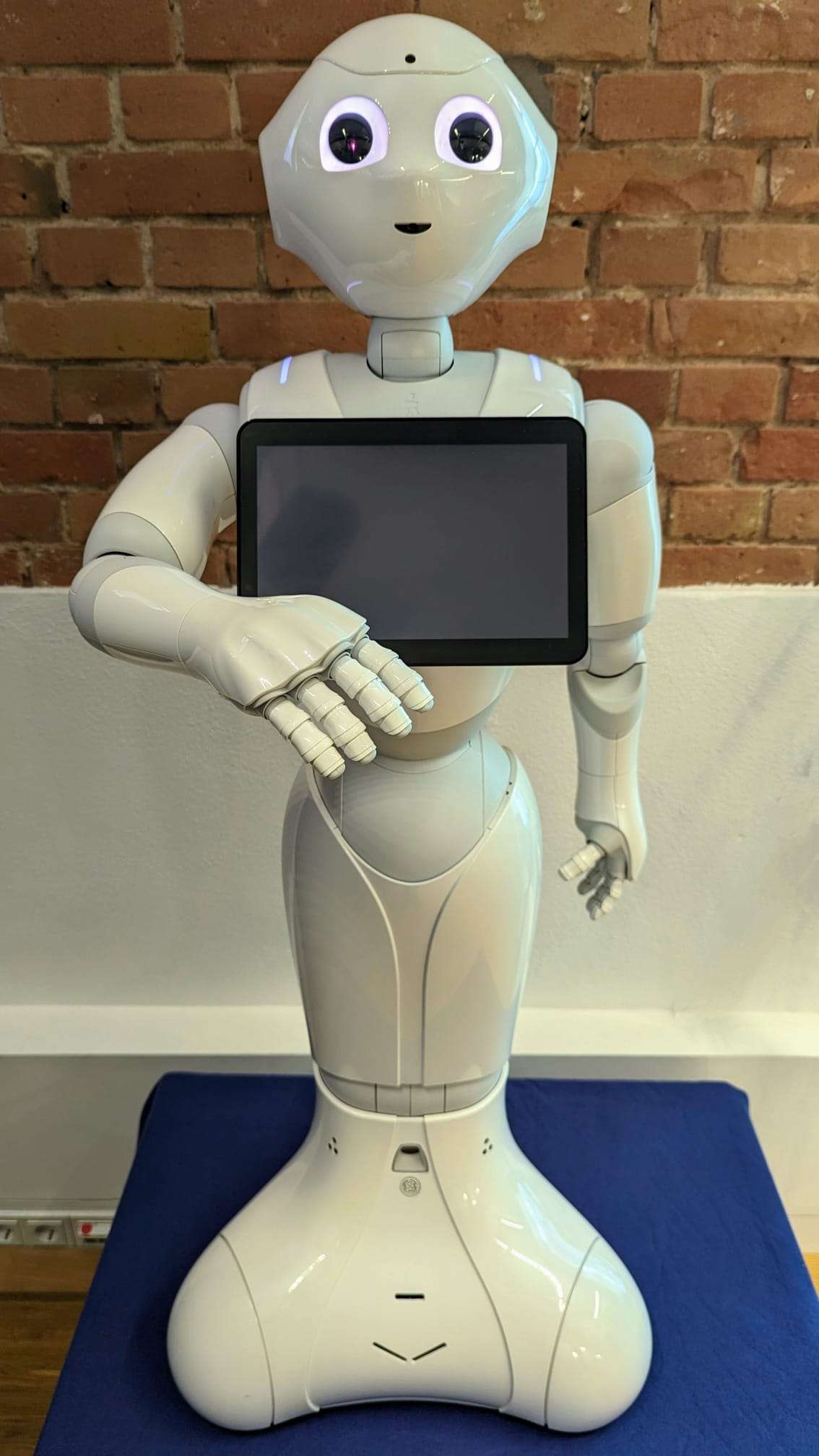}
    
    \caption{Geometric similarities between the degrees of freedom of a human's upper body and the humanoid robot Pepper. The shoulder roll, pitch, yaw, and elbow angle of a human can be directly mapped to Pepper's joint angles.}
    \label{fig:teleoperation}
    \vspace{-1em}
\end{figure*}

Since we are dealing with a Humanoid robot, we learn the robot motions using the kinematic similarities between the human and the robot \cite{fritsche2015first}. From the 3D positions of an arm, we extract the shoulder angles (yaw, pitch, and roll) and the bending of the elbow by using the geometry of the tracked skeleton, as seen in Figure~\ref{fig:teleoperation}. We defer the calculation of the wrist angle to our future work and instead use a fixed value for the wrist for each interaction. 

Since we 
extract joint angles from the human demonstrations, some inaccuracies are present due to slight differences between the kinematic structure of the human skeleton and the robot. Given the difference in the arm dimensions of the Pepper robot and a human, the spatial accuracy of the learned behaviors is quite limited. Additionally, given the smaller size of the Pepper robot, its reachability is also limited, thereby limiting the ability to learn task space trajectories. 

To bridge the mismatch in motion retargeting, we adapt the predicted motions during test time with Inverse Kinematics to reach the human partner's hand while using the predicted motions as a prior. The use of Inverse Kinematics enables the physical proximity needed by the interaction while keeping the robot configuration close to the demonstrated behaviors~\cite{gomez2020adaptation}.  Moreover, we do not need to use Inverse Kinematics all the time, but only in the segments involved in physical contact. We therefore inspect the underlying segments of the HMM manually to see which segments comprise of the contact-based portion of the trajectory and which ones do not, and subsequently perform the IK adaptation only in the contact-based segments.

Given an initial prediction of the robot's joint angle distribution $(\boldsymbol{\mu}_{\boldsymbol{q}}, \boldsymbol{\Sigma}_{\boldsymbol{q}})$ and a task space goal in 3D space $(\boldsymbol{\mu}_{\boldsymbol{x}}, \boldsymbol{\Sigma}_{\boldsymbol{x}})$ (which in this case is the human partner's hand position), we aim to find a joint angle configuration $\boldsymbol{q}^*$ that reaches $\boldsymbol{\mu}_{\boldsymbol{x}}$, but does not stray too far from $\boldsymbol{\mu}_{\boldsymbol{q}}$~\cite{gomez2020adaptation} which boils down to the following optimization problem
\begin{equation}
    \begin{split}
    \boldsymbol{q}^* = \argmaxC_{\boldsymbol{q}} \mathcal{N}&(f(\boldsymbol{q})|\boldsymbol{\mu}_{\boldsymbol{x}}, \boldsymbol{\Sigma}_{\boldsymbol{x}}) \mathcal{N}(\boldsymbol{q}|\boldsymbol{\mu}_{\boldsymbol{q}}, \boldsymbol{\Sigma}_{\boldsymbol{q}})\\
    = \argminC_{\boldsymbol{q}} \lambda_{\boldsymbol{x}} &(\boldsymbol{\mu}_{\boldsymbol{x}} - f(\boldsymbol{q}))^T\boldsymbol{\Sigma}_{\boldsymbol{x}}^{-1}(\boldsymbol{\mu}_{\boldsymbol{x}} - f(\boldsymbol{q}))  + \\
     \lambda_{\boldsymbol{q}} &(\boldsymbol{\mu}_{\boldsymbol{q}} - \boldsymbol{q})^T\boldsymbol{\Sigma}_{\boldsymbol{q}}^{-1}(\boldsymbol{\mu}_{\boldsymbol{q}} - \boldsymbol{q})
    \label{eq:ik-prior}
\end{split}
\end{equation}
where $f(\boldsymbol{q})$ is the forward kinematics model of the robot to estimate the end-effector location given the joint angles $\boldsymbol{q}$, and $\lambda_{\boldsymbol{x}}, \lambda_{\boldsymbol{q}}$ are relative weights balancing whether it is more important to stay close to the initial distribution or to reach the task space goal. In practice, the joint configuration prior $\boldsymbol{\mu}_{\boldsymbol{q}}$ comes from the VAE decoder $p(\boldsymbol{x}^r_t|\boldsymbol{z}^r_t)$, and $\boldsymbol{\mu}_{\boldsymbol{x}}$ comes from the human observation $\boldsymbol{x}^h_t$, and we set $\boldsymbol{\Sigma}_{\boldsymbol{q}}$ and $\boldsymbol{\Sigma}_{\boldsymbol{x}}$ to identity. We can then simplify Eq.~\ref{eq:ik-prior} to

\begin{equation}
    \boldsymbol{q}^* = \argminC_{\boldsymbol{q}} \lambda_{\boldsymbol{x}} \lVert\boldsymbol{\mu}_{\boldsymbol{x}} - f(\boldsymbol{q})\rVert^2 + \lambda_{\boldsymbol{q}} \lVert\boldsymbol{\mu}_{\boldsymbol{q}} - \boldsymbol{q}\rVert^2
    \label{eq:ik-prior-simplified}
\end{equation}
where $\lVert\cdot\rVert$ denotes the Euclidean norm.

Moreover, we do not need to employ IK during every segment of the interaction. Once the training converges, we identify the segments from the underlying HMM that correspond to the contact-based phases of the interaction. Based on the HMM's forward variable prediction from the human observation, we adapt the human-conditioned robot motions using IK if the current segment is contact-based. The incorporation of IK is further detailed in Algorithm~\ref{alg:vae-HMM-test}.

\begin{algorithm}[h!]
\small
\SetAlgoLined
\KwData{An observation of the human agent $\boldsymbol{x}^h_{1:t}$, trained VAE and HMM Models, a set of contact-based HMM states $\mathcal{I}$ for the given interaction}
\KwResult{Conditioned Trajectory for the second agent $\hat{\boldsymbol{x}}^r_{1:t}$}
    \For {$t \in [1,T]$}{
        Encode the human observation  $q(\boldsymbol{z}^h_t|\boldsymbol{x}^h_t)$\\
        Compute Latent Conditional $p(\boldsymbol{z}_t^r | q_t^h)$ (Eq.~\ref{eq:conditioning-withcov-K-i}-\ref{eq:conditioning-withcov-pz2z1})\\
        Decode the conditioned prediction $\boldsymbol{\mu}_{\boldsymbol{q}} = p(\hat{\boldsymbol{x}}^r_t|\hat{\boldsymbol{z}}^r_t)$\\
        \eIf{$\argmaxC_i \alpha_i(\boldsymbol{z}^h_t)$ is a contact-based segment}{
            $\hat{\boldsymbol{q}}_t = \argminC_{\boldsymbol{q}} \lambda_{\boldsymbol{x}} \lVert\boldsymbol{\mu}_{\boldsymbol{x}} - f(\hat{\boldsymbol{q}}_t)\rVert^2 + \lambda_{\boldsymbol{q}} \lVert\boldsymbol{\mu}_{\boldsymbol{q}} - \boldsymbol{q}\rVert^2$
        }{$\hat{\boldsymbol{q}}_t = \boldsymbol{\mu}_{\boldsymbol{q}}$}
        Send $\hat{\boldsymbol{q}}_t$ to the robot controller.
    }
    
 \caption{Conditioning on Human Observations and IK Adaptation}
 \label{alg:vae-HMM-test}
\end{algorithm}

In a contact-rich interaction like handshaking, along with moving in a coordinated manner, the robot must be compliant with the human's motion when in contact. 
When executing handshaking trajectories on the robot, we use the forward variable to detect when to lower the target joint stiffness once the robot enters the contact-based segments of the interaction. Without such as mechanism, the Pepper robot's arm would not comply with the human's hand, and the human would be unable to displace Pepper's hand for executing a proper handshake. 

During a handshaking interaction with the robot, to prevent sudden changes in stiffness arising from the misclassification of the segment, we disable back-transitions into the initial reaching segment. Additionally, since the forward variable is calculated using only the human partner's latent state during test time, we found some mismatches in the segment prediction compared to using the full joint human-robot states for timesteps near the transition boundary between the initial reaching segment and the subsequent segments. Therefore, taking a leaf out of Transition State Clustering~\cite{krishnan2017transition,hahne2024transition}, we learn an additional distribution over the states that get misclassified at the transition boundary between the reaching and contact phase. We use this additional transition state distribution to detect when the interaction proceeds into the contact-based segments when the probability of either the contact segment or the transition state exceeds that of the reaching segment. Doing so gives a better indication of when to lower the robot's stiffness and provides a more suitable interaction. Without such a scheme, due to the misclassification of the active segment, the joint stiffness would not always be lowered correctly, thereby resulting in a rigid and non-compliant handshake.

\section{Experiments and Results}
\label{sec:experiments}
In this section, we first provide our implementation details (Section~\ref{ssec:setup}) and the datasets used (Section~\ref{ssec:datasets}). We then present the results of predicting the controlled agent's trajectories after conditioning on the observed agent in Section~\ref{ssec:motion_pred} and finally, we discuss our user-study in Section~\ref{ssec:user-study} and show some additional results for a Bimanual Robot-to-Human Handover scenario in Section~\ref{ssec:bihandovers}.

\subsection{Experimental Setup}
\label{ssec:setup}
The VAEs are implemented using PyTorch~\cite{paszke2019pytorch} with 2 hidden layers in the encoder and decoder each with sizes (40, 20) and (20, 40) respectively and a 5-dimensional latent space with Leaky ReLU activations~\cite{Maas13rectifiernonlinearities} at all layers except the output layer. 
The weights are initialized using Xavier's initialization~\cite{glorot2010understanding}. The networks are trained with $\beta~=~5\times10^{-3}$, 10 Monte Carlo samples per input data and using the Adam optimizer with weight decay~\cite{loshchilov2018decoupled} with a learning rate of $5\times10^{-4}$. In the HHI scenarios, we share the parameters of the VAEs for both human agents as the inputs are structurally similar. The networks were trained for 400 epochs. The best out of 4 seeds were used to report the results. 

A separate HMM is trained for each interaction. The HMMs are implemented using a custom PyTorch version of PbDLib\footnote{\url{https://git.ias.informatik.tu-darmstadt.de/prasad/pbdlib-torch}}~\cite{pignat2017learning}. 
Each HMM has 6 hidden states, which we found as the most numerically stable. The HMMs are initialized by splitting each of the latent trajectories in the training set into equally sized segments over time. 
To prevent numerical instabilities arising from vanishing values in the HMM covariance matrices, we add a small positive regularization constant of $10^{-4}$ to the diagonal elements. When sampling from the conditional distribution, due to numerical errors in computing the Cholesky decomposition of the covariance matrices, rather than adding a constant value to all elements, we add linearly increasing regularization based on the dimensions, going from $9.1\times10^{-5}$ for the first dimension to $10^{-4}$ for the last dimension. Additionally, we adapt the covariance matrices of the conditional distribution by iteratively adding an even smaller constant that is proportional to the absolute value of the smallest eigenvalue of the covariance matrix to the diagonal elements~\cite{higham1988computing}.

We implement our approach on the Pepper robot~\cite{pandey2018mass}, which is a 1.2m tall humanoid robot that has the same degrees of freedom in its arm as a human. For Inverse Kinematics, we use a modified version of the IKPy library~\cite{manceron_pierre_2022_6551158}
\footnote{\url{https://github.com/souljaboy764/ikpy}}. We set $\lambda_{\boldsymbol{x}}=1$ and $\lambda_{\boldsymbol{q}}=0.01$ for the objective function in Eq.~\ref{eq:ik-prior-simplified}. 

\subsection{Dataset}
\label{ssec:datasets}
\subsubsection{B\"utepage et al.~\cite{butepage2020imitating}}\footnote{\url{https://github.com/souljaboy764/human_robot_interaction_data}}
\label{ssec:buetepage-dataset}record HHI and HRI demonstrations of 4 interactions: Waving, Handshaking, and two kinds of fist bumps. The first fistbump called \enquote{Rocket Fistbump}, involves bumping fists at a low level and then raising them upwards while maintaining contact with each other. The second is called \enquote{Parachute Fistbump} in which partners bump their fists at a high level and bring them down while simultaneously oscillating the hands sideways, while in contact with each other. Since our testing scenario involves the humanoid robot Pepper~\cite{pandey2018mass}, we additionally extract the joint angles from one of the human partner's skeletons from the above-mentioned HHI data for the Pepper robot using the similarities in DoFs between a human and Pepper~\cite{fritsche2015first,prasad2021learning} (Figure~\ref{fig:teleoperation}), which we denote this as \enquote{HRI-Pepper}. B\"utepage et al.~\cite{butepage2020imitating} additionally record demonstrations of these actions with a human partner interacting with an ABB YuMi-IRB 14000 robot controlled via kinesthetic teaching. We call this scenario as \enquote{HRI-Yumi}.

In the HHI scenario (and the HRI-Pepper scenario), there are 181 trajectories (32 - Waving, 38 - Handshake, 70 - Rocket Fistbump, 49 - Parachute Fistbump) of which 80\% of the trajectories (149 trajectories) are used for training and the rest (32 trajectories) for testing. In the HRI-Yumi scenario, there are 41 trajectories (10 each for Waving, Handshaking, and Rocket Fistbump and 11 for Parachute Fistbump) of which we use a similar split with 32 trajectories for training and 9 for testing. We use a time window of 5 observations as the input for a given timestep as done in~\cite{butepage2020imitating}. We downsample the data to 20Hz to match our testing scenario.

We use the 3D positions and the velocities (represented as position deltas) of the right arm joints (shoulder, elbow, and wrist), with the origin at the shoulder, leading to an input size of 90 dimensions (5x3x6: 5 timesteps, 3 joints, 6 dimensions) for a human partner. For the HRI-Pepper and HRI-Yumi scenarios, we use a similar window of joint angles, leading to an input size of 20 dimensions (5x4) for the 4 joint angles of Pepper's right arm and an input size of 35 dimensions (5x7) for the 7 joint angles of Yumi's right arm.

\subsubsection{Nuitrack Skeleton Interaction Dataset}
\label{ssec:nuisi-dataset}(NuiSI) \footnote{\url{https://github.com/souljaboy764/nuisi-dataset}} is a dataset which we collected ourselves of the same 4 interactions as in~\cite{butepage2020imitating}, namely Waving, Handshaking, Rocket Fistbump and Parachute Fistbump. While the dataset in B\"utepage et al.~\cite{butepage2020imitating} has clean motions of the interaction partners, they require high-accuracy inertial sensors in whole-body suits. Therefore, for a more realistic setting, we capture the interaction partners using low-cost hardware.

While other datasets exist that perform skeleton tracking with a Kinect Camera~\cite{shahroudy2016ntu,van2016spatio,shu2016learning}, the participants in these datasets are recorded from the side. Therefore, due to partial occlusions, there is a high degree of noise in skeleton tracking, which made it difficult to train a suitable model for the interactions. Therefore, we record two human partners interacting with one another using one Intel Realsense D435 camera per partner (two in total), such that each camera captures a full frontal view of the interaction. We use Nuitrack\footnote{\url{https://nuitrack.com/}}
for tracking the upper body skeleton joints in each frame at 30Hz. As done above with~\cite{butepage2020imitating}, we additionally extract the joint trajectories for the Pepper robot, which leaves us finally with both the HHI and HRI-Pepper scenarios. 
We have 44 trajectories (12 - Waving, 11 - Handshaking, 12 - Rocket Fistbump, 9 - Parachute Fistbump), of which we similarly use 80\% of the trajectories for training (33 trajectories) and the rest (11 trajectories) for testing. 

The skeletons are then rotated such that they follow the orientation shown in Figure~\ref{fig:teleoperation} with the $x$-axis in the forward direction, the $y$-axis going from right to left, and the $z$-axis in the upward direction. For training, the data is processed similarly as mentioned above with a window size of 5 time steps. Given the relatively low amount of samples, we finetune the network trained on the aforementioned dataset of~\cite{butepage2020imitating}. As done with~\cite{butepage2020imitating}, we extract the joint angles for the Pepper robot from the skeleton trajectories for the HRI-Pepper scenario~\cite{fritsche2015first,prasad2021learning}.



\subsection{Conditioned Prediction Results}
\label{ssec:motion_pred}

We test the conditioning ability of our approach compared to~\cite{butepage2020imitating}\footnote{Results reported using our implementation of~\cite{butepage2020imitating}.} to evaluate the accuracy of the generated motions of the controlled agent after observing the human interaction-partner. We evaluate the approaches over the four interactions of the dataset in \cite{butepage2020imitating} and our collected data (NuiSI). We calculate the Mean Squared Error (MSE) averaged over each joint and the 5-time step window.

\begin{table}[h!]
    \caption{Prediction MSE (in cm) for the second human partner's trajectories after observing the first human partner averaged over all joints and timesteps. (* -- $p<0.05$, ** -- $p<0.01$, lower is better)}
    \label{tab:pred-hhi}
    \centering
\begin{tabular}{|c|c|l|c|}
\hline
Dataset & Action & \multicolumn{1}{c|}{MILD~v1} & \cite{butepage2020imitating} \\ \hline
\multirow{4}{*}{\cite{butepage2020imitating}} & Hand Wave & \textbf{0.788 $\pm$ 1.226}$^{**}$ &  4.121 $\pm$ 2.252 \\ \cline{2-4} 
& Handshake          & \textbf{1.654 $\pm$ 1.549}$^*$ &  1.181 $\pm$ 0.859 \\ \cline{2-4} 
& Rocket Fistbump    & \textbf{0.370 $\pm$ 0.682} &  0.544 $\pm$ 1.249 \\ \cline{2-4} 
& Parachute Fistbump & \textbf{0.537 $\pm$ 0.579}$^{**}$ &  0.977 $\pm$ 1.141 \\ \hline
\multirow{4}{*}{NuiSI} & Hand Wave & \textbf{0.408 $\pm$ 0.538}$^{**}$ & 3.168 $\pm$ 3.392 \\ \cline{2-4} 
& Handshake          & \textbf{0.311 $\pm$ 0.259}$^{**}$ & 1.489 $\pm$ 3.327 \\ \cline{2-4} 
& Rocket Fistbump    & \textbf{1.142 $\pm$ 1.375}$^{**}$ & 3.576 $\pm$ 3.082 \\ \cline{2-4} 
& Parachute Fistbump & \textbf{0.453 $\pm$ 0.578}$^{**}$ & 2.008 $\pm$ 2.024 \\ \hline
\end{tabular}
\end{table}

\begin{table*}[h!]
    \centering
        \caption{Prediction MSE (in radians) for robot trajectories after observing the human interaction partner averaged over all joints and timesteps. (Lower is better, lowest values highlighted in bold, significance shown in Table~\ref{tab:hri-significance})}
    \label{tab:pred-hri}
\resizebox{0.99\textwidth}{!}{
\begin{tabular}{|c|c||c|c|c|c|c||c|}
\hline
Dataset & Action & MILD~v1 & MILD~v2.1 & MILD~v2.2 & MILD~v3.1 & MILD~v3.2 & B\"utepage et al. \cite{butepage2020imitating}\\ \hline
\multirow{4}{*}{HRI-Yumi \cite{butepage2020imitating}} & Hand Wave & 1.705 $\pm$ 0.521 & 1.349 $\pm$ 1.972 & 1.641 $\pm$ 1.968 & 1.033 $\pm$ 1.204 & 1.143 $\pm$ 1.330 & \textbf{0.225 $\pm$ 0.302}\\ \cline{2-8} 
 & Handshake & 0.290 $\pm$ 0.148 & 0.073 $\pm$ 0.040 & \textbf{0.068 $\pm$ 0.052} & 0.104 $\pm$ 0.056 & 0.123 $\pm$ 0.069 & 0.133 $\pm$ 0.214 \\ \cline{2-8} 
 & Rocket Fistbump & 0.428 $\pm$ 0.175 & 0.236 $\pm$ 0.167 & 0.183 $\pm$ 0.122 & 0.130 $\pm$ 0.074 & \textbf{0.128 $\pm$ 0.071} & 0.147 $\pm$ 0.119 \\ \cline{2-8} 
 & Parachute Fistbump & 0.425 $\pm$ 0.150 & 0.028 $\pm$ 0.042 & 0.033 $\pm$ 0.034 & \textbf{0.028 $\pm$ 0.034} & 0.028 $\pm$ 0.035 & 0.181 $\pm$ 0.155\\ \hline
\multirow{4}{*}{HRI-Pepper \cite{butepage2020imitating}} & Hand Wave & 0.267 $\pm$ 0.152 & 0.161 $\pm$ 0.228 & 0.165 $\pm$ 0.189 & \textbf{0.103 $\pm$ 0.103} & 0.106 $\pm$ 0.105 & 0.664 $\pm$ 0.277 \\ \cline{2-8} 
 & Handshake & 0.327 $\pm$ 0.253 & 0.111 $\pm$ 0.092 & 0.153 $\pm$ 0.154 & 0.061 $\pm$ 0.048 & \textbf{0.056 $\pm$ 0.041} & 0.184 $\pm$ 0.141\\ \cline{2-8} 
 & Rocket Fistbump & 0.161 $\pm$ 0.095 & 0.035 $\pm$ 0.068 & 0.035 $\pm$ 0.068 & 0.021 $\pm$ 0.037 & \textbf{0.018 $\pm$ 0.035} &  0.033 $\pm$ 0.045 \\ \cline{2-8} 
 & Parachute Fistbump & 0.265 $\pm$ 0.178 & 0.116 $\pm$ 0.176 & 0.112 $\pm$ 0.181 & 0.095 $\pm$ 0.151 & \textbf{0.088 $\pm$ 0.148} & 0.189 $\pm$ 0.196 \\ \hline
\multirow{4}{*}{HRI-Pepper (NuiSI)} & Hand Wave & 0.760 $\pm$ 0.325 & 0.050 $\pm$ 0.084 & 0.060 $\pm$ 0.087 & \textbf{0.046 $\pm$ 0.059} & 0.049 $\pm$ 0.059 & 0.057 $\pm$ 0.093 \\ \cline{2-8} 
 & Handshake & 0.225 $\pm$ 0.114 & 0.025 $\pm$ 0.022 & 0.025 $\pm$ 0.020 & 0.021 $\pm$ 0.015 & \textbf{0.020 $\pm$ 0.014} & 0.083 $\pm$ 0.075 \\ \cline{2-8} 
 & Rocket Fistbump & 0.354 $\pm$ 0.238 & 0.077 $\pm$ 0.095 & 0.080 $\pm$ 0.088 & \textbf{0.077 $\pm$ 0.067} & 0.079 $\pm$ 0.072 & 0.101 $\pm$ 0.086  \\ \cline{2-8} 
 & Parachute Fistbump & 0.201 $\pm$ 0.072 & 0.032 $\pm$ 0.038  & 0.028 $\pm$ 0.040 & 0.025 $\pm$ 0.028 & \textbf{0.022 $\pm$ 0.027} & 0.049 $\pm$ 0.040 \\ \hline
\end{tabular}
}

\end{table*}

We do not train with the conditional loss in the Human-Human scenarios since we use shared weights, the decoder already learns to reconstruct the ground truth samples for the conditional distribution. Additionally, we found that the interplay between the shared weights and the conditional training discussed in Section~\ref{ssec:cond-kl} would cause the HMM posterior to collapse into a single unimodal distribution. 
Therefore, we only show comparisons of our vanilla approach presented in Section~\ref{ssec:vae-HMM}, which we denote as \enquote{MILD~v1}. 

The results of MILD~v1 in predicting the interaction partner's trajectories in the HHI scenarios can be seen in Table~\ref{tab:pred-hhi}. It can be seen that MILD~v1 with its simplistic nature of using an HMM for the latent dynamics performs significantly better (as verified with a Mann-Whitney U Test) than~\cite{butepage2020imitating} where the VAEs are trained with an uninformative standard normal distribution as a prior. Although additional LSTMs are employed in~\cite{butepage2020imitating} to learn the latent dynamics, since the VAEs are trained with an uninformative prior, their approach fails to accurately reconstruct motions the learnt latent dynamics. In contrast, the latent dynamics of each segment of the interaction is captured well by the HMM which therefore acts as an informative prior for the VAEs which is reflected in the improved prediction accuracy of MILD~v1.

Coming to HRI scenarios, starting from the initial incorporation of the HMM prior as shown in Section~\ref{ssec:vae-HMM} (denoted as \enquote{MILD~v1}), we explore two variants of the last conditional training term in Eq.~\ref{eq:new-elbo}.
The first is with reconstructing the conditional latent predictions (i.e. the mean in Eq.~\ref{eq:conditioning-withcov-mu-t}) of samples drawn from the posterior distribution both with and without the use of the VAE posterior covariance (in Eq.~\ref{eq:conditioning-withcov-K-i}). This can be summarized mathematically as $\mathbb{E}_{q(\boldsymbol{z}^h_t|\boldsymbol{x}^h_t)}\log p(\boldsymbol{x}^r_t|\boldsymbol{\mu}^r_t)$ where $\boldsymbol{\mu}^r_t$ is calculated using Eq.~\ref{eq:gmr-conditioning-mu-t} with samples drawn from $q(\boldsymbol{z}^h_t|\boldsymbol{x}^h_t)$. 
We explore this variant both without and with the posterior covariance in Eq.~\ref{eq:conditioning-withcov-K-i}, denoted as \enquote{MILD~v2.1} and \enquote{MILD~v2.2} respectively
    
The second variant uses the posterior mean and covariance to calculate the conditional distribution and subsequently reconstruct samples drawn from the conditional distribution. This can be summarized mathematically as $\mathbb{E}_{p(\boldsymbol{z}^r_t|q^h_t)}\log p(\boldsymbol{x}^r_t|\boldsymbol{z}^r_t)$ where $p(\boldsymbol{z}^r_t|q^h_t)$ is calculated using Eq.~\ref{eq:conditioning-withcov-K-i}~-~\ref{eq:conditioning-withcov-pz2z1} and the samples drawn from this distribution are reconstructed. 
We explore this variant both without and with the posterior covariance in Eq.~\ref{eq:conditioning-withcov-K-i}, which we denote as \enquote{MILD~v3.1} and \enquote{MILD~v3.2} respectively. 

The key differences between the different variants are highlighted in Table~\ref{tab:mild_variants}. To summarize, the prior means and covariances from the HMM are used by both variants in the conditioning, denoted by the orange terms in Eq.~\ref{eq:conditioning-withcov-K-i}~-~\ref{eq:conditioning-withcov-pz2z1}. The key difference between the variants is how the posterior distribution terms (denoted in magenta in Eq.~\ref{eq:conditioning-withcov-K-i}~-~\ref{eq:conditioning-withcov-pz2z1}) are used. In MILD~v3.1 and MILD~v3.2, we directly use the posterior mean and covariance as shown in Eq.~\ref{eq:conditioning-withcov-K-i}~-~\ref{eq:conditioning-withcov-pz2z1}. In the case of MILD~v2.1 and MILD~v2.2, we do not directly use the posterior mean and covariance, but first draw samples from the posterior distribution $\boldsymbol{z}_t^h \sim \mathcal{N}({\color{magenta}\boldsymbol{\mu}_{\boldsymbol{z}}(\boldsymbol{x}^h_t)}, {\color{magenta}\boldsymbol{\Sigma}_{\boldsymbol{z}}(\boldsymbol{x}^h_t)})$ which are then used in the conditioning in Eq.~\ref{eq:conditioning-withcov-mu-i} instead of the posterior mean ${\color{magenta}\boldsymbol{\mu}_{\boldsymbol{z}}(\boldsymbol{x}^h_t)}$ which can be written as $\boldsymbol{\hat{\mu}}^r_i = {\color{orange}\boldsymbol{\mu}^r_i} + \boldsymbol{K}_i(\boldsymbol{z}_t^h - {\color{orange}\boldsymbol{\mu}^h_i})$ to directly get the conditional samples that then get reconstructed.

\begin{table*}
\caption{Differences in conditioning and sampling strategies of the variants of MILD.}
    \label{tab:mild_variants}
    \centering
    \begin{tabular}{|c|c|c|} \hline 
         Variant &  Conditional Training Inputs & Samples given to Decoder\\ \hline 
         MILD~v1 &  None & Only Posterior Samples\\ \hline 
         MILD~v2.1 &  Posterior Samples & \multirow{2}{*}{Posterior Samples and their corresponding conditioned outputs}\\ \cline{1-2}
         MILD~v2.2 &  Posterior Samples and Covariance & \\ \hline 
         MILD~v3.1 &  Posterior Mean & \multirow{2}{*}{Posterior Samples, Conditional Samples}\\ \cline{1-2}
         MILD~v3.2 &  Posterior Mean and Covariance & \\ \hline
    \end{tabular}
    
    \vspace{-2em}
\end{table*}

As seen in Table~\ref{tab:pred-hri}, reconstructing samples from the conditional distribution (MILD~v3.1 and v3.2) provides much better results as compared to conditioning samples drawn from the posterior (MILD~v2.1 and v2.2), both of which perform better than MILD~v1 that does not use conditional training. The samples drawn from the conditional distribution would be more representative of the type of samples that the decoder would see during run time. While enough samples from the posterior, when conditioned, can also estimate this distribution, empirically, this fails to match up to sampling from the conditional distribution. One argument for this is that reconstructing conditional samples enables learning a joint latent space more susceptible to the HMM conditioning.

Furthermore, the overall improvement in performance compared to MILD~v1 and~\cite{butepage2020imitating} highlights the advantage of incorporating conditional prediction into the training process for reactive motion generation. The importance of incorporating reactive motion generation into the training can also be seen in the improved performance of~\cite{butepage2020imitating} over MILD~v1 in the HRI scenarios as compared to the HHI scenarios.
This improvement comes from the fact that MILD uses the HMMs just as a latent prior, whereas in~\cite{butepage2020imitating},  the authors explicitly train a separate HRI dynamics model for predicting the robot motions from the latent trajectories of both the human and the robot. During testing, however, the HRI dynamics network does not have access to the ground truth target of the robot which is used to train the network. Therefore, the HRI dynamics is predicted in an autoregressive manner, which deteriorates the performance due to out-of-distribution data. In this regard, it can be seen that explicitly incorporating such conditional out-of-distribution samples can lead to better results, as seen in the improved performance of the variants of MILD (v2.1 - v3.2).

Some examples of the learned behaviors, along with the progression of the HMM in the latent space can be seen in Figure~\ref{fig:hri-samples}, where we show a sample interaction for handshake (Figure~\ref{fig:hri-handshake}) and rocket fistbump (Figure~\ref{fig:hri-rocket}) on the Pepper robot. As it can be seen, the HMM captures the sequencing between the multiple modes of the latent space to generate suitable motions for real-world HRI scenarios. This is additionally validated via a user study (Section~\ref{ssec:user-study}) which shows the ability of our approach to generalize well to various users, despite being trained on demonstrations of just two partners.




\subsection{HRI User Study}
\label{ssec:user-study}
To see the effectiveness of our approach in producing acceptable physical behaviors, we conducted a user study where participants interacted with the robot. We evaluate our proposed approach both with and without IK adaptation, denoted as \enquote{MILD-IK} and \enquote{MILD} (while we use \enquote{MILD~v3.2}, for ease of notation, we shorten it to \enquote{MILD}) against a baseline IK algorithm (Eq.~\ref{eq:ik-plain}) that uses the human's hand pose as the target location (\enquote{Base-IK}). We run this study for two interactions, handshake and rocket fistbump. The study design was given a positive vote by the Ethics Commission at TU Darmstadt (Application EK 48/2023).

\begin{figure}[h!]
    \centering
    \includegraphics[width=0.6\linewidth]{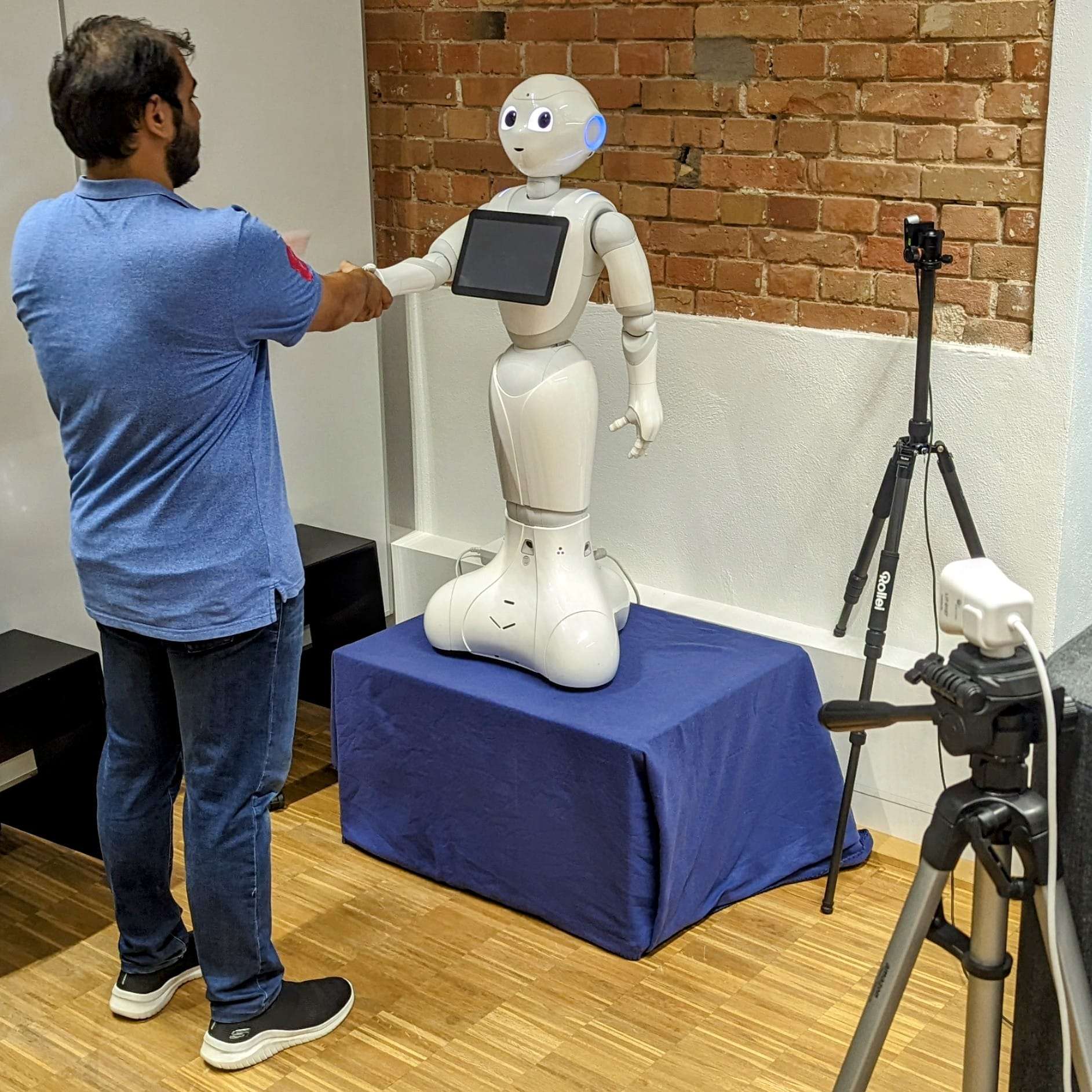}
    \caption{Setup of the User Study for interacting with the Pepper robot. As Pepper is quite short (1.2m), it is placed on a pedestal to match the height of a human partner. The camera behind Pepper tracks the human partner's motion, which is used to generate Pepper's motions.}
    \label{fig:study_setup}
    \vspace{-2em}
\end{figure}

\input{tabs_n_figs/hri_images}

\subsubsection{Procedure}
The study took place in a laboratory setting (Figure~\ref{fig:study_setup}) where participants were first guided to a desk to fill out a consent form and a pre-questionnaire which included demographic information, prior experiences with robots, their attitude towards robots~\cite{syrdal2009negative}, attitude towards physical interactions, and personality questions to gauge extroversion~\cite{rammstedt2007measuring}. We assessed these measures to gain additional information about our sample. Participants were then shown an instruction video of two humans performing an interaction (either a handshake or a rocket fistbump). Participants were then given the experiment protocol to read wherein they were instructed that they had to lead the interaction with the robot. 
Additionally, some general instructions were provided to the participants about how they should position themselves to maintain their posture and the limit of the arm movements they can perform. This was done to prevent occlusions in the skeleton tracking and to maintain the robot’s reachability limits.

Participants were then guided behind a barrier where they would see the robot for the first time and would then stand at an adequate distance from the robot. When they were ready, an initial interaction was performed with a hard-coded motion. 
This is to get the participant habituated to the way the robot moves and counter any novelty effects that may occur and to have the participant better understand how to perform the interaction with the robot. After the initial run, the participant was informed that the experimental trials would begin. The sequence of each trial was as follows:
\begin{enumerate}
    \item Before each trial, the participant would see a video stream of the skeleton tracker to better position themselves.
    \item Once the position was set and the tracking was stable, the experimenter would signal them to start the trial.
    \item The participant would start the interaction and the robot would respond reactively.
    \item Once the participant goes back to a neutral position with their hands by their side, the robot arm would go back to a neutral position, marking the end of the trial.
    \item This process would then repeat once again after which the participant was asked to fill out a questionnaire. 
\end{enumerate}
This process 
constitutes a single session and was repeated 3 times (3 sessions in total) wherein the robot was controlled by one of the three aforementioned algorithms (Base-IK, MILD, MILD-IK). The algorithms were shown in randomized order to all participants to avoid sequence effects. The participant was neither informed of the randomization nor the algorithm they were interacting with. After all 3 sessions, the participant was asked to fill out a final questionnaire where they had to rank the sessions based on their preference and answer some open-ended questions about the sessions. 

\subsubsection{Participant Sample}
A total of 20 users (8 female, 12 male) participated in our study and were recruited through the university environment. Of the 20 participants, 10 performed the Rocket fistbump (4 female, 6 male) and the rest performed a handshake. The mean age of the participants was 27.85 years (SD: 3.51). 
Participants had an average level of experience with robots overall on a scale from 1 (no experience at all) to 5 (a lot of experience) with a mean of 2.90 (SD: 1.45). They had quite a positive attitude towards robots on a scale from 1 (very negative) to 7 (very positive) with a mean of 5.90 (SD: 1.16). This is also consistent with the ratings regarding the attitudes towards robots for the following items on a scale of 1 (Strongly Agree) to 7 (Strongly Disagree). Participants largely had a high agreement towards feeling relaxed when interacting with robots (Mean: 5.60, SD: 1.05) and high levels of disagreement towards being paranoid when interacting with a robot (Mean: 1.85, SD: 0.93) and towards feeling nervous standing in front of a robot (Mean: 1.80, SD:1 .06). Participants had a positive outlook towards physical interactions in general on a scale from 1 (distant) to 7 (open) with a mean of 6.10 (SD: 0.84). This was also confirmed with the Big 5 extroversion scale \cite{rammstedt2007measuring} with a mean extroversion of 4.75 (SD: 1.34) out of 7. 

\subsubsection{Methodology}
The study followed a within-subject design where participants interacted with a Pepper robot controlled by each of the algorithms (Base-IK, MILD and MILD-IK) twice in a randomized order, leading to 6 HRIs per participant. Since humans perceive different types of physical touch differently~\cite{gallace2010science,saarinen2021social,burgoon1991relational}, we aimed to remove any influence the type of interaction (handshake or fistbump) might have on how participants view the robot during different interactions. We wanted to keep the focus on understanding how our proposed algorithm is perceived by users. Therefore, each participant had to either perform a handshake or a fistbump, not both. Additionally, to prevent any sudden jumps in the motion of the robot, a weighted moving average filter was used to smoothen out the predicted motions of the robot.

We break the 6 trials into 3 sessions, where each session corresponds to two trials of a given algorithm. Each session was evaluated with 16 different items adapted from the Godspeed~\cite{bartneck2008godspeed} and the SASSI~\cite{hone2000towards} questionnaires, each rated on a 5-Point scale (1 - Strongly disagree, 5 - Strongly agree):
\begin{itemize}
        \item The interaction with the robot was pleasant.
        \item The interaction with the robot was exciting.
        \item The interaction with the robot was human-like.
        \item The interaction with the robot was natural.
        \item The interaction with the robot was friendly.
        \item The interaction with the robot was comfortable.
        \item The interaction with the robot was well-timed.
        \item The interaction with the robot was accurate.
        \item The interaction with the robot was annoying.
        \item The interaction with the robot was awkward.
        \item The interaction with the robot was scary.
    \item The robot interacted in an aggressive way.
    \item I am satisfied with the way the robot interacted with me.
    \item The second trial in the session was more effortless than the first.
\end{itemize}

At the end of the experiment, we asked the participants to rank the 3 sessions (algorithms) in their order of preference.

We run a one-way Repeated Measures ANOVA to compare the responses of the different algorithms followed by paired sample t-tests for a post hoc analysis (Table~\ref{table:anova-results}).

\begin{figure}[h!]
    \centering
    \includegraphics[width=0.8\linewidth]{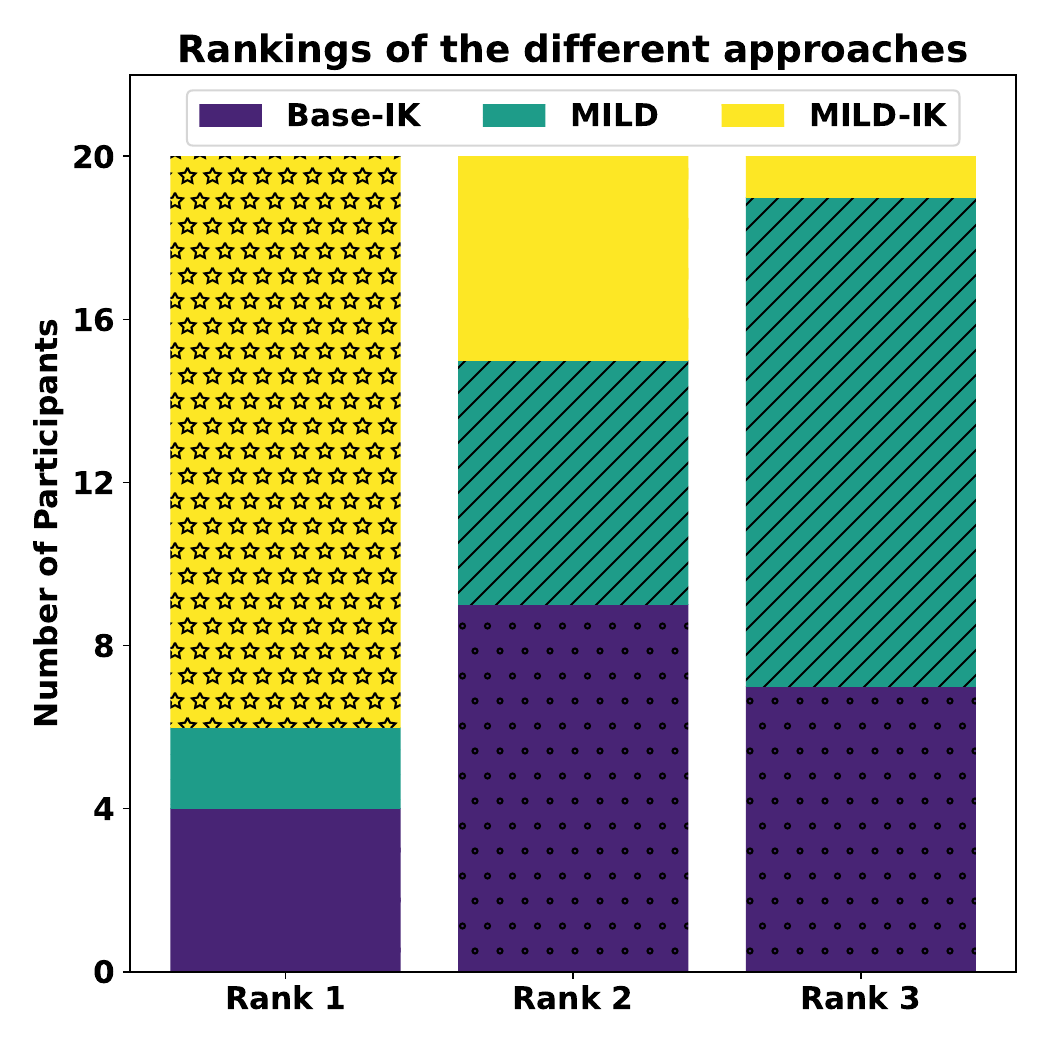}
    \vspace{-1em}
    \caption{Participant's ranking of the algorithms. (MILD-IK - yellow stars, MILD - green lines, Base-IK - purple dots). Most participants ranked MILD-IK first, better than both Base-IK and MILD.}
    \label{fig:algo-rankings}
    \vspace{-2em}
\end{figure}


\subsubsection{Study Results}

\begin{figure*}[h!]
    \centering
    \includegraphics[width=\linewidth]{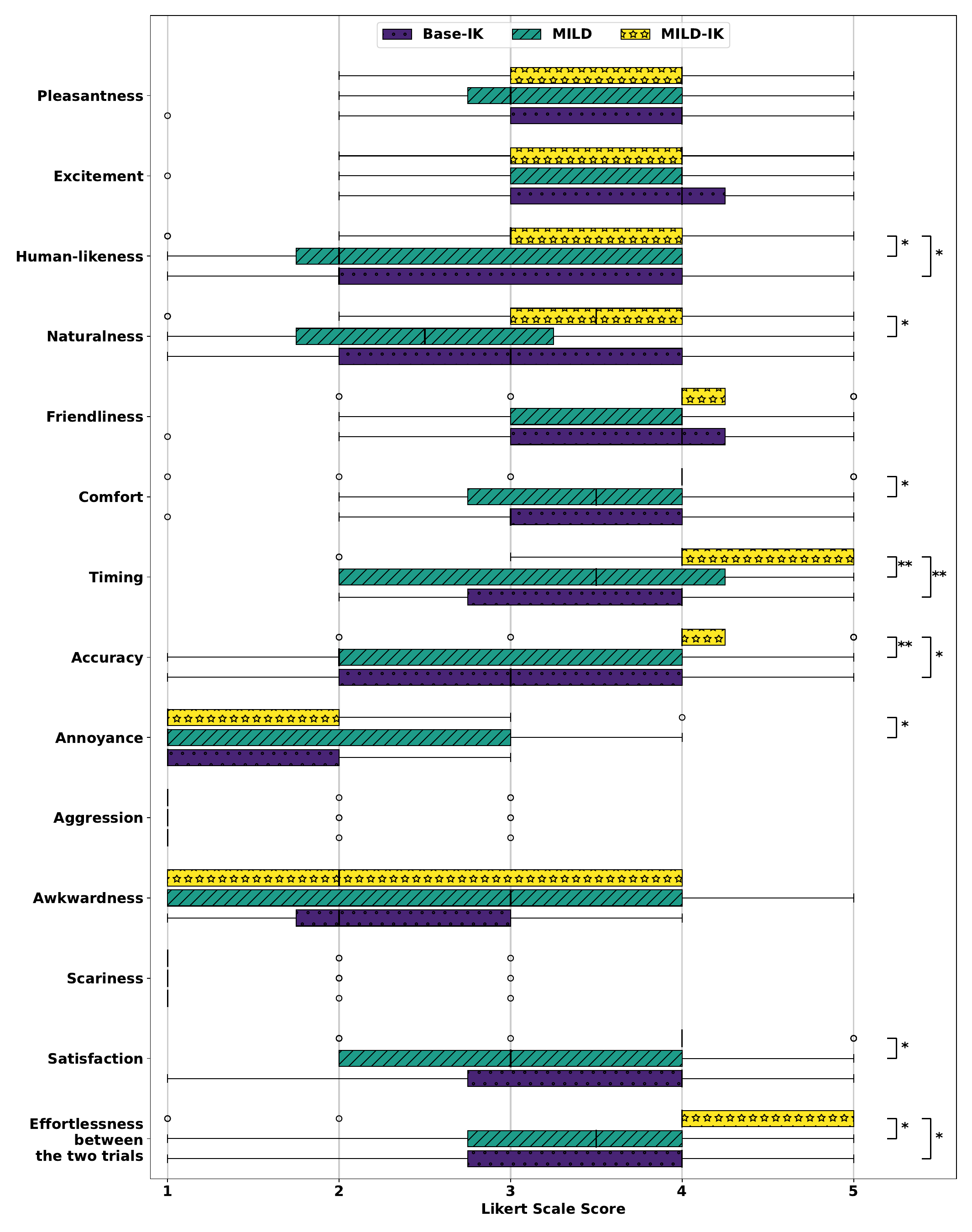}
    \caption{Boxplot of the comparison between the three different approaches. (* - $p < 0.05$, ** - $p < 0.01$)}
    \label{fig:userstudy_boxplot}
\end{figure*}

As seen in Figure~\ref{fig:algo-rankings}, MILD-IK was by far ranked in the first place much more than MILD which was by far the least ranked, and Base-IK which was mostly ranked second. Just using MILD without IK was the least preferred among the algorithms, even though it was programmed keeping the interactiveness in mind, but due to the motion retargeting issues mentioned in Section~\ref{ssec:ik-adaptation}, it is unable to reach the participant's hand accurately, leading to a low acceptance by the participants which can further be seen in the results in Figure~\ref{fig:userstudy_boxplot}.
All approaches have similarly high levels of pleasantness, excitement, and friendliness and similarly low levels of annoyance, aggression, and scariness. This goes to show that the overall interaction scenario was well perceived by the participants and can additionally be attributed to an overall positive attitude toward robots. 

Since the Base-IK approach does not have any prior over the IK solutions, it would lead to poses where the elbow is pointed outwards and the hand is turned sideways, which participants remarked was unnatural and awkward. Even though it was able to reach the human's hand location, this unnatural pose of the robot hand prevented the Base-IK approach from being able to reach the human hand in a graspable manner, due to the orientation of the end-effector. This is also reflected via a significantly lower accuracy rating for Base-IK and can further be seen in relatively lower (although non-significant) trends for naturalness, comfort, and satisfaction.

Participants found MILD awkward at times since the predicted motions would sometimes fall short of the partner's hand and the robot would thereby not reach the participants' hand correctly. This can especially be seen in the low accuracy that is given to MILD. This reachability issue gets mitigated with MILD-IK as it generates more natural poses. 
However, with MILD-IK, since the prediction of when to start the IK adaptation comes from the HMM, there would be a distinct moment when the robot hand would reach the human's hand, which we attribute as a possible reason for a relatively higher awkwardness rating. 

\begin{figure}
    \centering
    

    \begin{tabular}{ccc}
         \rotatebox[origin=lc]{90}{Base-IK}&\includegraphics[width=0.18\textwidth]{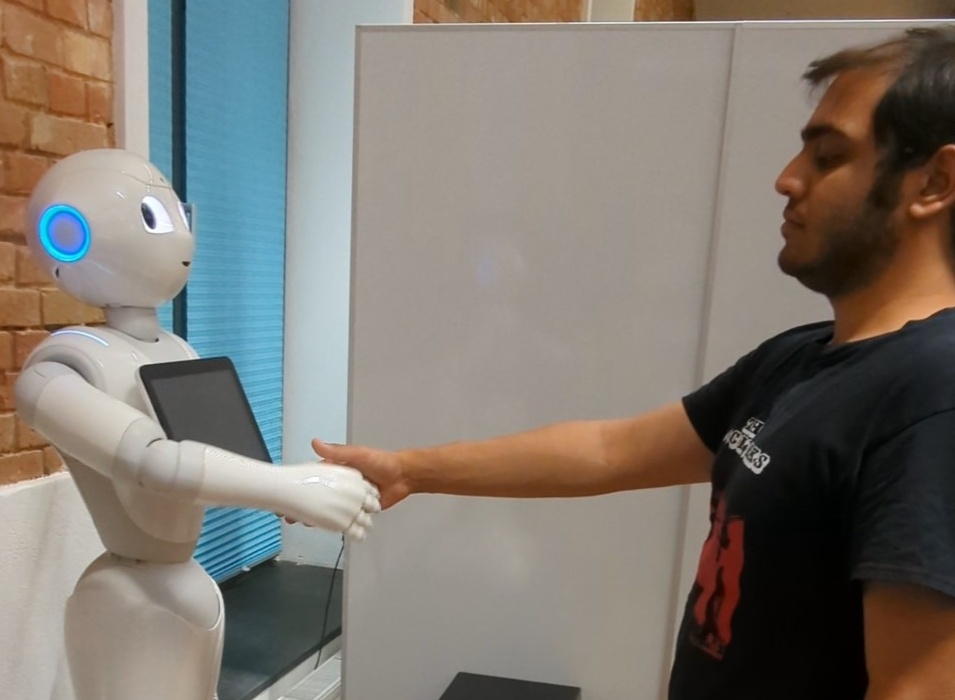} & \includegraphics[width=0.18\textwidth]{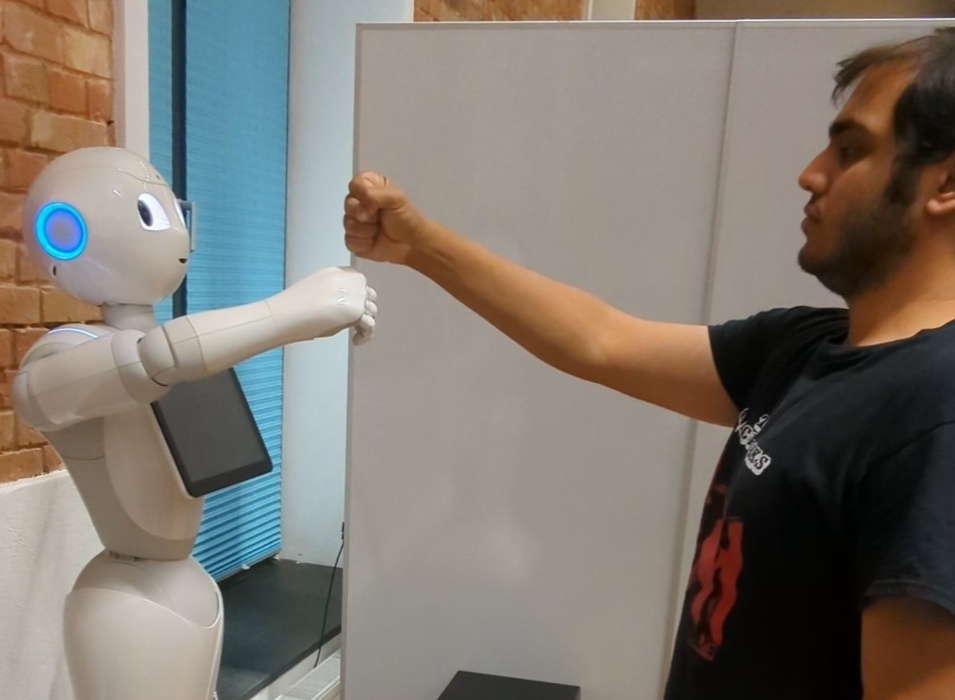} \\
        \rotatebox[origin=lc]{90}{MILD}&\includegraphics[width=0.18\textwidth]{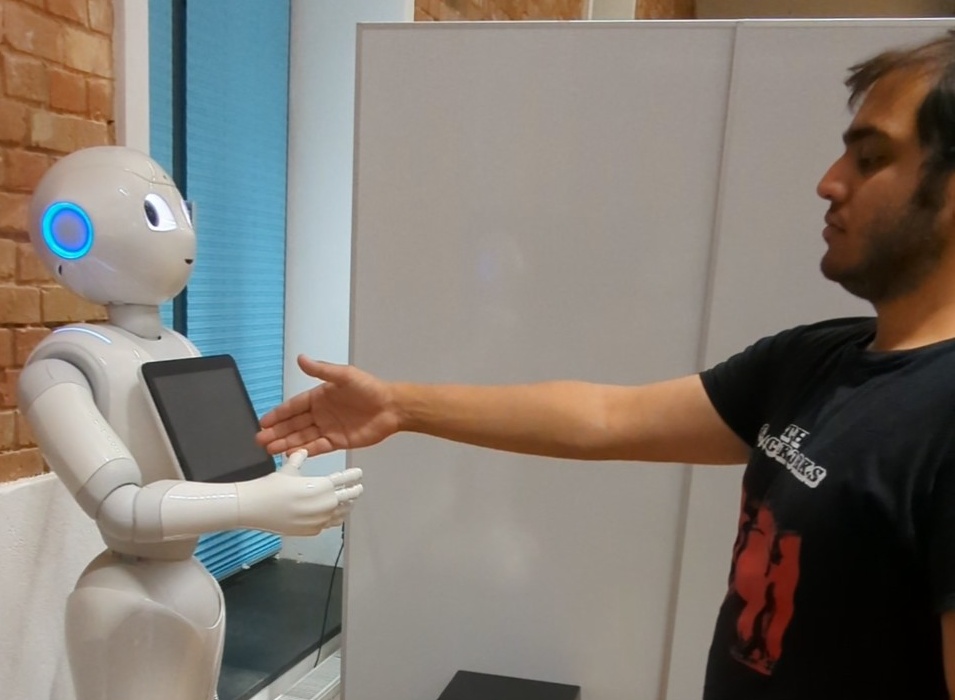} & \includegraphics[width=0.18\textwidth]{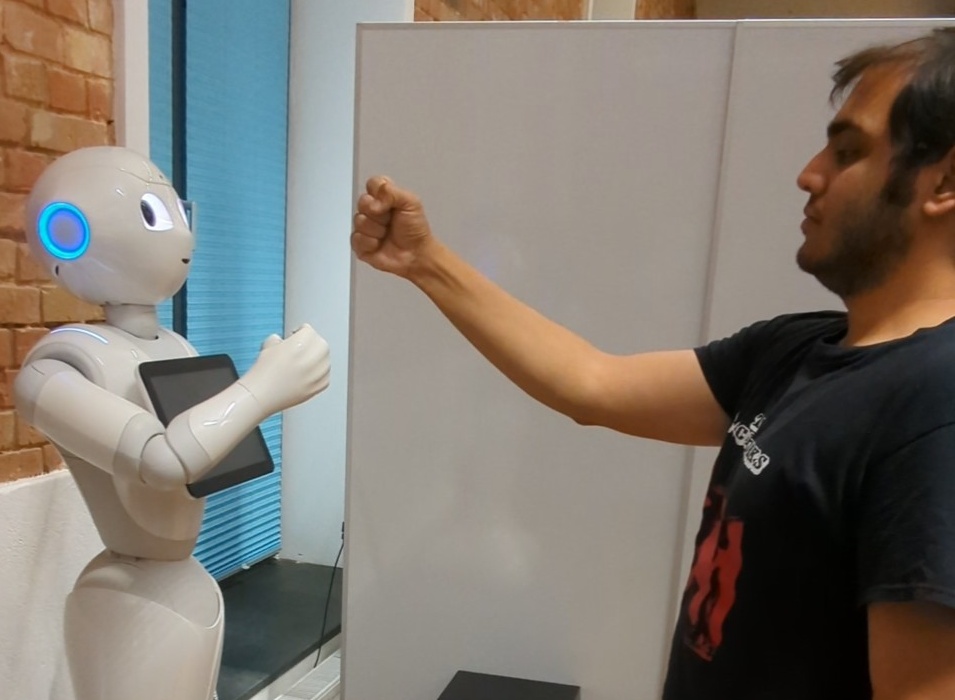} \\
        \rotatebox[origin=lc]{90}{MILD-IK}&\includegraphics[width=0.18\textwidth]{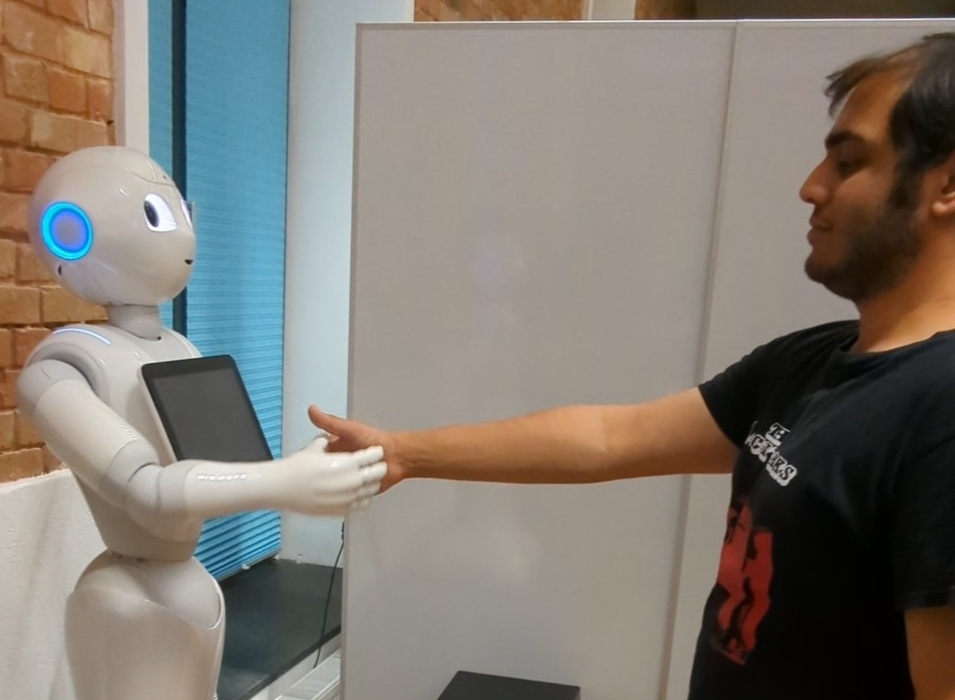} & \includegraphics[width=0.18\textwidth]{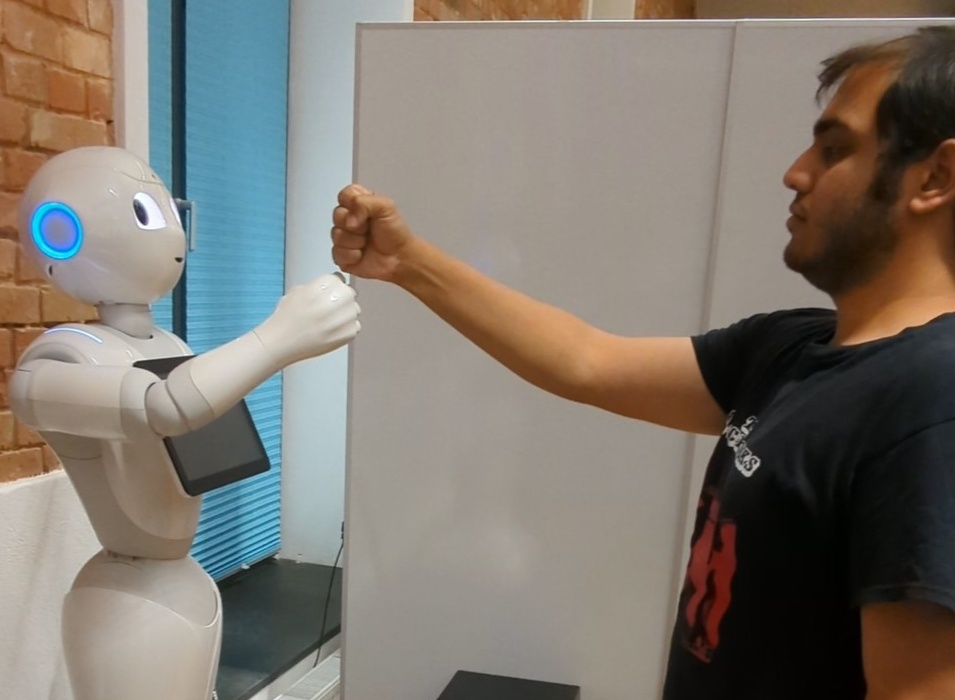}
    \end{tabular}
    
    \caption{Peculiarities of the different algorithms. The \enquote{Base-IK} approach reaches the human's hand but awkwardly with the robot hand rotated inwards and the elbow pointing out. MILD maintains a human-like posture but falls short of reaching the partner's hand due to the mismatch in motion retargeting. However, MILD-IK accurately reaches the human's hand while maintaining a human-like posture.}
    \label{fig:oddities}
    \vspace{-2em}
\end{figure}

Overall, the high acceptability of MILD-IK is also reflected via significantly better ratings of the human-likeness, timing, and perceived accuracy. The effortlessness between two trials in an interaction was also rated significantly higher for MILD-IK showing that it was easier for participants to get habituated to the movements of the robot. MILD-IK also achieves a higher rating for timing, which is attributed to the ability of the HMM to generate the receding motion in a reactive manner, unlike Base-IK which would still try to reach the hand until the participant would go back to a neutral pose. Some of these peculiarities can be seen in Figure~\ref{fig:oddities}.

\subsubsection{Objective Performance Metrics}

\begin{figure}[h!]
    \centering
    \includegraphics[width=\linewidth]{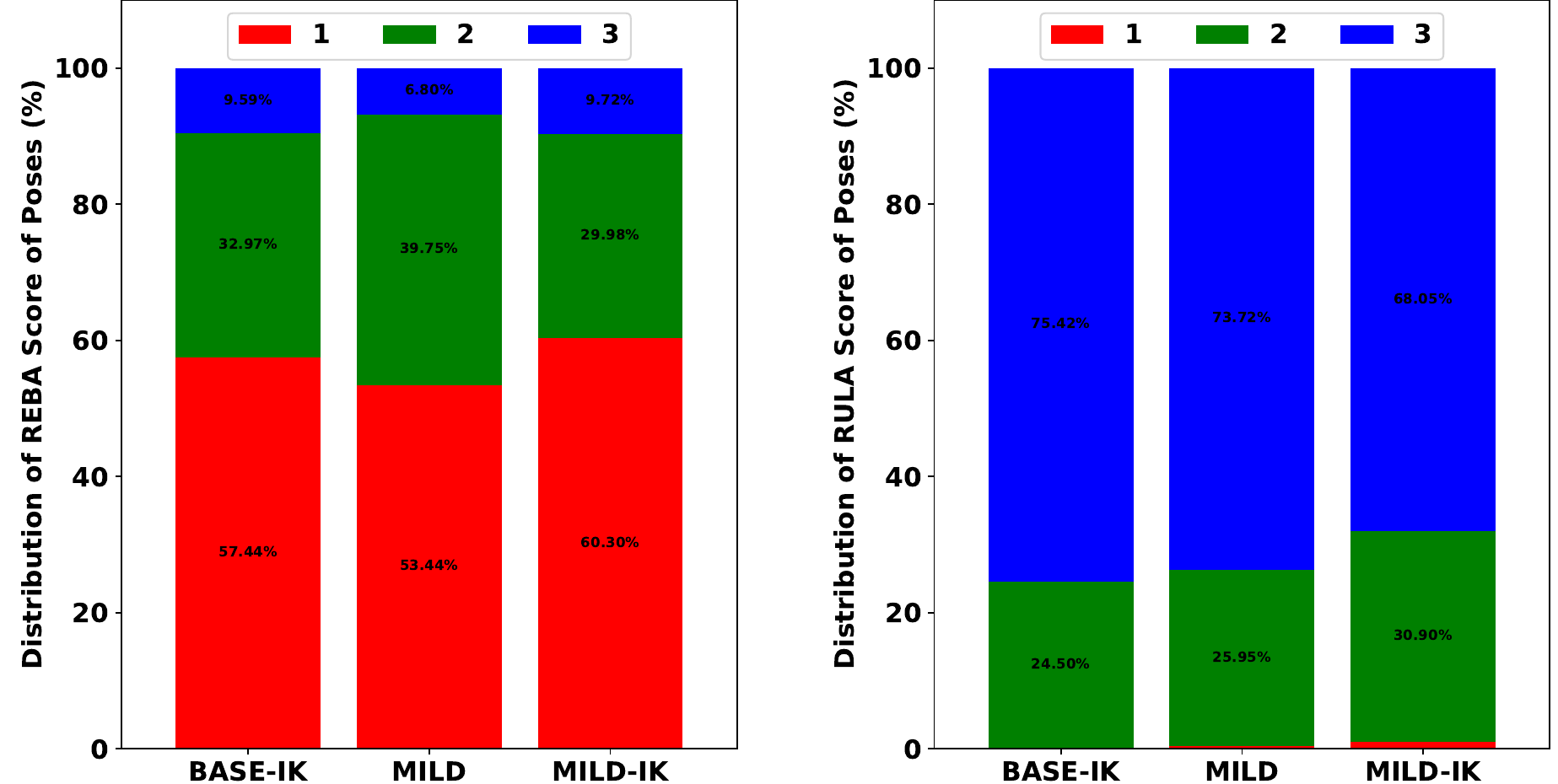}
    \caption{Distribution of REBA and RULA scores across all the poses of the human partner for different algorithms.}
    \label{fig:ergonomics}
    \vspace{-2em}
\end{figure}

\begin{figure}[h!]
    \centering
    \includegraphics[width=0.7\linewidth]{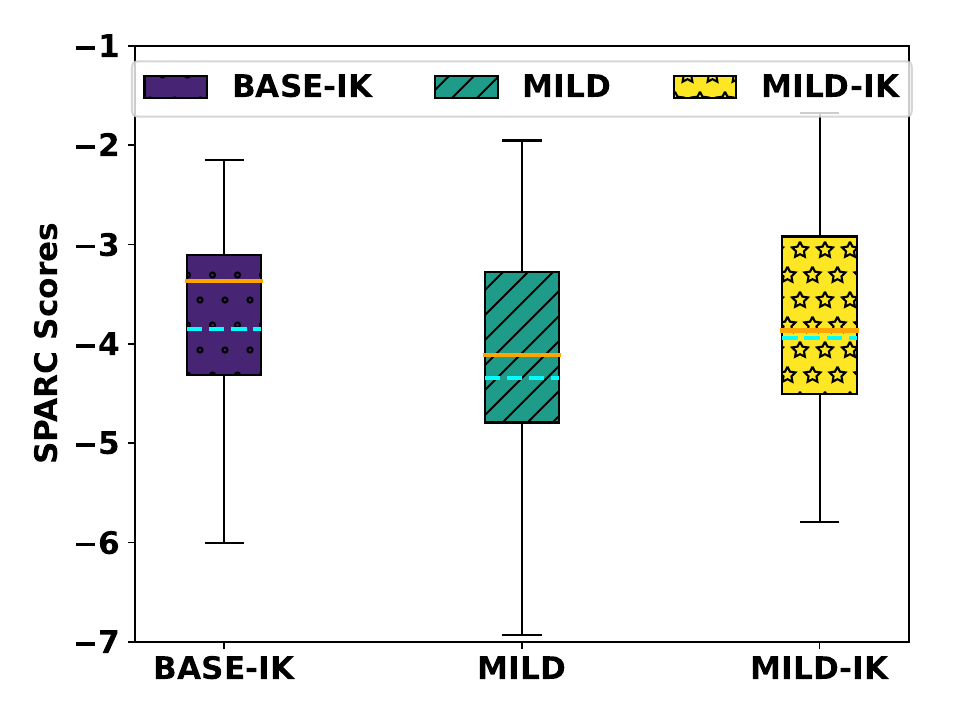}
    \caption{Distribution of SPARC scores across the wrist trajectories of the human partner for different algorithms. The orange line depicts the median and the dotted cyan line depicts the mean.}
    \label{fig:sparc}
    \vspace{-1em}
\end{figure}

We evaluate some objective performance metrics relating to the human partner's motions. We analyze the ergonomics of the arm movements of the participants using the REBA~\cite{hignett2000rapid} and RULA~\cite{mcatamney2004rapid} scores (Figure~\ref{fig:ergonomics}). We additionally analyze the smoothness of the human movements using SPARC~\cite{balasubramanian2015analysis} (Figure~\ref{fig:sparc}).

\input{tabs_n_figs/anova_table}

We found the distribution of the REBA and RULA scores to be very similar across the different algorithms, and within the range of Negligible- to Low-Risk posture scores. Similarly, there were no significant differences in the SPARC scores as well. 
This lack of significance comes mainly from the instructions that participants received for maintaining their posture and arm location, which was needed to minimize occlusions in the skeleton tracking and to comply with the robot's joint limits. Consequently, the deviation in the postures of the participants was negligible across the different trials. Regarding the smoothness of the human partner's motion, given the limited range of motion, in addition to the noise in skeleton tracking, we found no significant differences between the different algorithms.

\begin{figure*}[h!]
\centering
    \includegraphics[width=0.19\textwidth,trim={3.5cm 0 3.5cm 0},clip]{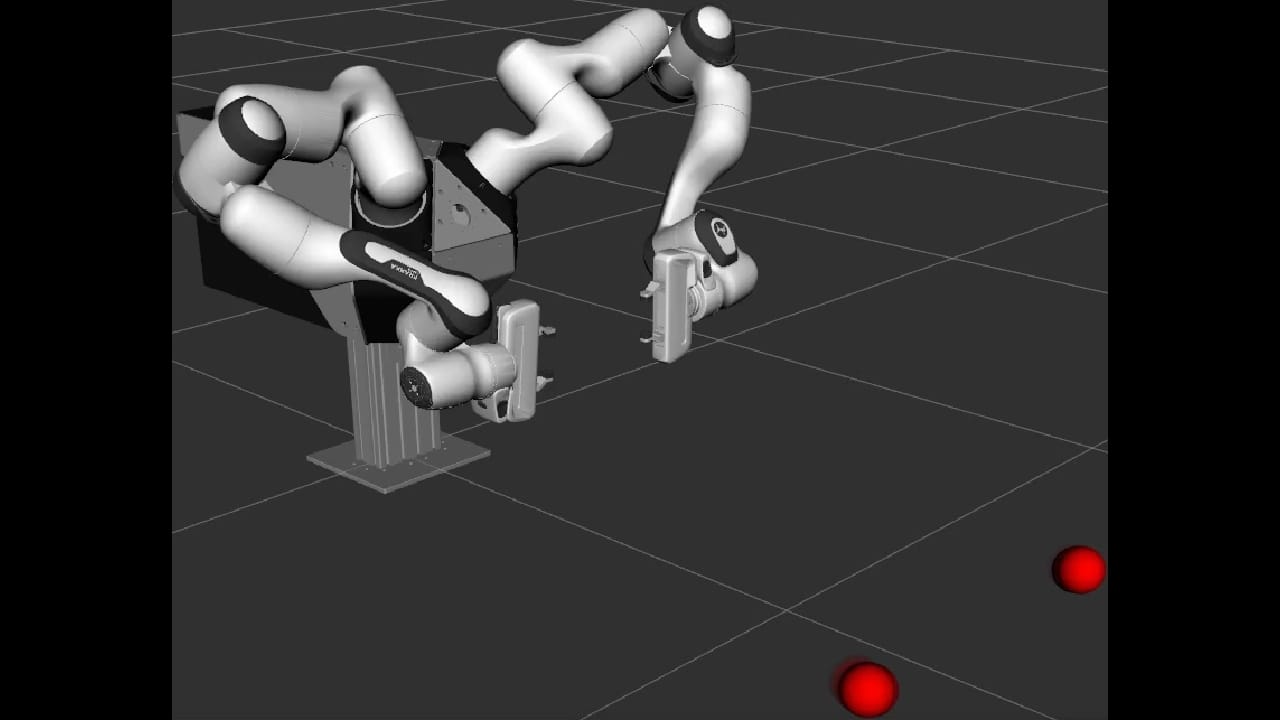}
    \includegraphics[width=0.19\textwidth,trim={3.5cm 0 3.5cm 0},clip]{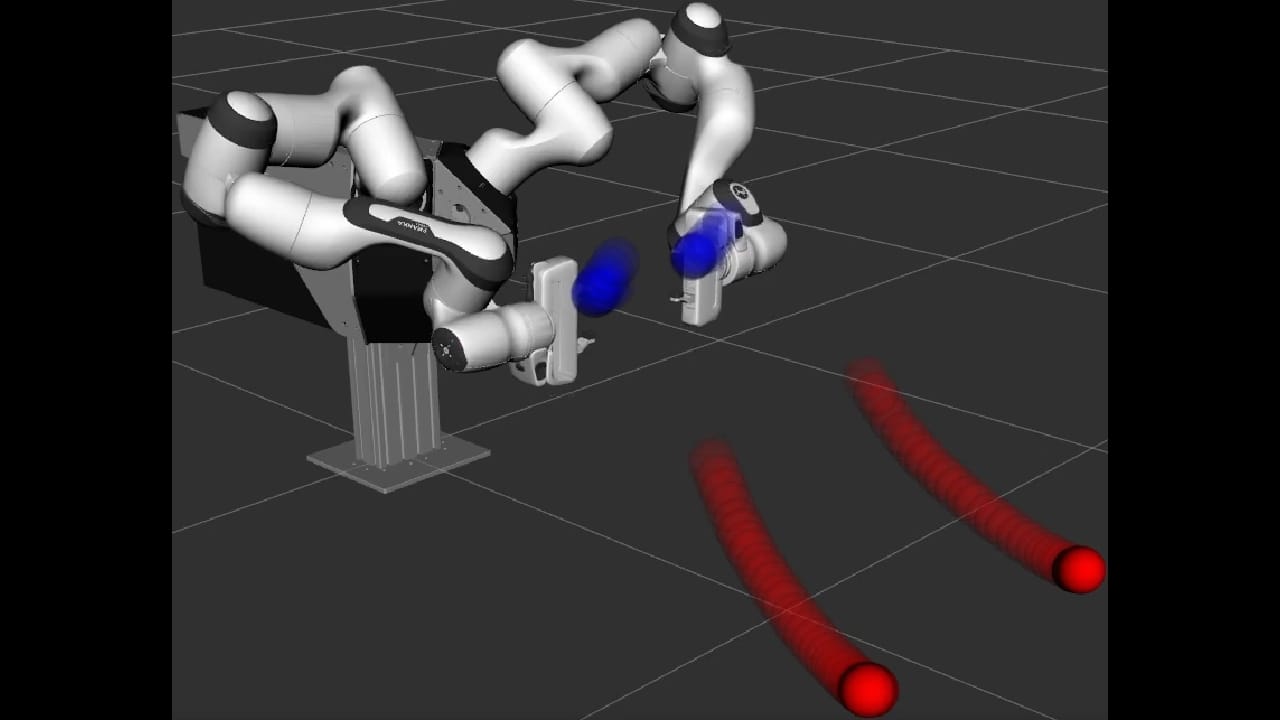}
    \includegraphics[width=0.19\textwidth,trim={3.5cm 0 3.5cm 0},clip]{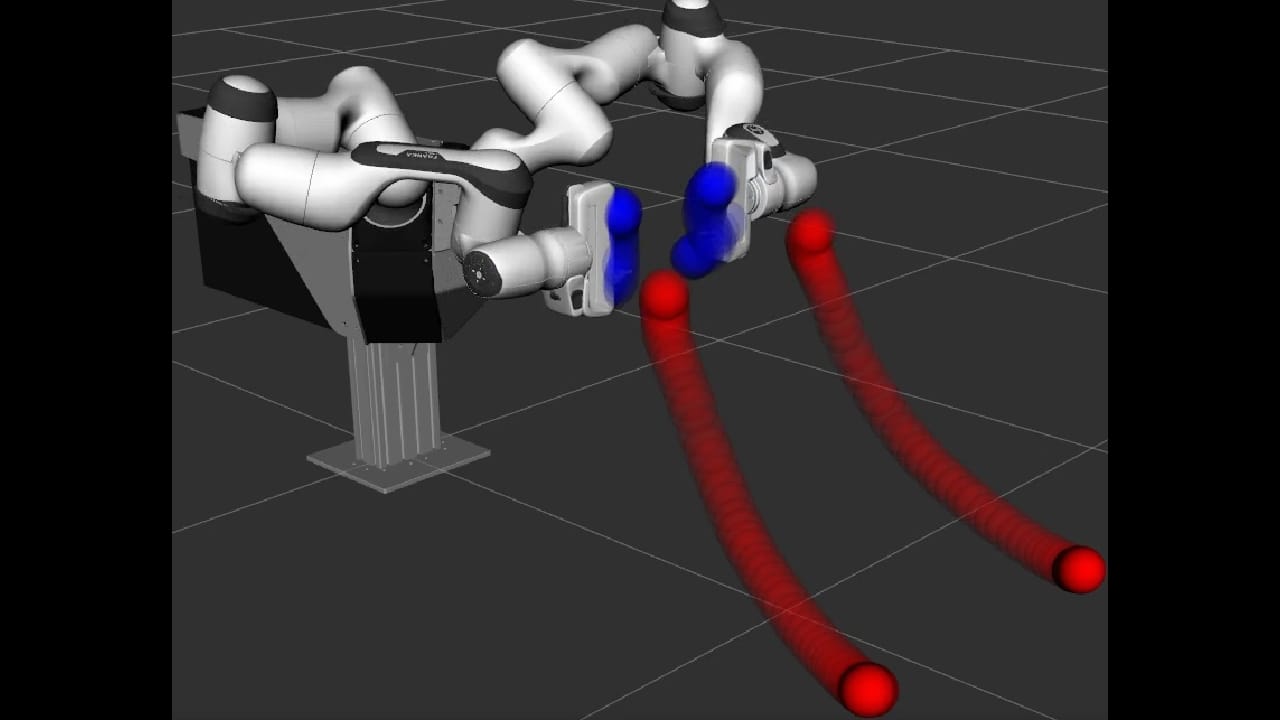}
    \includegraphics[width=0.19\textwidth,trim={3.5cm 0 3.5cm 0},clip]{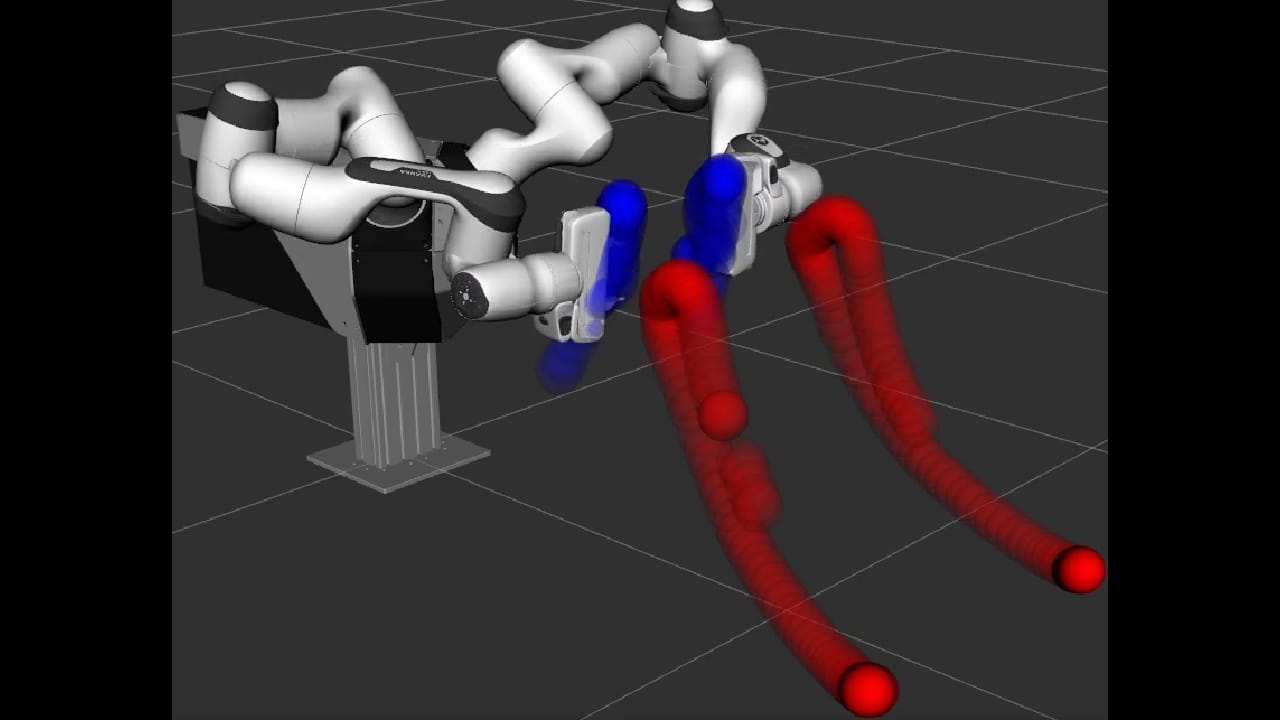}
    \includegraphics[width=0.19\textwidth,trim={3.5cm 0 3.5cm 0},clip]{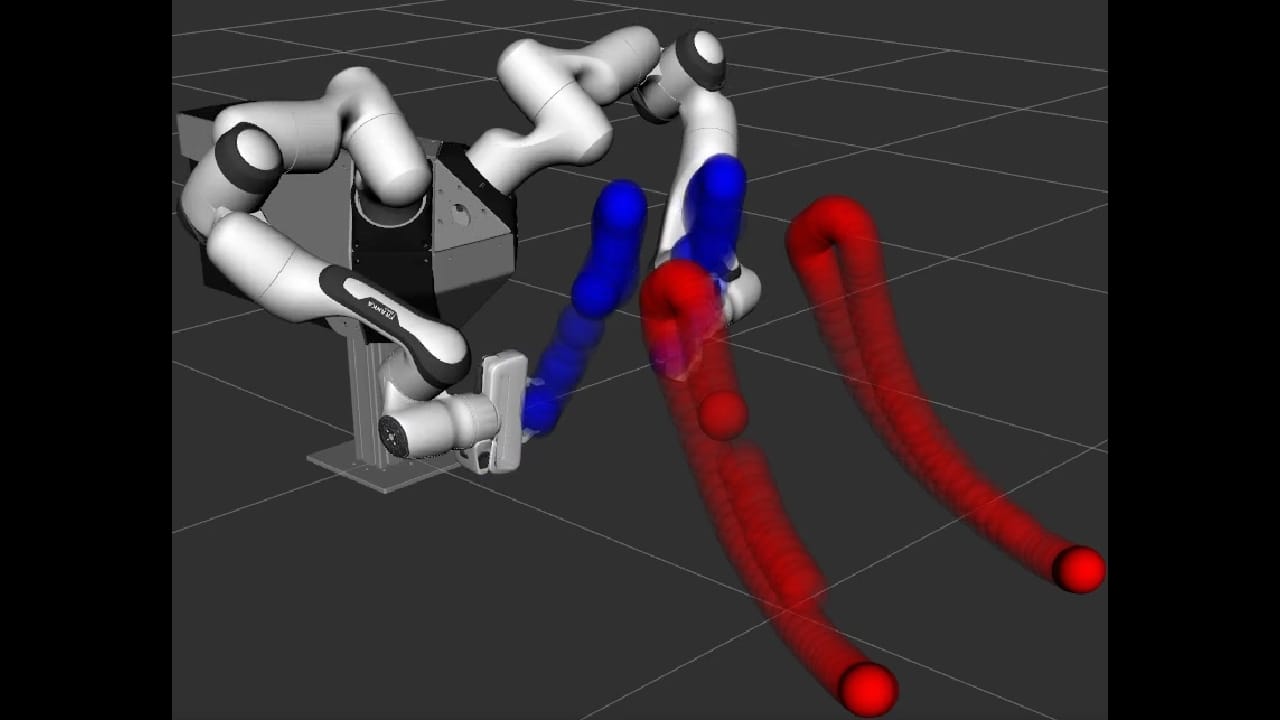}\\
    \includegraphics[width=0.19\textwidth,trim={4cm 0 0 0},clip]{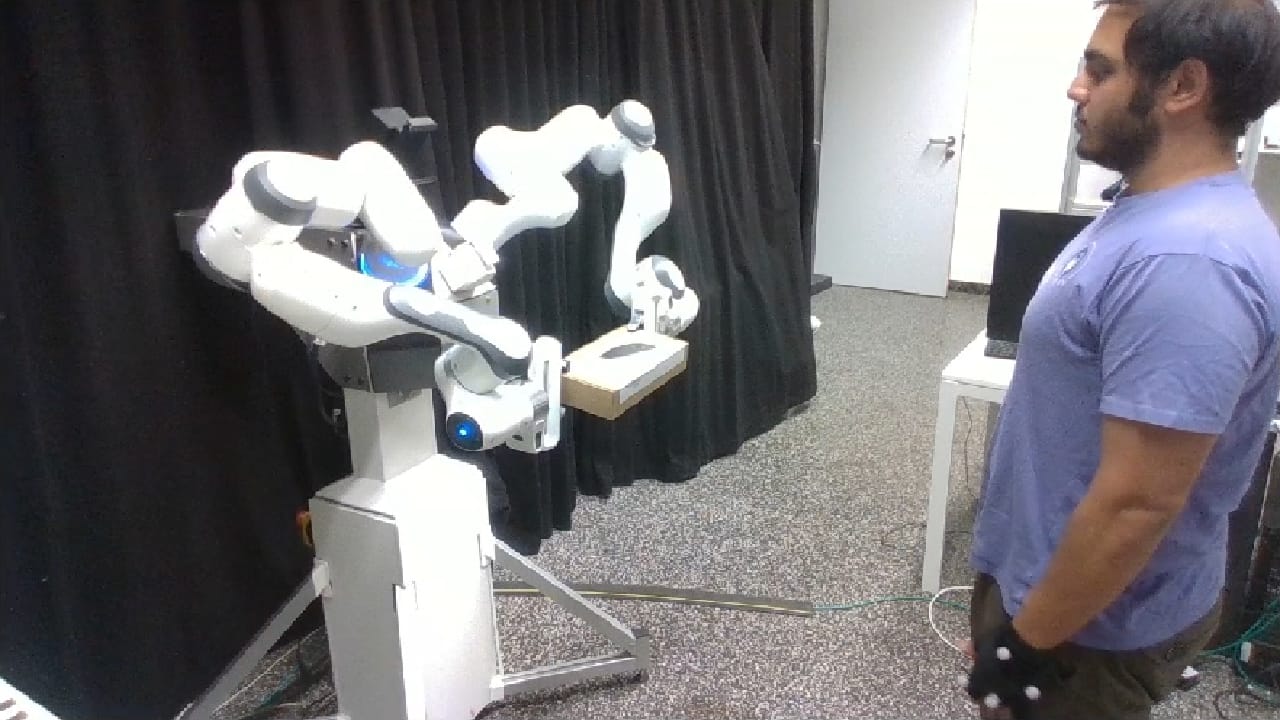}
    \includegraphics[width=0.19\textwidth,trim={4cm 0 0 0},clip]{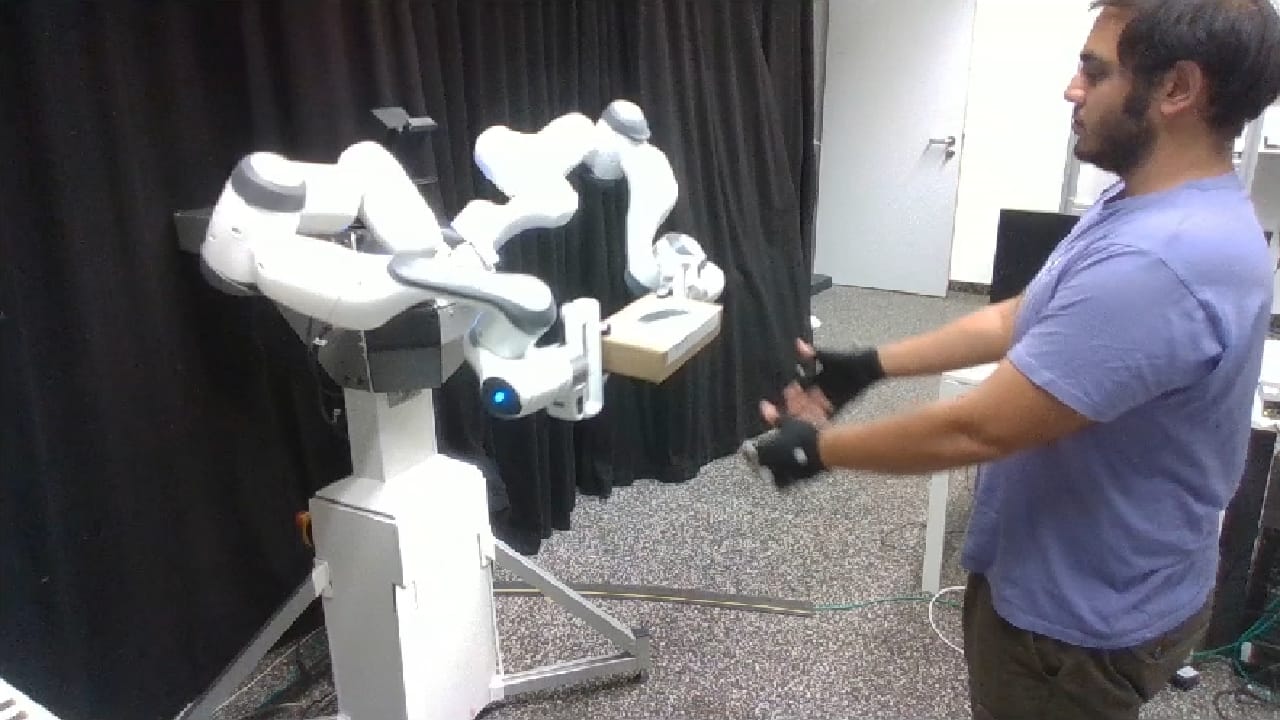}
    \includegraphics[width=0.19\textwidth,trim={4cm 0 0 0},clip]{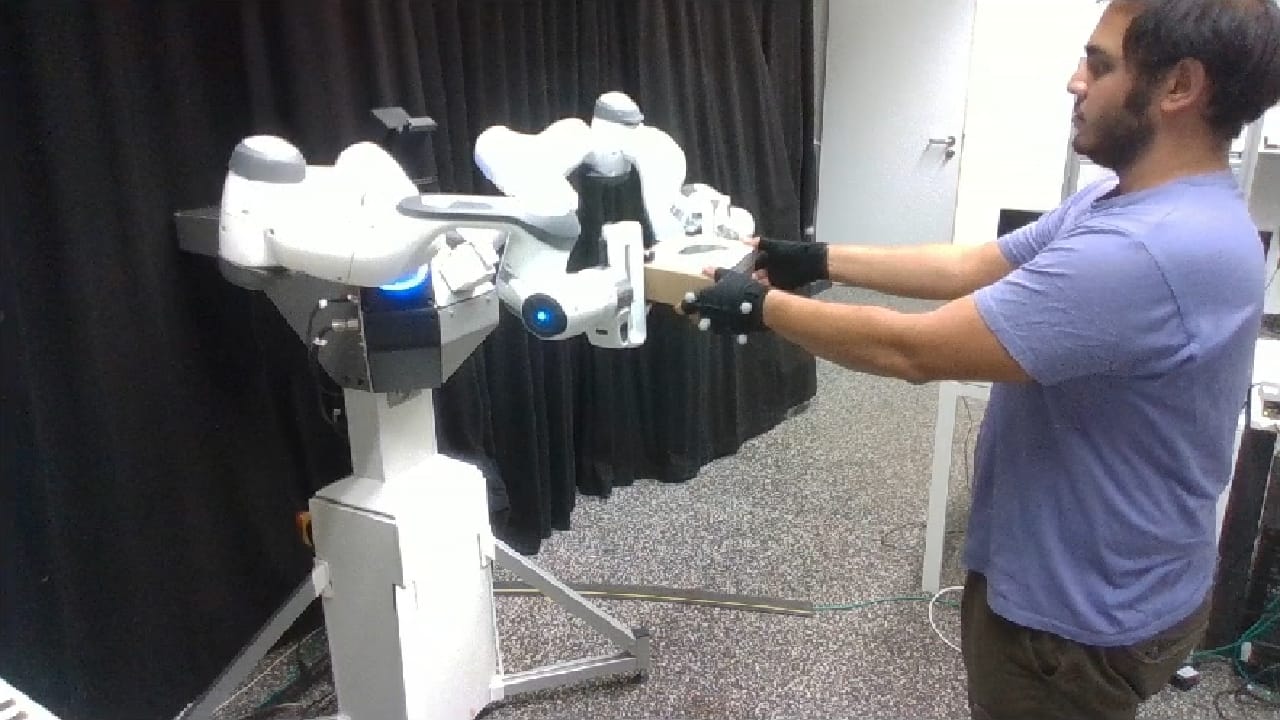}
    \includegraphics[width=0.19\textwidth,trim={4cm 0 0 0},clip]{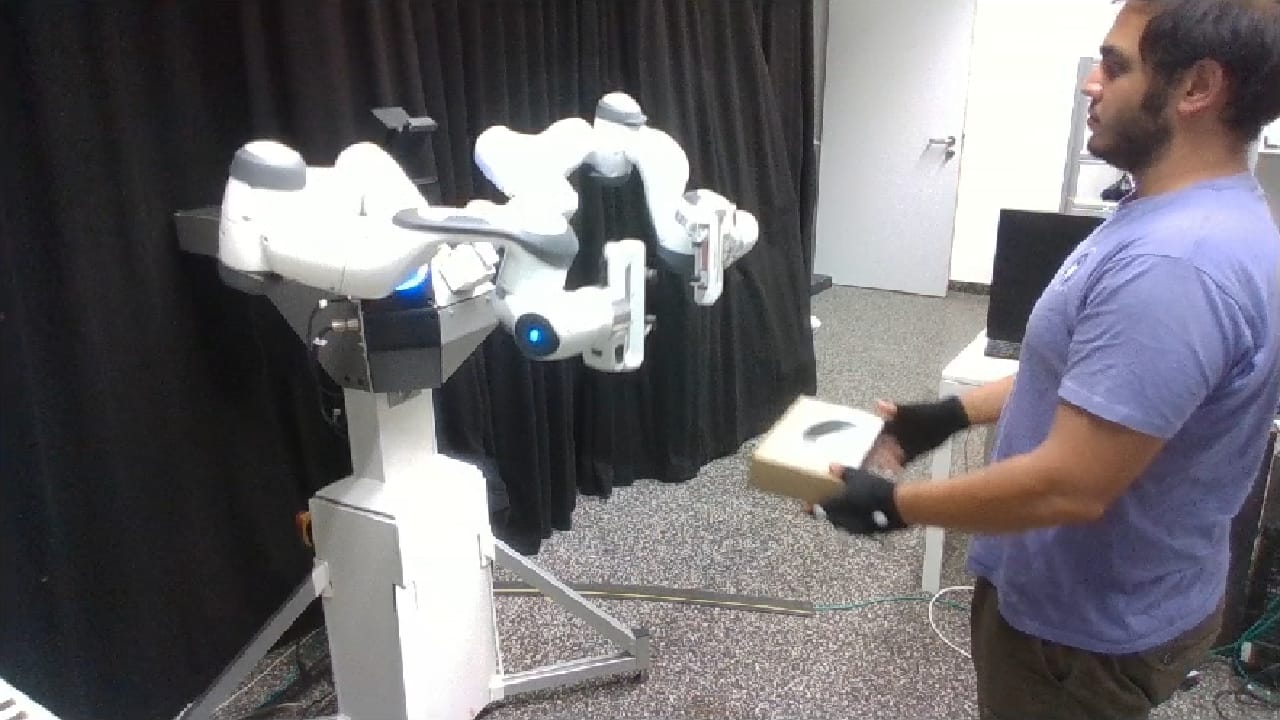}
    \includegraphics[width=0.19\textwidth,trim={4cm 0 0 0},clip]{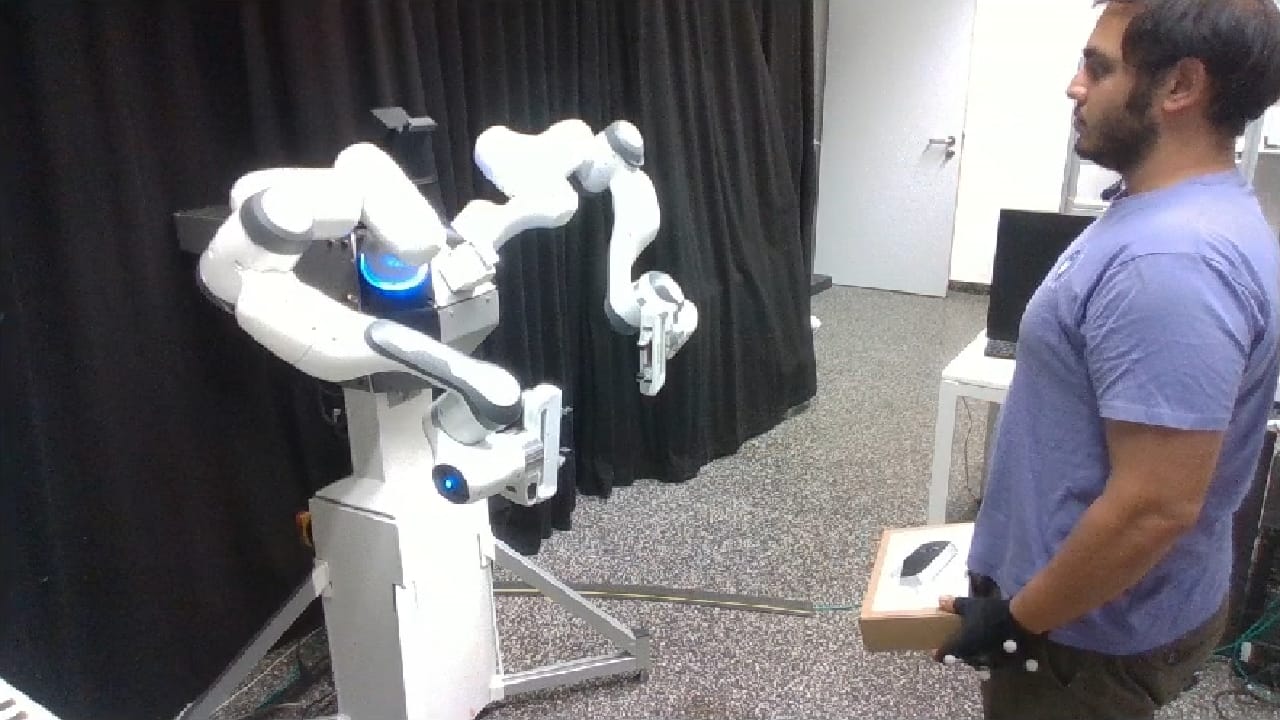}
   
\caption{Example of a Bimanual Robot-to-Human Handover Interaction generated by MILD. The top row shows the observed trajectory of the human in {\color{red} red}, and the generated trajectory of the robot in {\color{blue} blue}. The progression of the interaction can be seen in the bottom row.}
\label{fig:handover}
\vspace{-1.5em}
\end{figure*}

\subsection{Bimanual Robot-to-Human Handovers}
\label{ssec:bihandovers}
To further test the effectiveness of MILD on a more complex task, we showcase the ability of MILD to reactively predict motions for a Bimanual Robot in a Robot-to-Human handover scenario, wherein the robot has to hand over an object to the human partner. For this handover experiment, we use a Bimanual Franka Emika Panda robot setup, \enquote{Kobo} that runs a dual arm cartesian impedance controller to command each arm in the 3D task space. Therefore, we train MILD directly using task space trajectories from the Bimanual Handovers dataset in~\cite{kshirsagar2023dataset}. We re-scale the hand trajectories of the giver in the training data to fit the robot's task space limits and train a model using the hand trajectories of both the giver and the receiver. Since we directly predict task space trajectories from Human-Human Interactions and do not need the inverse kinematics adaptation as done with Pepper, we use MILD~v1 to learn the robot motions as the giver, with the human as the receiver. We find that MILD generates suitable response motions for various objects, an example of which can be seen in Figure~\ref{fig:handover}. However, some failures still occur, mainly from the lack of incorporating object-specific information, such as the object dimensions or geometry. While the failures could be mitigated with some post-processing to optimize the generated motions~\cite{goksu2024kinematically}, this is currently out of the scope of our work. Please refer to the supplementary video for further examples.

\section{Conclusion and Future Work}
\label{sec:conclusion}
In this paper, we proposed a system for learning real-world HRI from demonstrations of Human-Human Interactions (HHI). We first learned latent interaction dynamics from the HHI demonstrations in a modular manner using Hidden Markov Models (HMMs) and then demonstrated how the learned dynamics were utilized for learning HRI behaviors. We further showed how the learning of HRI behaviors was improved by incorporating the conditional distribution of the HMM into the training process. This enabled more accurate reactive motion generation during test time which we found is an important aspect of achieving a competitive performance. We then demonstrated the adaptation of the reactively generated robot trajectories during test time with Inverse Kinematics, thereby successfully combining the spatial accuracy of task-space adaptation with the ease of learning joint-space trajectories in a manner that improved the exactness of the learned behaviors. For a contact-rich task like handshaking, we additionally showed how the HMM segment predictions can be used for stiffness modulation to improve the overall perceived quality of the interaction. Through a user study, we found that our method is better perceived by human users in terms of human-likeness, timing, accuracy, and effortlessness. Our user study further validates the effectiveness of our approach in generalizing well to multiple users despite being trained on data from just two interaction partners. 
We additionally show the effectiveness of MILD in successfully generating Bimanual Robot-to-Human Handovers for different objects in a timely and reactive manner, which shows the usefulness of our approach even on more complex interaction scenarios. 
\subsection{Limitations}

While the current approach generated acceptable and accurate response trajectories, there are still some limitations to our approach which we highlight below. 
Currently, while training the VAE and HMM together as in the HHI scenario with the conditional loss as in the HRI scenario, we frequently encountered mode collapse of the HMM hidden states. This is why we resort to freezing the learned HMM and Human VAE for training on the HRI scenarios. 
This could stem from a bad approximation of the forward variable that does not accurately capture the progression of the hidden states but would rather favor just a few or a single state, thereby exacerbating the mode collapse. While this might be solved by propagating gradients through the forward variable, we found that this brings numerical issues arising from the recurrent nature of the forward variable computation, which leads to vanishing/exploding gradients (as in typical RNNs). Mitigating this issue would enable one to incorporate the HMMs into the training process in a more mathematically sound manner.

Although we train the HMMs jointly over the latent spaces of both interacting agents, during testing, we use only the human observations. 
This can lead to inaccuracies in predicting the state of the interaction when solely using human observations as compared to the joint set of observations of both agents (as highlighted in the appendix). Incorporating the current state of the robot and the relative geometry between the human and the robot into the training process could improve the overall predictive performance of the network.

\vspace{-1.1em}
\subsection{Future Work}

On the practical side, coming to the interaction with the robot, we purposely left out auxiliary behaviors such as speech or gaze that make the robot more \enquote{alive} as our focus was to evaluate the different interaction algorithms. These auxiliary behaviors could help improve the overall perceived quality of the interaction. 
The influence of the inherent personality traits also affects the interaction~\cite{chaplin2000handshaking}. Further research into quantifying such influences, for example, based on the mental states of the human partner~\cite{abdulazeem2023human} or with underlying emotions~\cite{ammi2015haptic,stock2022survey}, to adapt the robot actions and personalize the interactions can subsequently provide a more natural interaction and improve the perception of the robot.
For handshaking, we had to incorporate our own mechanism to ensure compliance during the contact-based segments of the interaction. Alternatively, using an underlying Cartesian Impedance Controller can also help ensure compliance during the interaction. While some works have looked into such compliance in more static scenarios~\cite{bolotnikoval2018compliant}, further research is required to adapt this to more contact-rich and dynamic tasks like handshaking for the Pepper robot.

From the learning side, the current bottleneck of our approach, both in terms of training times and in achieving accurate predictions, is the HMM. Currently, we train the HMMs independently from the VAE in a separate step. In this regard, one could look at incorporating this into the variational framework in a more principled manner~\cite{arenz2023a} or training the HMMs and VAEs in a more closely coupled manner by propagating the gradients of the expectation-maximization through the VAE could yield more suitable representations for the task at hand.
Alternatively, we plan to look at going beyond HMMs by using neural variants for incorporating the multimodality, such as Mixture Density Networks~\cite{bishop1994mixture} and additionally handle inputs from multiple sensors, which are important for functional HRI tasks in settings involving shared autonomy with a human. Additionally, incorporating a GAN-like discriminator~\cite{wang2022co} or diffusion-based losses~\cite{ng2023diffusion} could improve performance over a VAE for more complex tasks.

\vspace{-1em}
\section*{Acknowledgments}
This work was supported by the German Research Foundation (DFG) Emmy Noether Programme (CH 2676/1-1), the EU’s Horizon Europe project 
\enquote{ARISE} (Grant no.: 101135959), the German Federal Ministry of Education and Research (BMBF) Projects \enquote{IKIDA} (Grant no.: 01IS20045), \enquote{KompAKI} (Grant no.: 02L19C150), and \enquote{RiG} (Grant no.: 16ME1001), the F\"orderverein f\"ur Marktorientierte Unternehmensf\"uhrung, Marketing und Personal management e.V., the Leap in Time Stiftung. The authors gratefully acknowledge the compute provided by the high-performance computer Lichtenberg II at TU Darmstadt, funded by BMBF and the State of Hesse.
The authors thank Louis Sterker for helping with the experimental setup and Sven Schultze for helping with the NuiSI dataset. The authors thank Snehal Jauhri, S. Phani Teja, and the reviewers for their constructive comments on the paper. We also thank Alap Kshirsagar, Judith B\"utepage, Emmanuel Pignat, Pierre Manceron, and Roberto Calandra 
for open-sourcing their work that helped develop this paper.

\bibliographystyle{IEEEtrans}
\bibliography{references}  

\newpage


\setcounter{section}{1}
\setcounter{subsection}{0}
\setcounter{table}{0}
\renewcommand{\thesection}{\Alph{section}}
\renewcommand{\thesubsection}{\Alph{section}.\arabic{subsection}}
\renewcommand{\thetable}{\Alph{section}.\arabic{table}}

\input{tabs_n_figs/robot_mse_pvalues}

\begin{IEEEbiography}
[{\includegraphics[width=1in]{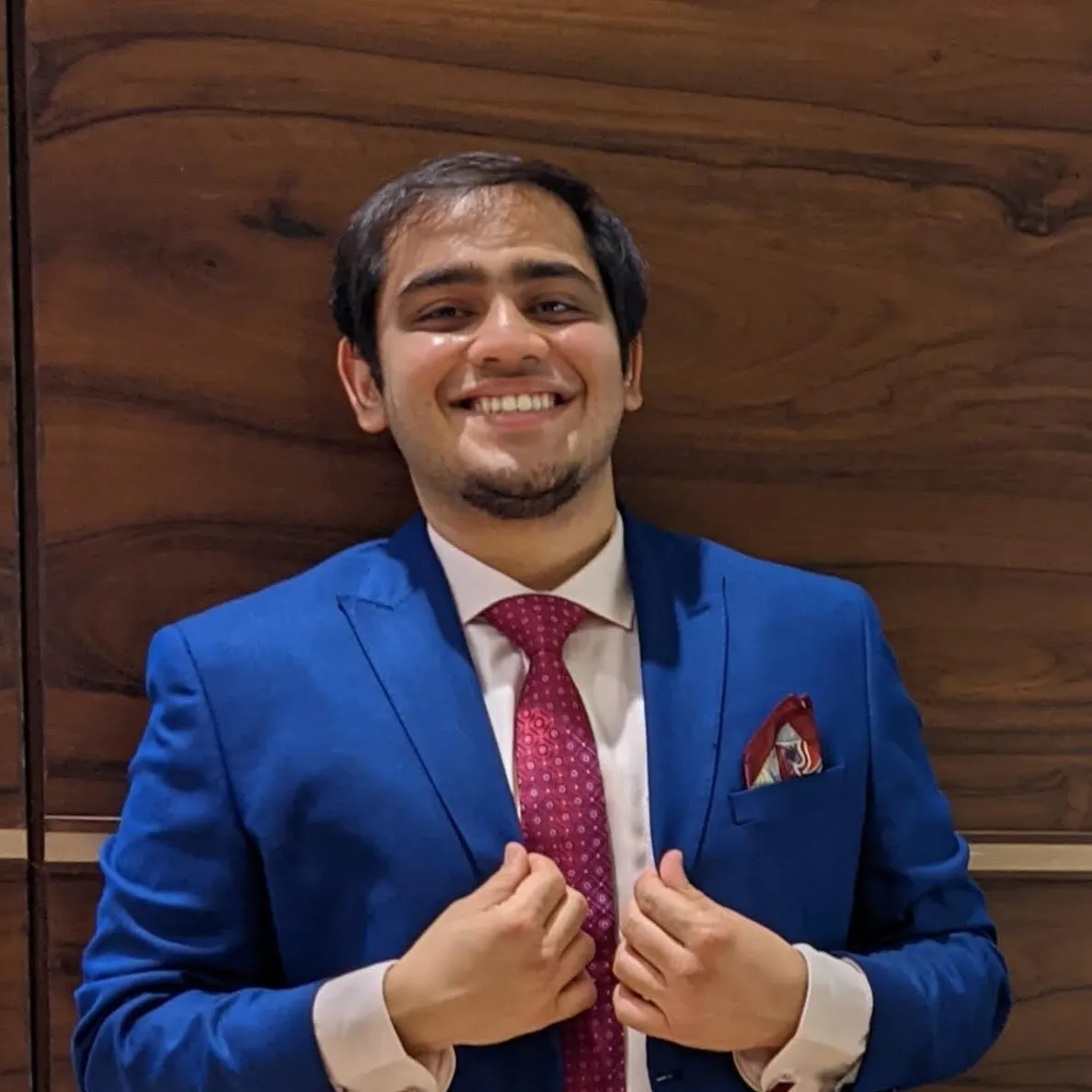}}]{Vignesh Prasad} is a Postdoctoral Researcher at the Interactive Robot Perception and Learning (PEARL) in the Computer Science Department of the Technical University of Darmstadt, Germany, working on on Robot Learning, Computer Vision, and Human-Robot Interaction. He pursued a Master's degree in Computer Science with a focus on robotics at the Robotic Research Center (RRC), IIIT Hyderabad, India. He obtained a Ph.D. from the Intelligent Autonomous Systems Group (IAS) at the Technical University of Darmstadt, Germany, in 2023.
\end{IEEEbiography}
\vspace{-5em}
\begin{IEEEbiography}
[{\includegraphics[width=1in]{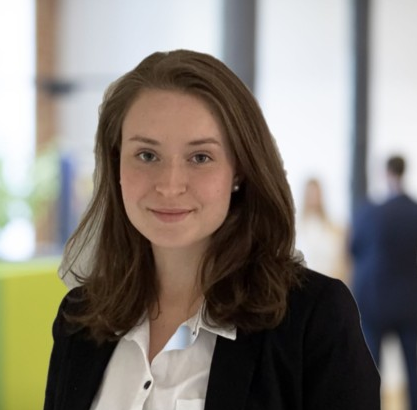}}]{Lea Heitlinger} received her Ph.D. at the Chair of Marketing and Human Resource
Management at the Technical University of Darmstadt, Germany, in 2024. Her research focuses on the psychological aspects of the integration of service robots in our everyday lives from both customer and employee perspectives. Prior to her doctoral studies, she received a Bachelor's and Master's degree in Psychology from the University of Heidelberg, Germany, with a focus on Organizational Behavior \& Adaptive Cognition.
\end{IEEEbiography}
\vspace{-5em}
\begin{IEEEbiography}
[{\includegraphics[width=1in]{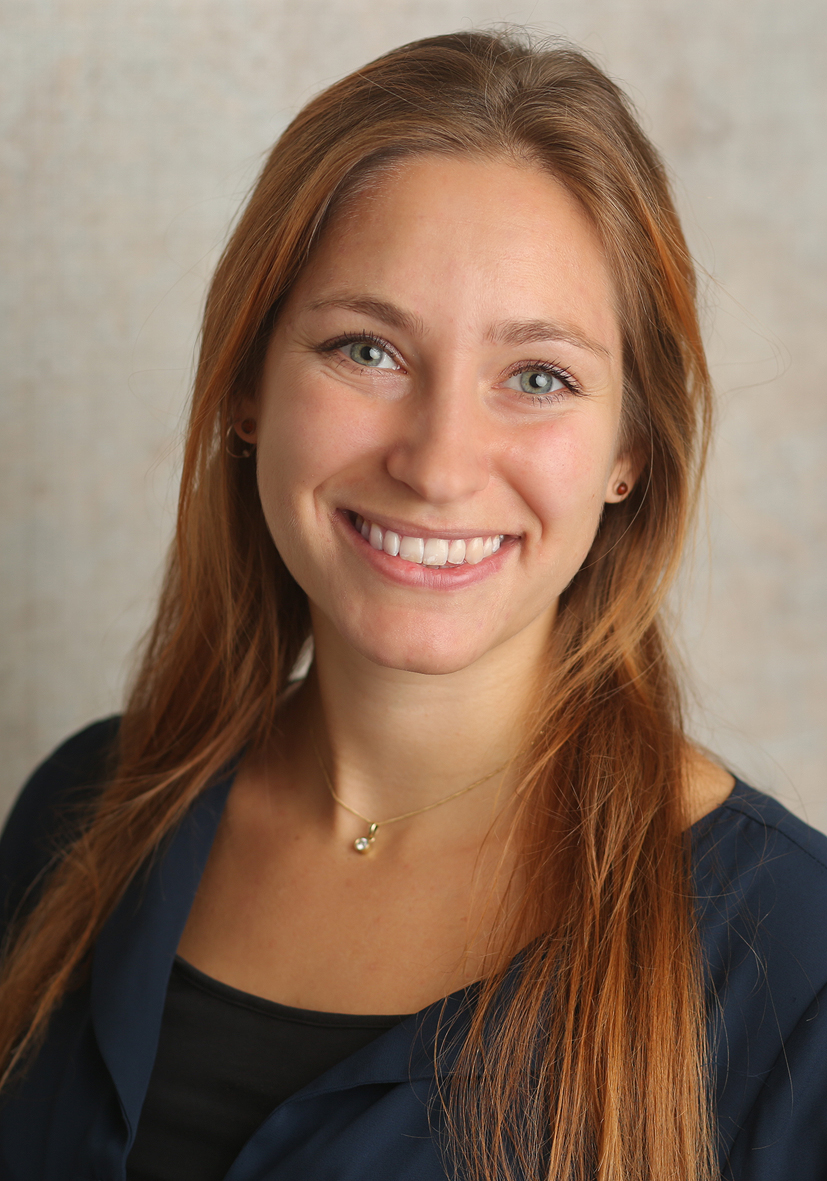}}]{Dorothea Koert} is an Independent Research Group Leader of the interdisciplinary BMBF junior research group IKIDA. She has Master's degrees in Autonomous Systems and Computational Engineering with a focus on Robotics and completed her Ph.D. in 2020 with the Intelligent Autonomous Systems Group (IAS) at the Technical University of Darmstadt. Her research focus is at the intersection of Interactive Machine Learning and Human-Robot Interaction. In 2019 she received the AI-Newcomer award of the German Informatics Society. 
\end{IEEEbiography}

\vspace{-5em}
\begin{IEEEbiography}[{\includegraphics[width=1in]{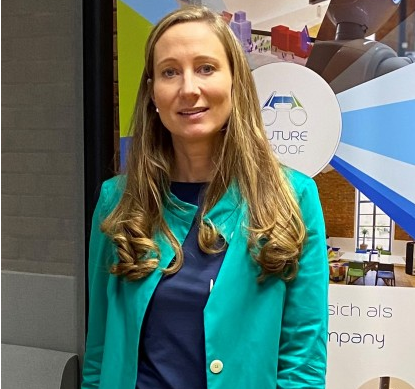}}]{Ruth Stock-Homburg} is a Full (W3) Professor of Marketing and Human Resources Management at the Technical University of Darmstadt. She received a Ph.D. in Management from the University of Mannheim in 2000 and was Germany's youngest Professor in the field of Business Administration at the University of Hohenheim in 2005. In 2018, she completed a second Ph.D. in Psychology from the University of Hagen.
She was honored as the most productive German female professor in the field of business administration in 2005, 2009, and 2014 and received several best paper awards from the American Marketing Association. 
\end{IEEEbiography}
\vspace{-5em}

\begin{IEEEbiography}
[{\includegraphics[width=1in]{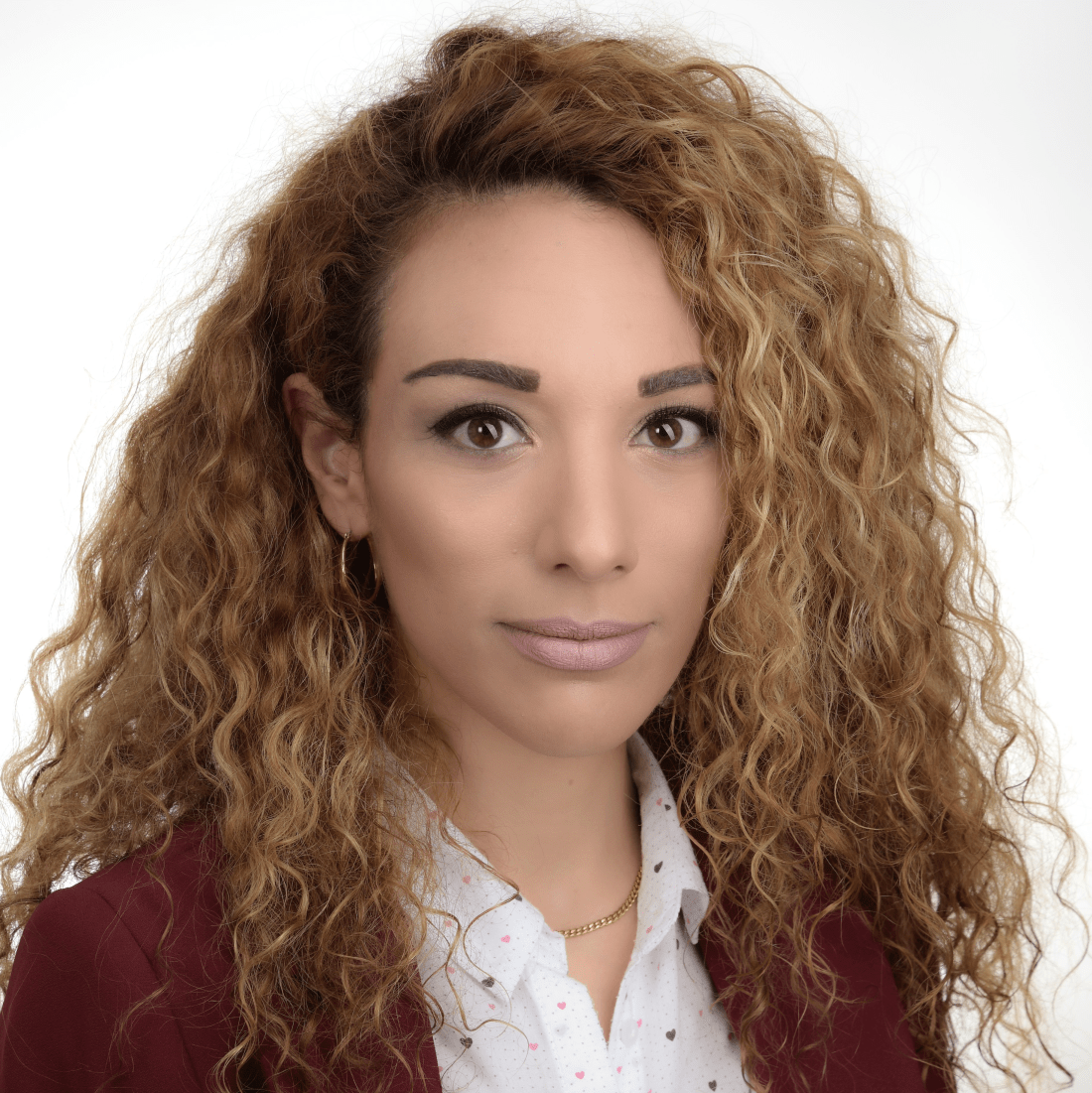}}]{Georgia Chalvatzaki} is a Full Professor (W3) for Interactive Robot Perception and Learning (PEARL) at the Computer Science Department of the Technical University of Darmstadt, Germany. She was awarded an AI Emmy Noether DFG research grant in 2021 and received several awards (IROS 2022 Best Paper Award in Mobile Manipulation,  Outstanding Associate Editor RA-L 2023, Daimler and Benz Foundation Scholarship 2022, 2021 AI Newcomer German Informatics Society, Robotics Science and Systems Pioneer 2020, etc.). She completed her Ph.D. in 2019 at the Electrical and Computer Engineering School of the National Technical University of Athens, Greece.
\end{IEEEbiography}
\vspace{-5em}
\begin{IEEEbiography}
[{\includegraphics[width=1in]{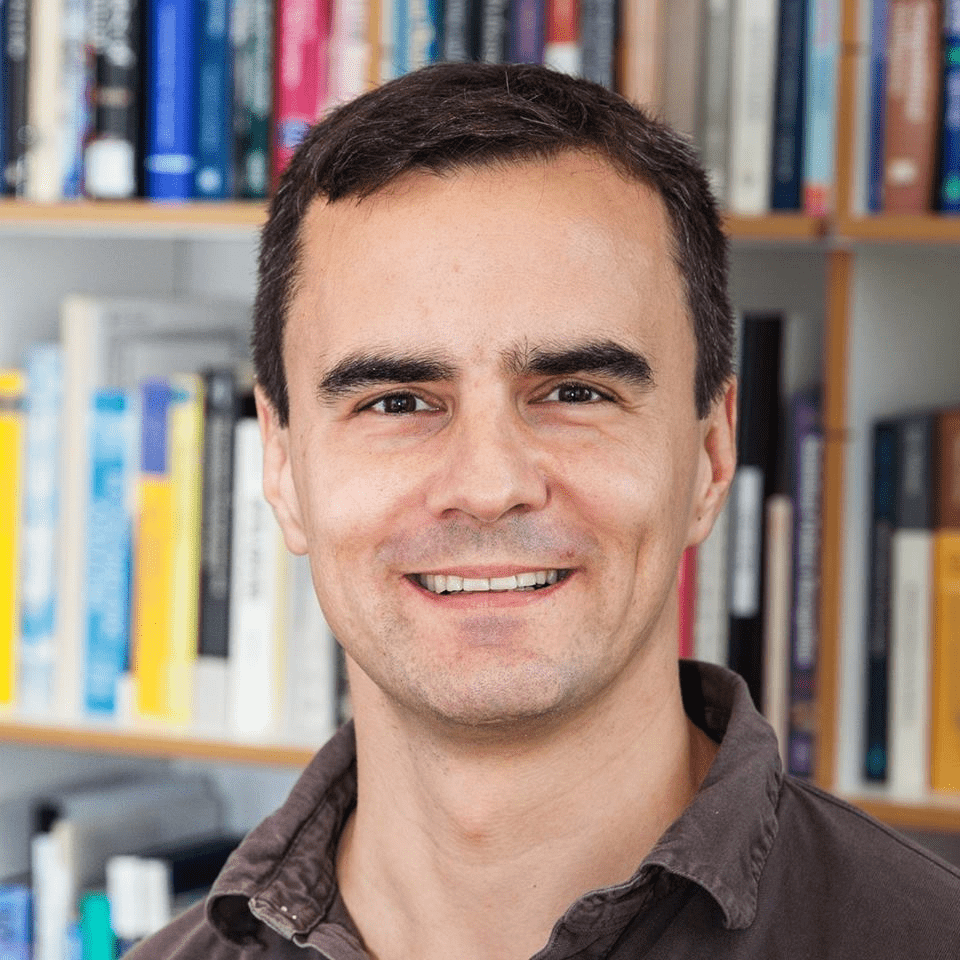}}]{Jan Peters} is a full professor (W3) for Intelligent Autonomous Systems at the Computer Science Department of the Technical University of Darmstadt, dept head of the research department on Systems AI for Robot Learning (SAIROL) at the German Research Center for Artificial Intelligence (Deutsches Forschungszentrum für Künstliche Intelligenz, DFKI), and
a founding research faculty member of The Hessian Center for Artificial Intelligence. He has received the Dick Volz Best 2007 US Ph.D. Thesis Runner-Up Award, RSS - Early Career Spotlight, INNS Young Investigator Award, and IEEE Robotics \& Automation Society’s Early Career Award, as well as numerous best paper awards. He received an ERC Starting Grant and was appointed an IEEE fellow, AIAA fellow and ELLIS fellow.
\end{IEEEbiography}

\end{document}

%% file: tabs_n_figs/hri_images.tex
\begin{figure*}[h!]
\centering
   \centering
     \begin{subfigure}[b]{0.93\textwidth}
         \centering
         \includegraphics[width=0.19\textwidth]{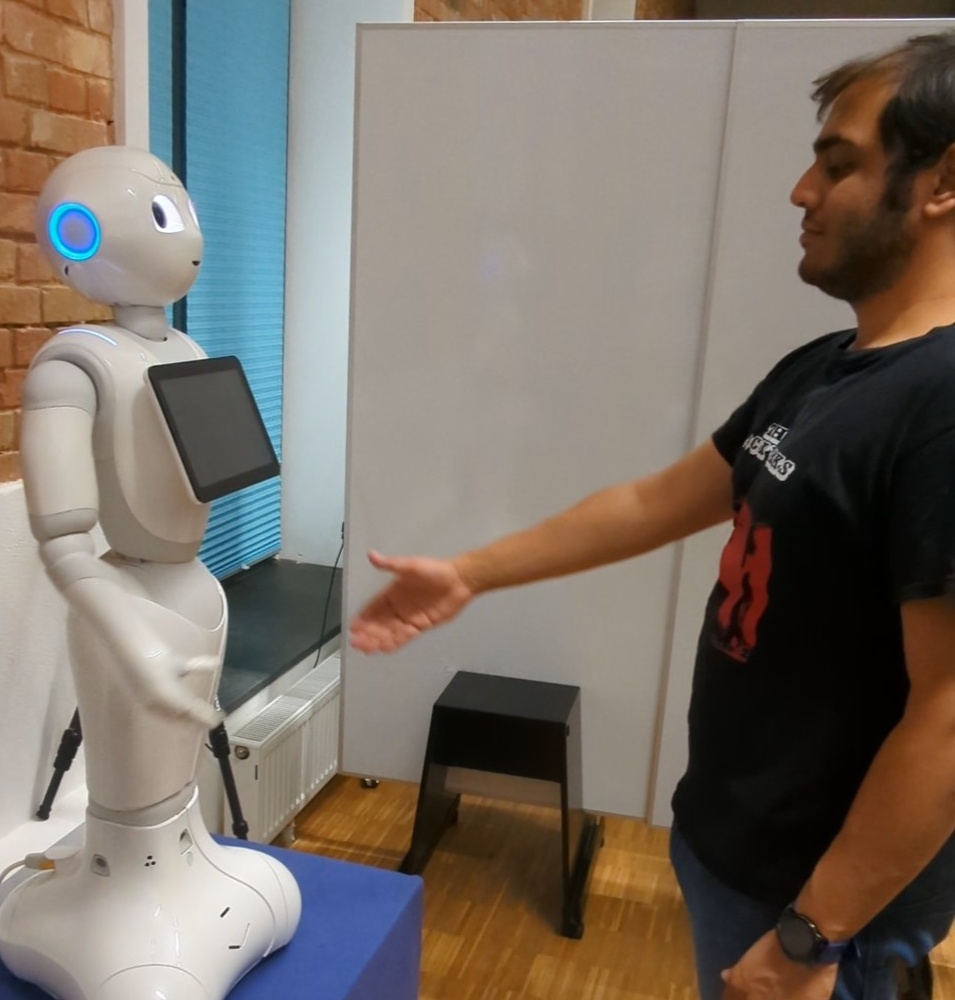} \hfill \includegraphics[width=0.19\textwidth]{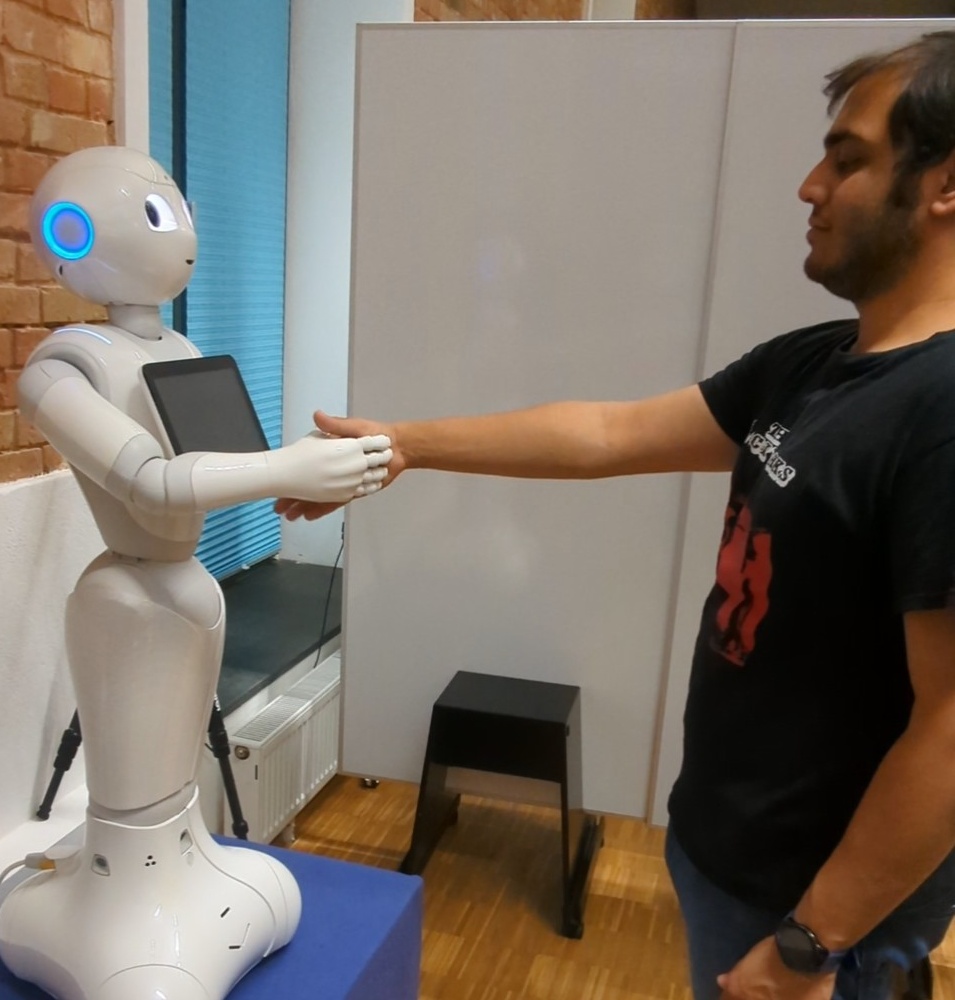} \hfill \includegraphics[width=0.19\textwidth]{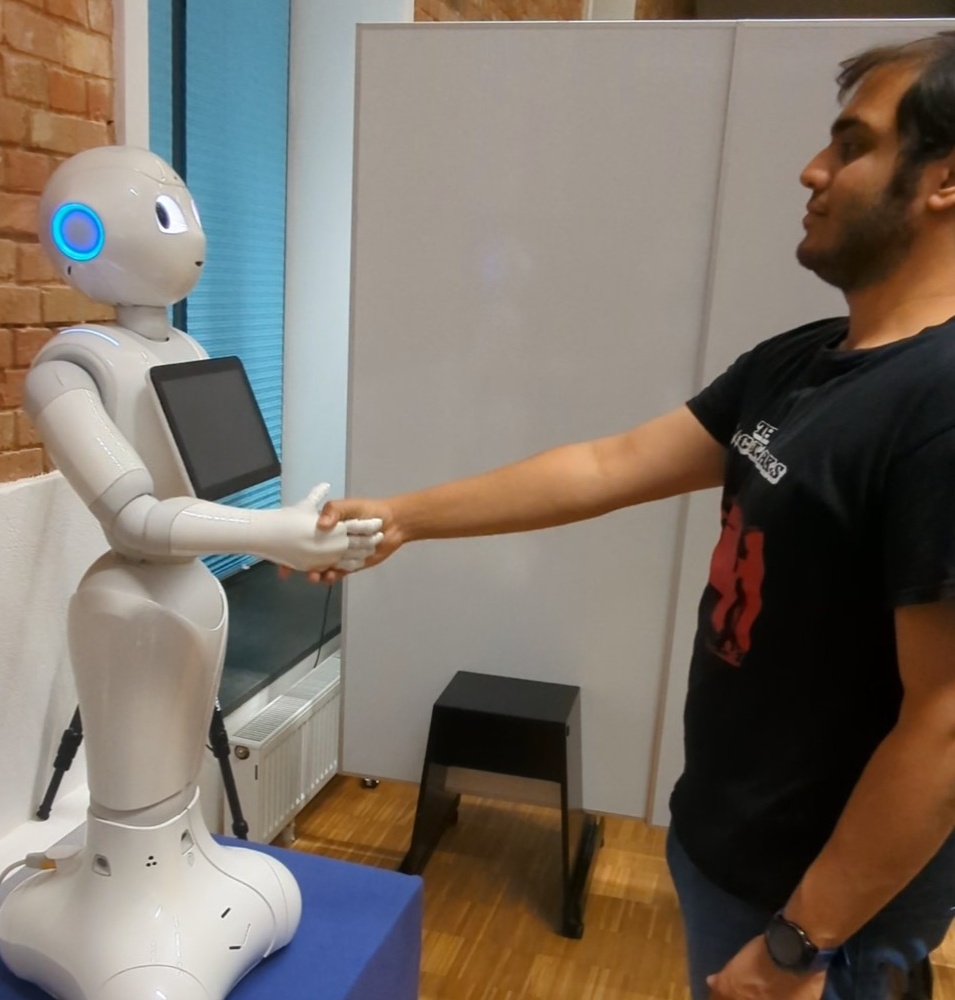} \hfill \includegraphics[width=0.19\textwidth]{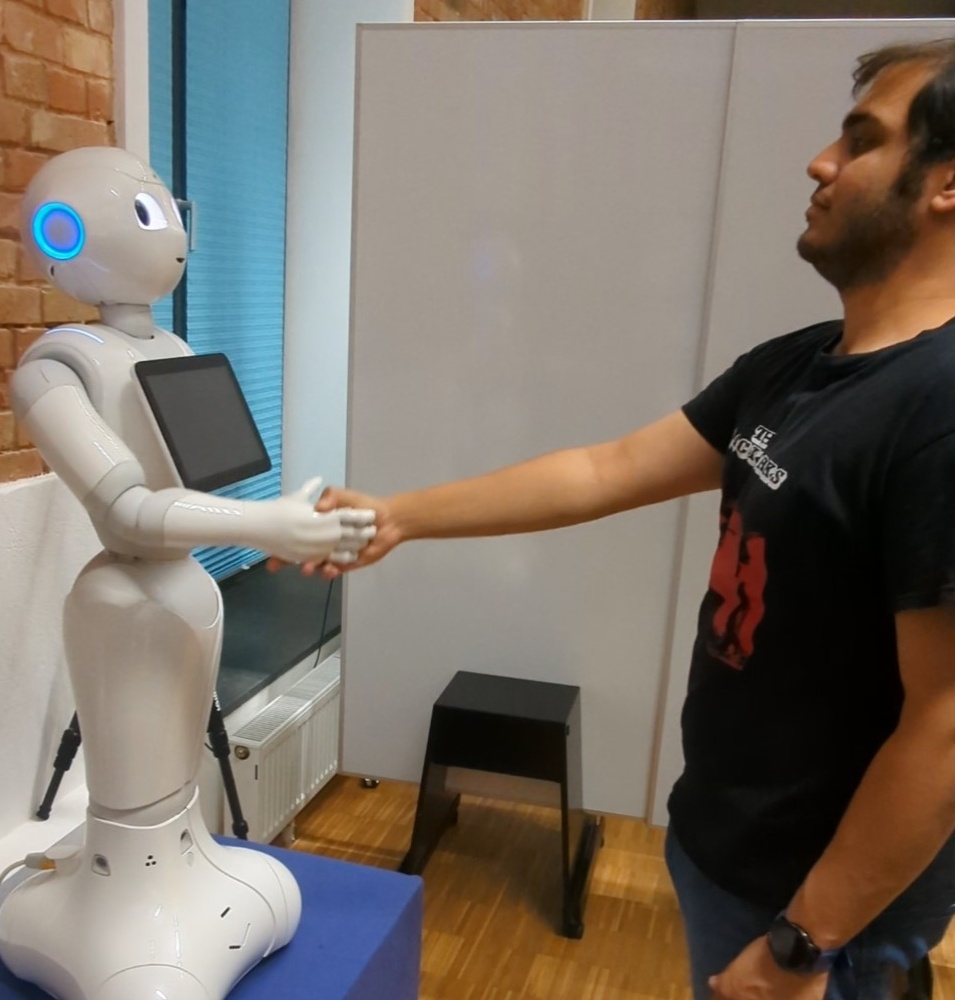} \hfill \includegraphics[width=0.19\textwidth]{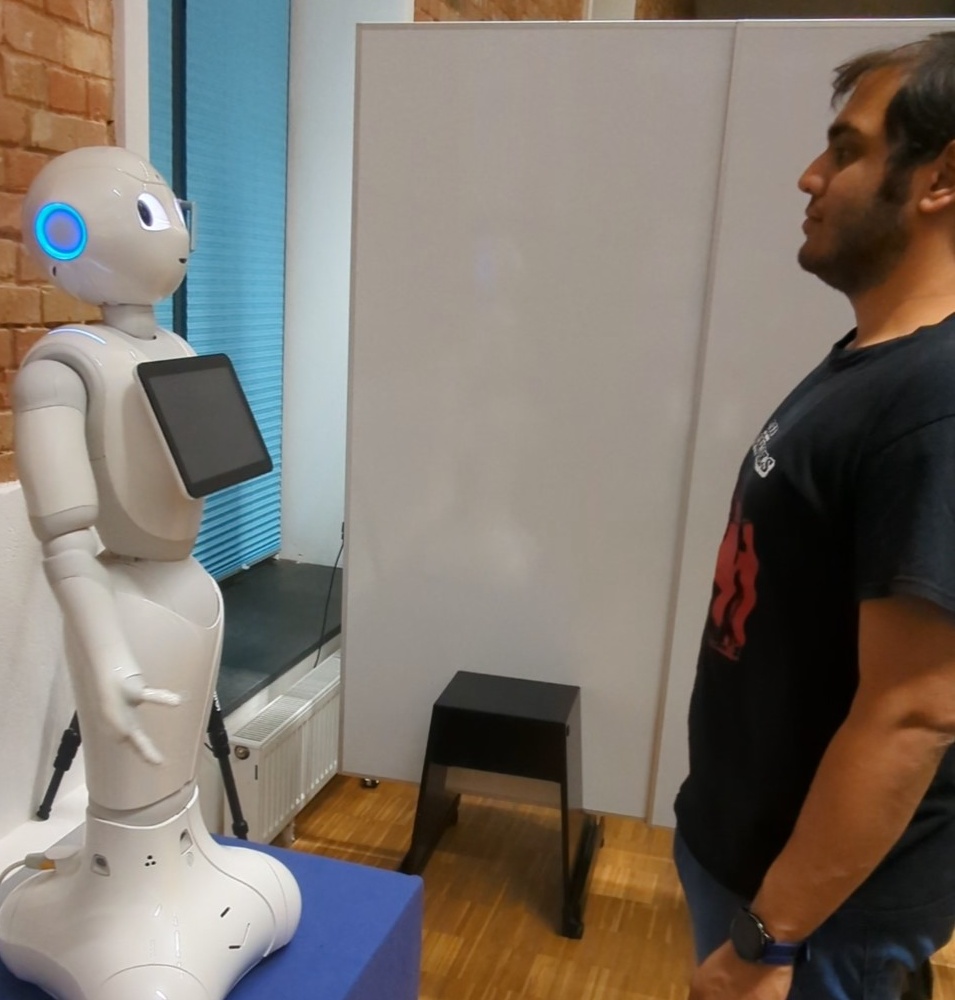} \\
         \includegraphics[width=0.19\textwidth]{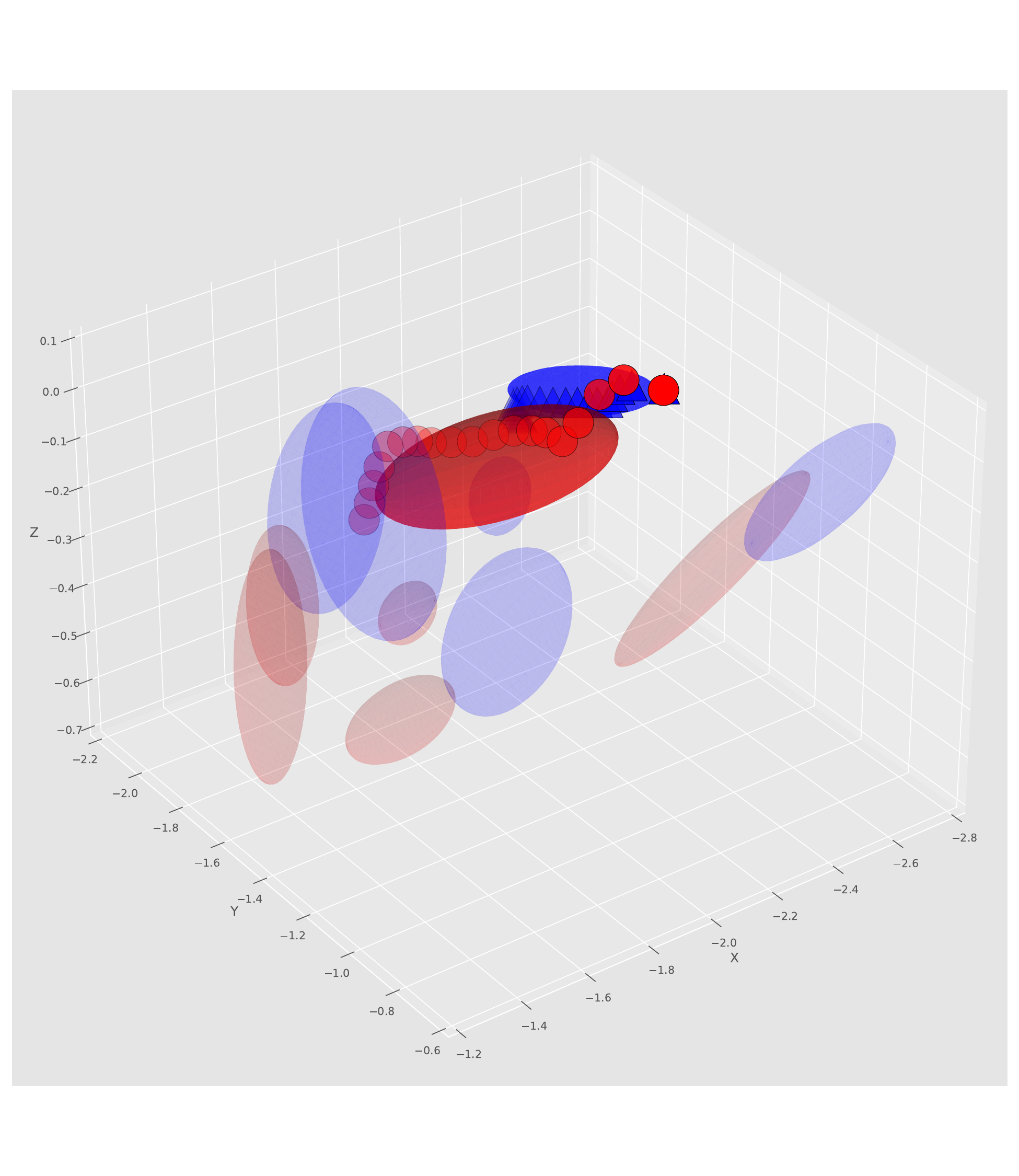} \hfill \includegraphics[width=0.19\textwidth]{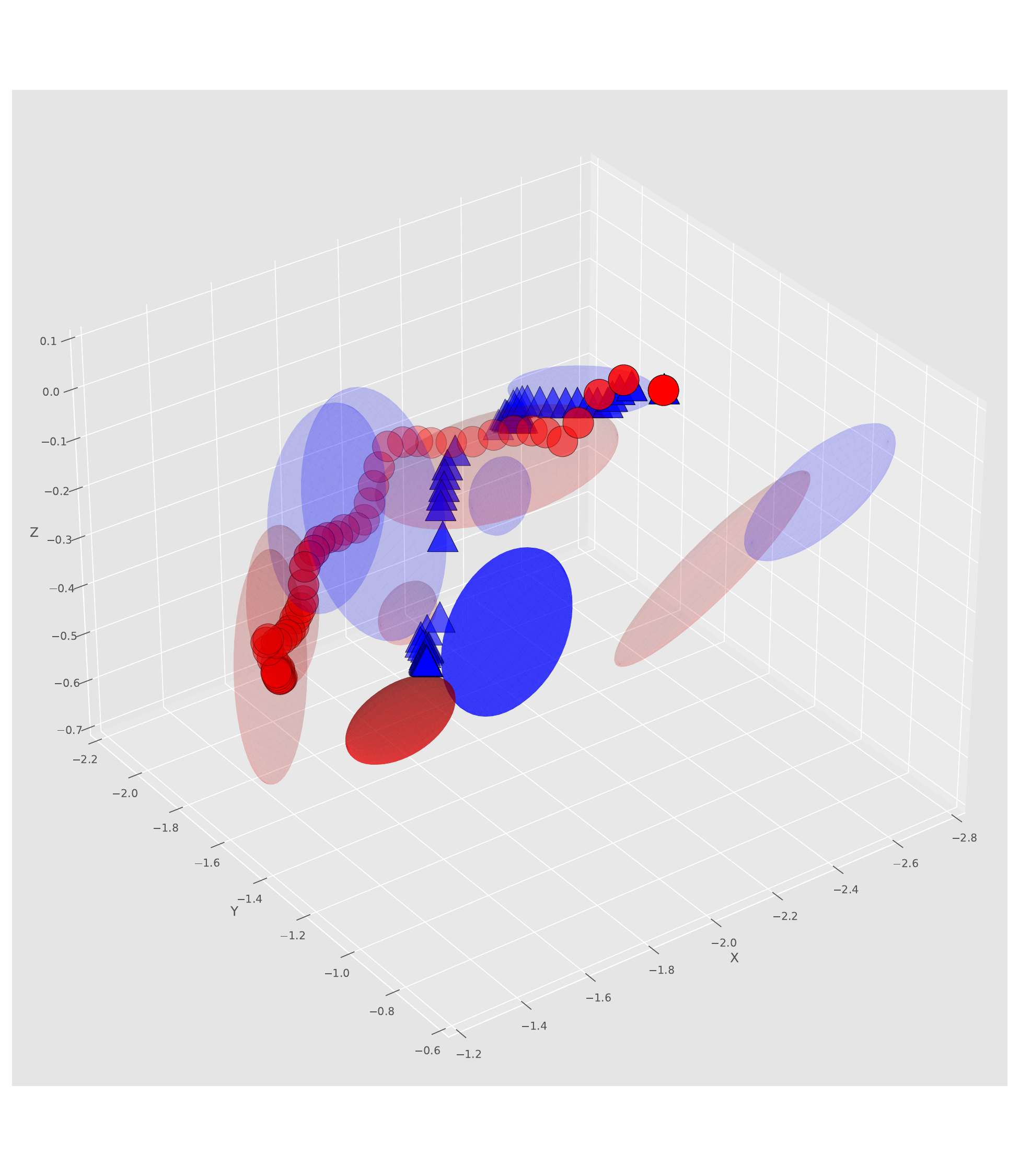} \hfill \includegraphics[width=0.19\textwidth]{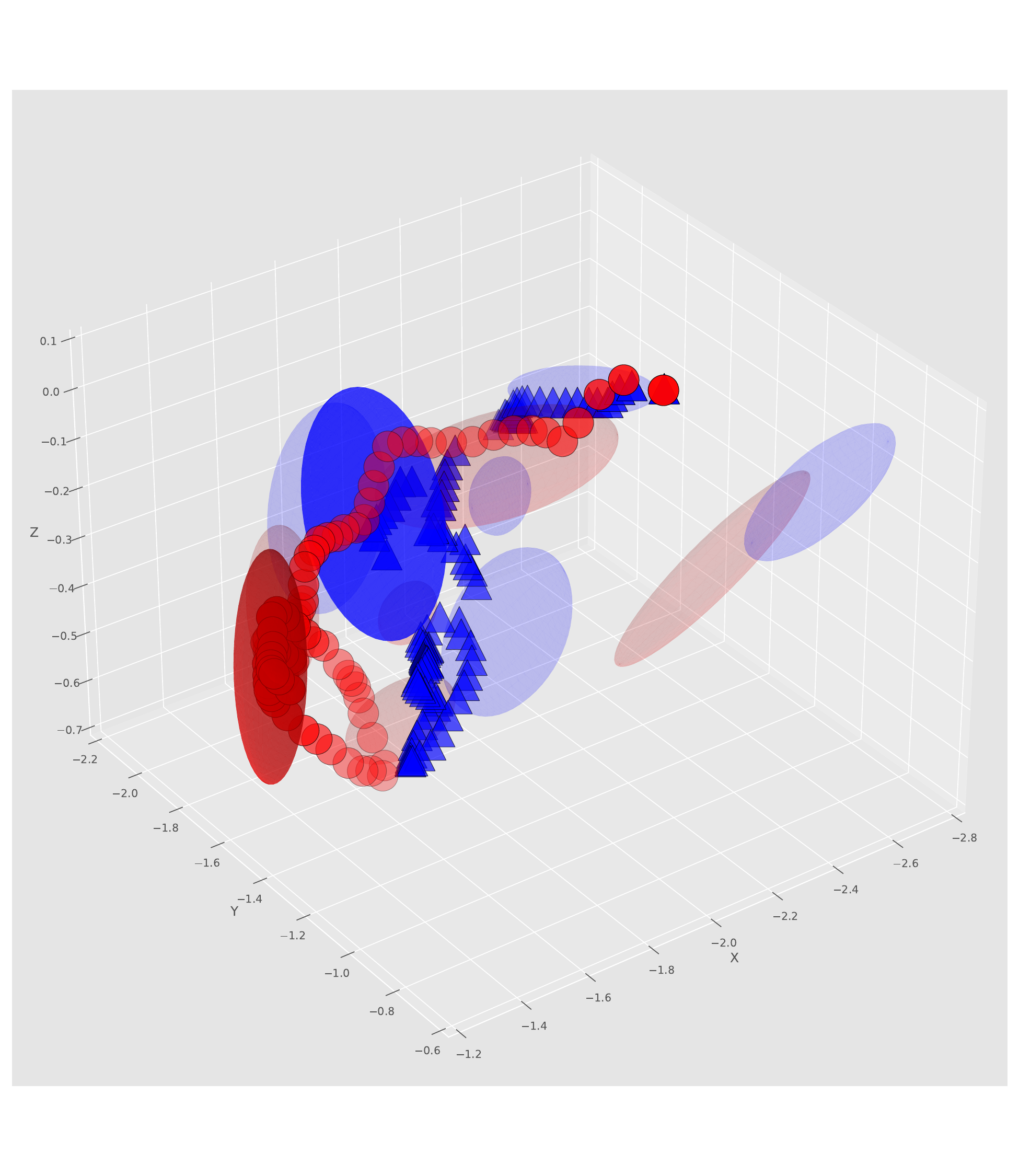} \hfill \includegraphics[width=0.19\textwidth]{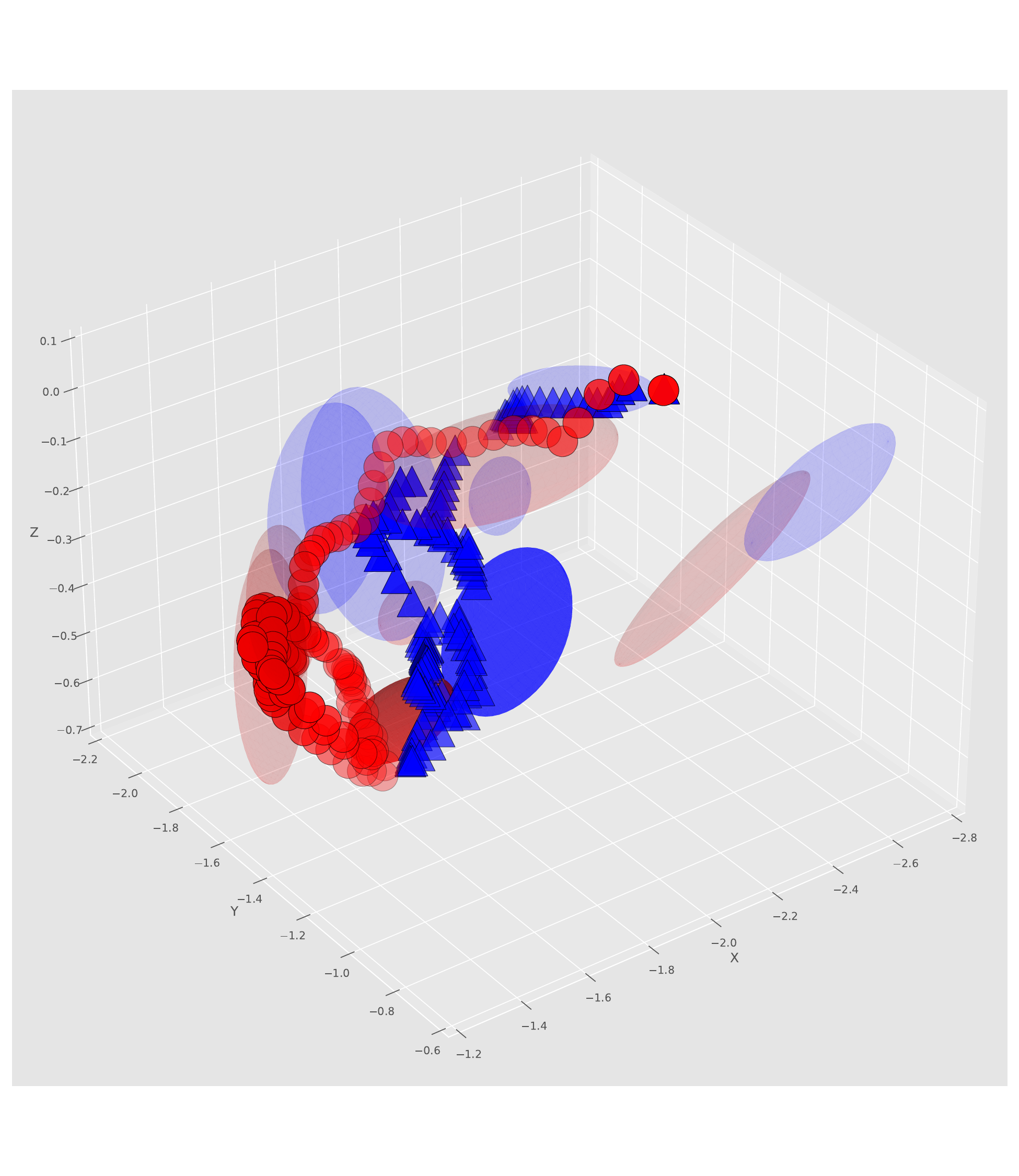} \hfill \includegraphics[width=0.19\textwidth]{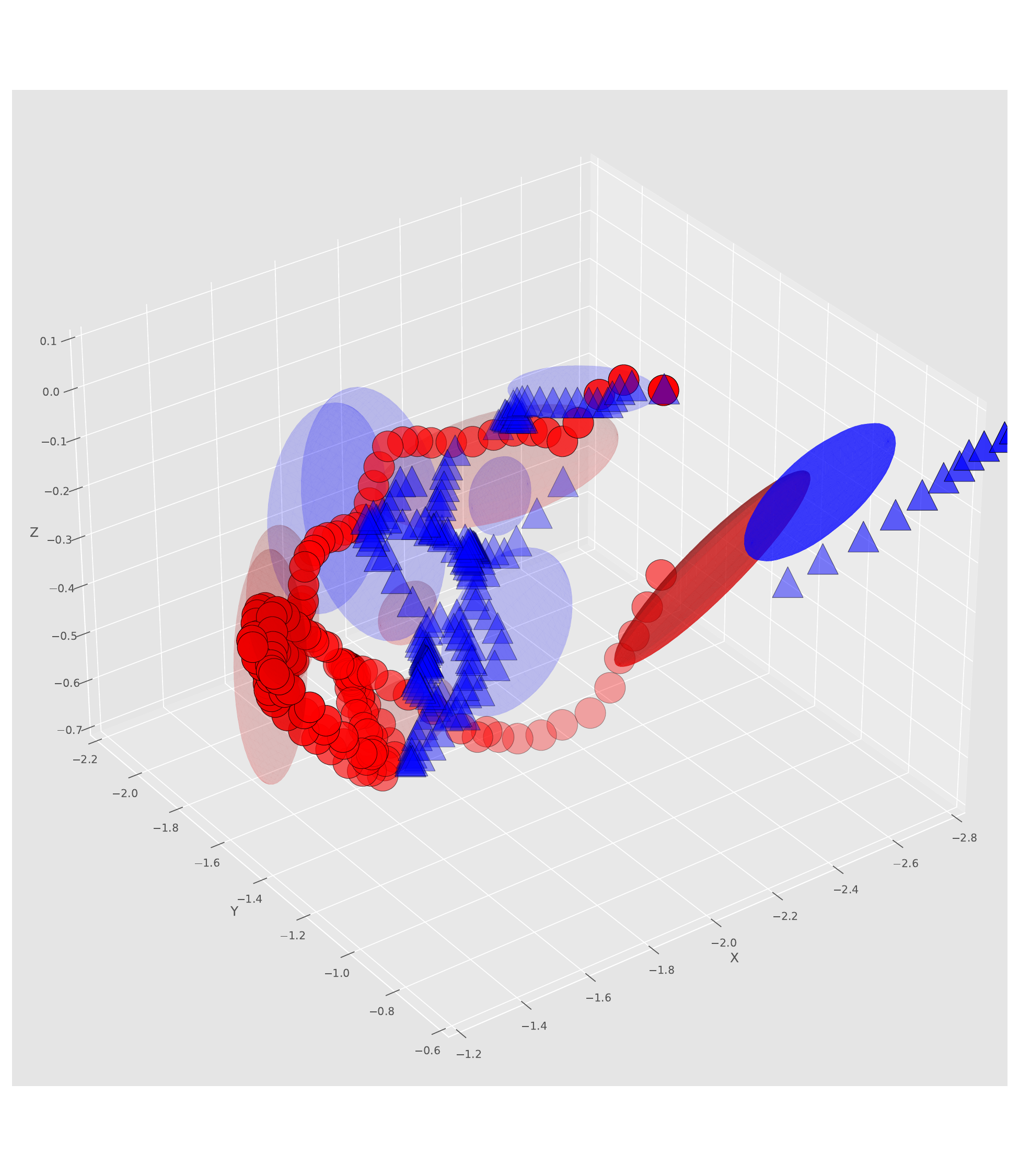} \\ 
         \includegraphics[width=\textwidth]{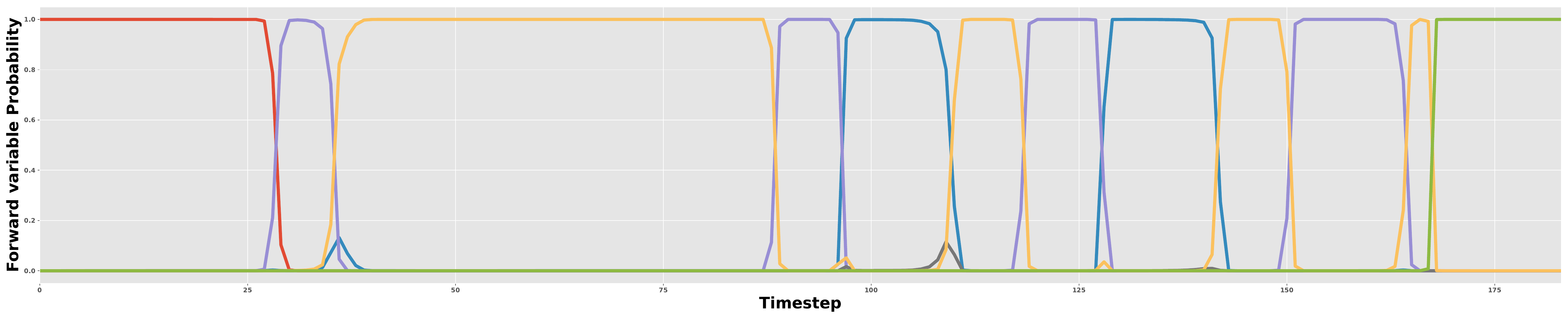}
         \caption{Sample Handshake HRI}
         \label{fig:hri-handshake}
     \end{subfigure}
     \begin{subfigure}[b]{0.93\textwidth}
         \centering
         \includegraphics[width=0.19\textwidth]{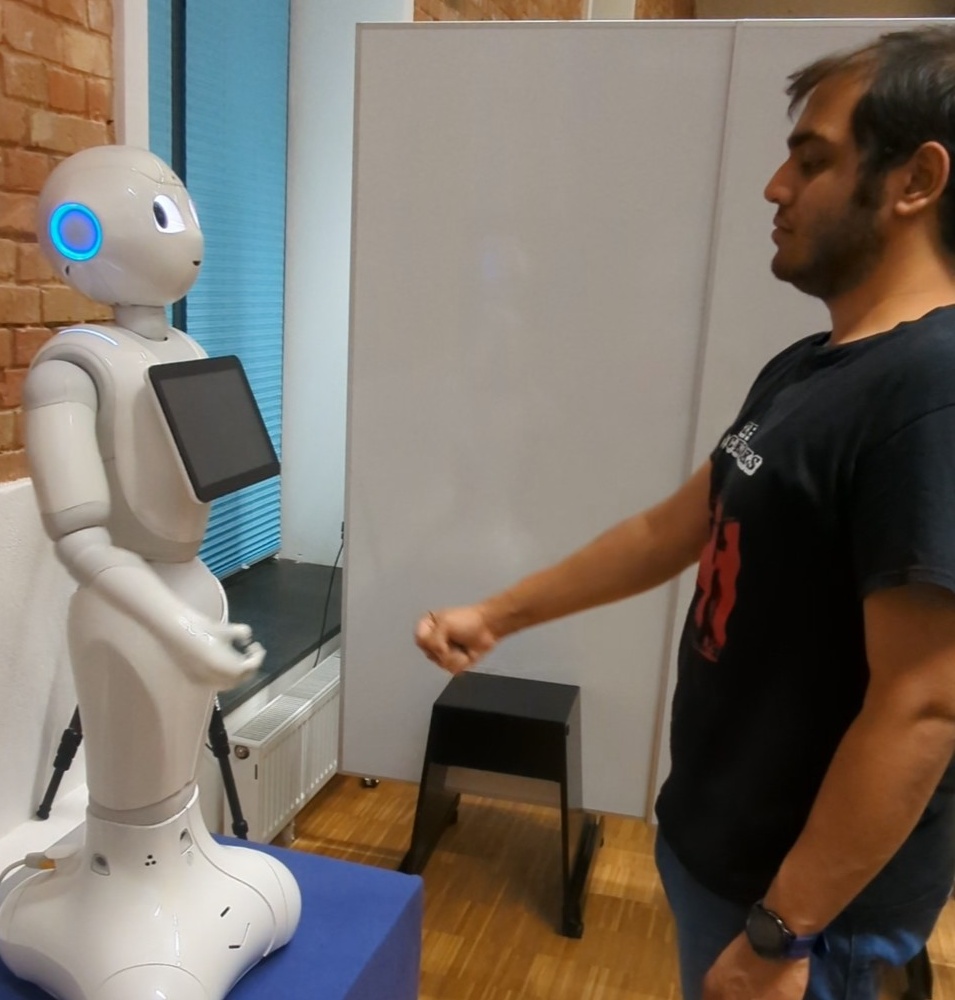} \hfill \includegraphics[width=0.19\textwidth]{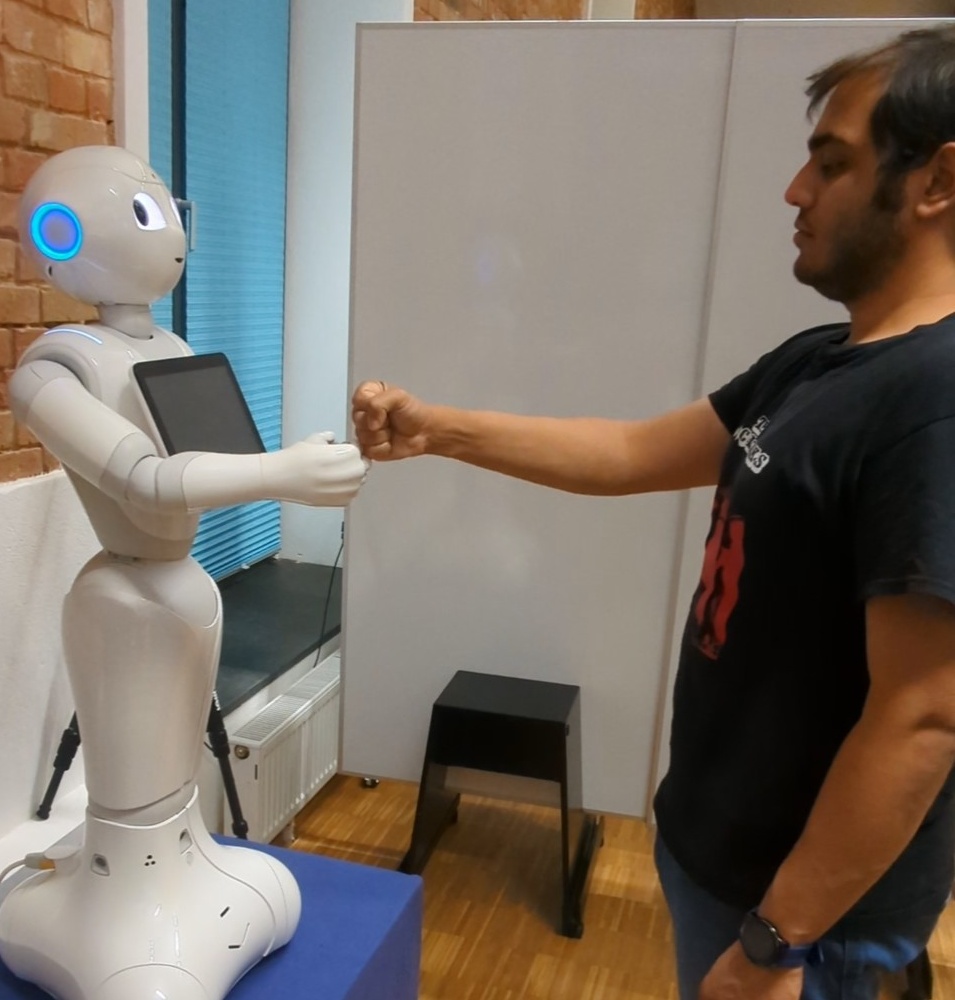} \hfill \includegraphics[width=0.19\textwidth]{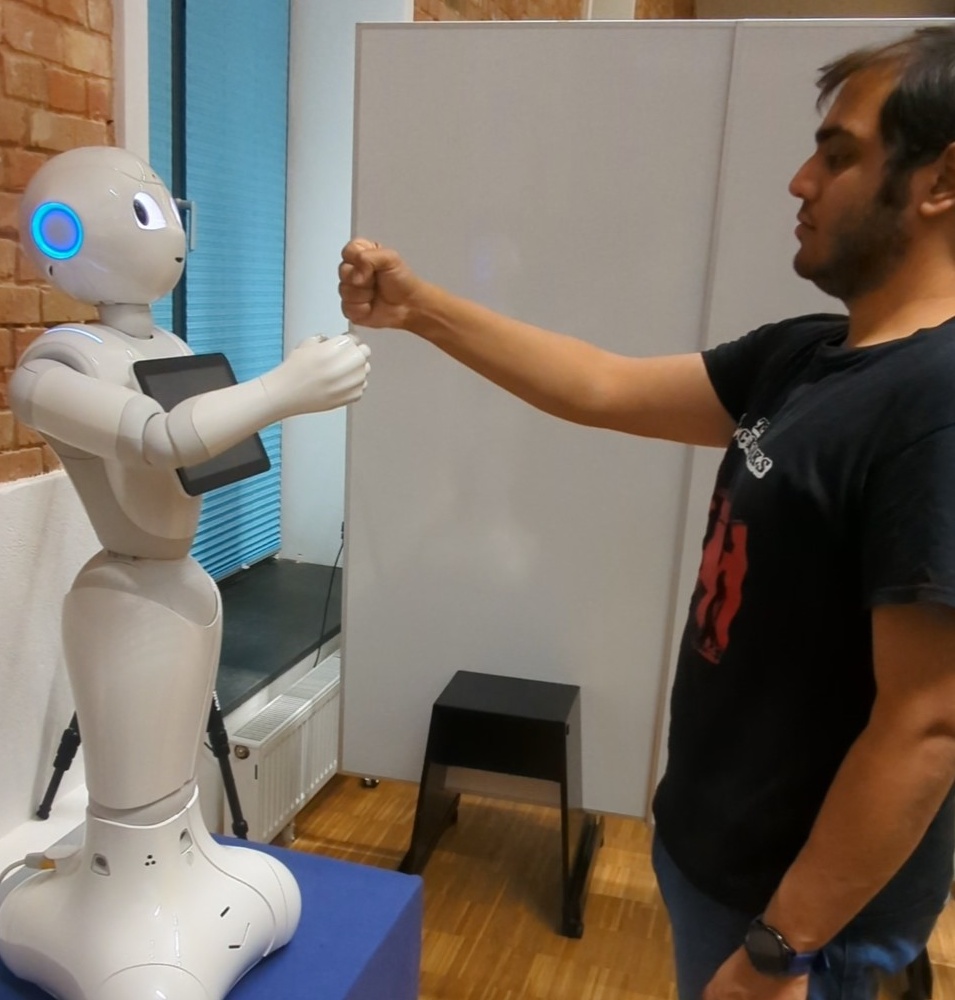} \hfill \includegraphics[width=0.19\textwidth]{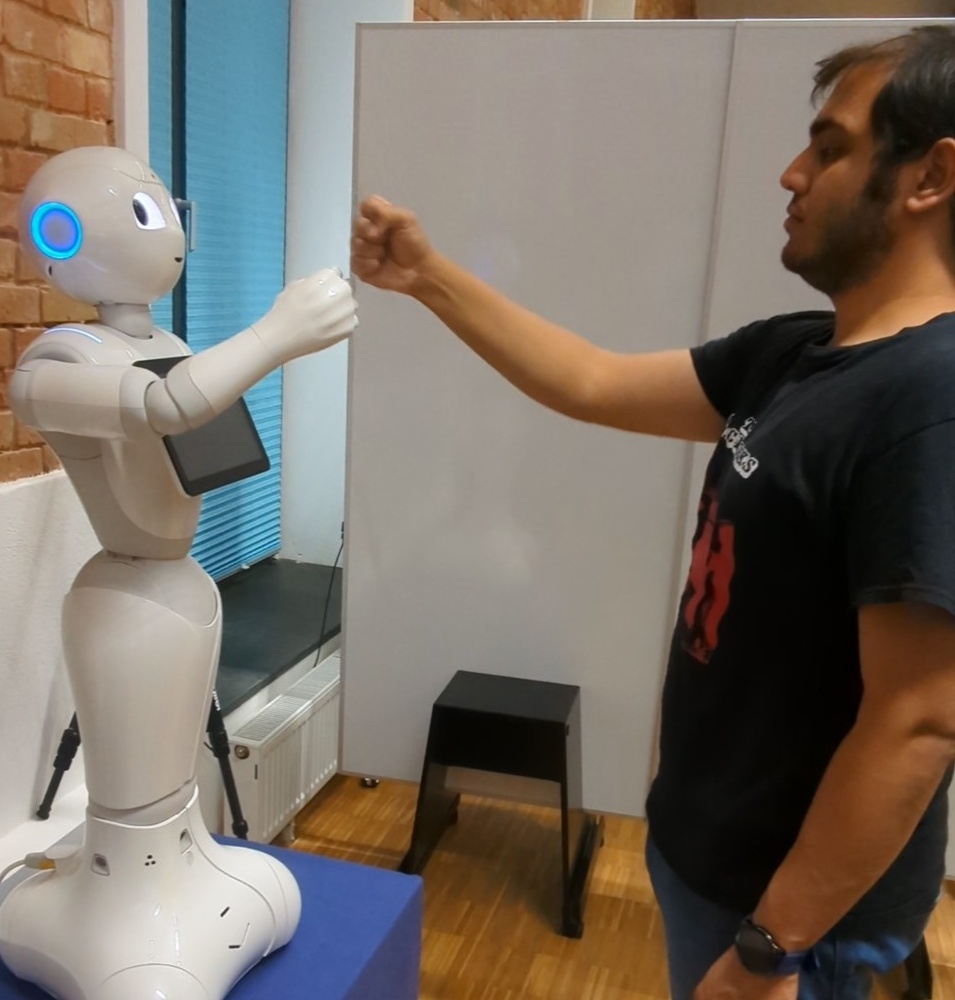} \hfill \includegraphics[width=0.19\textwidth]{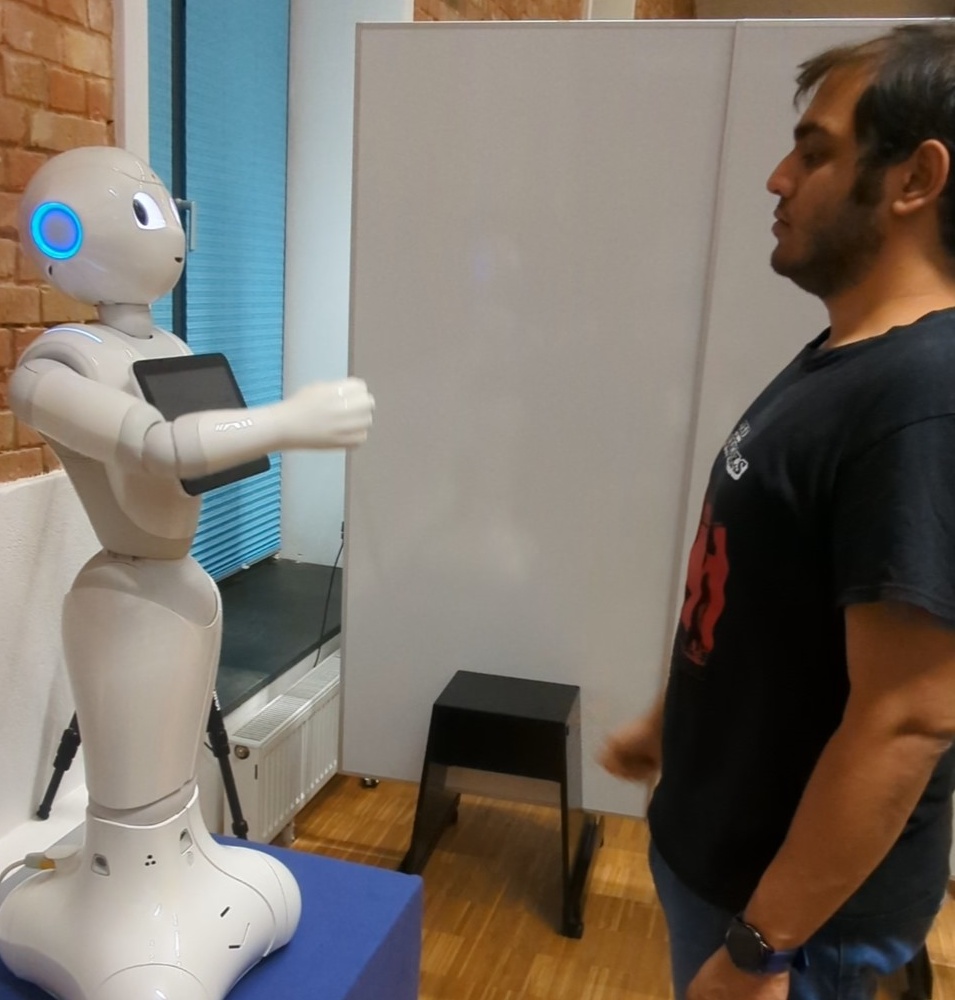} \\
         \includegraphics[width=0.19\textwidth]{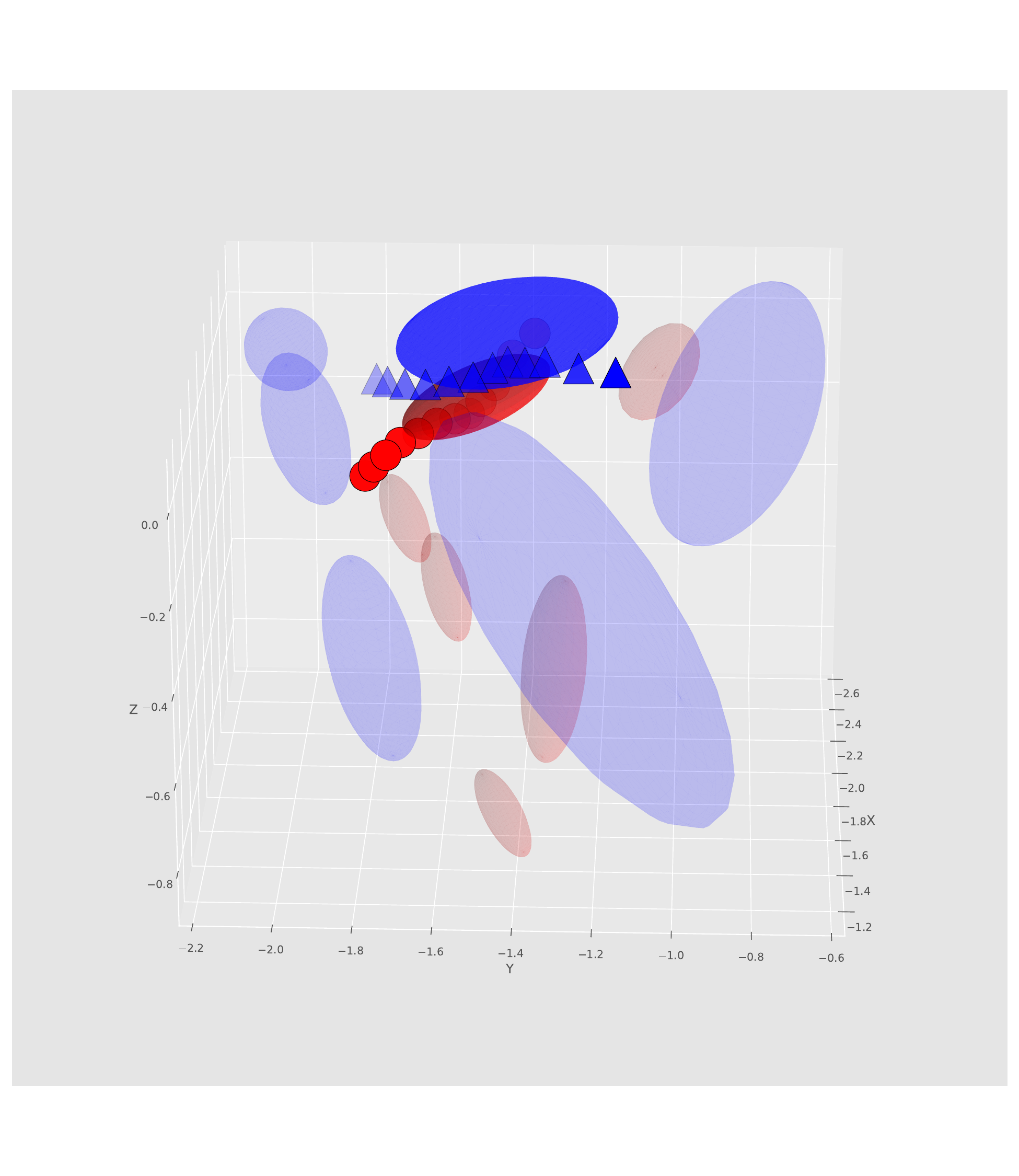} \hfill \includegraphics[width=0.19\textwidth]{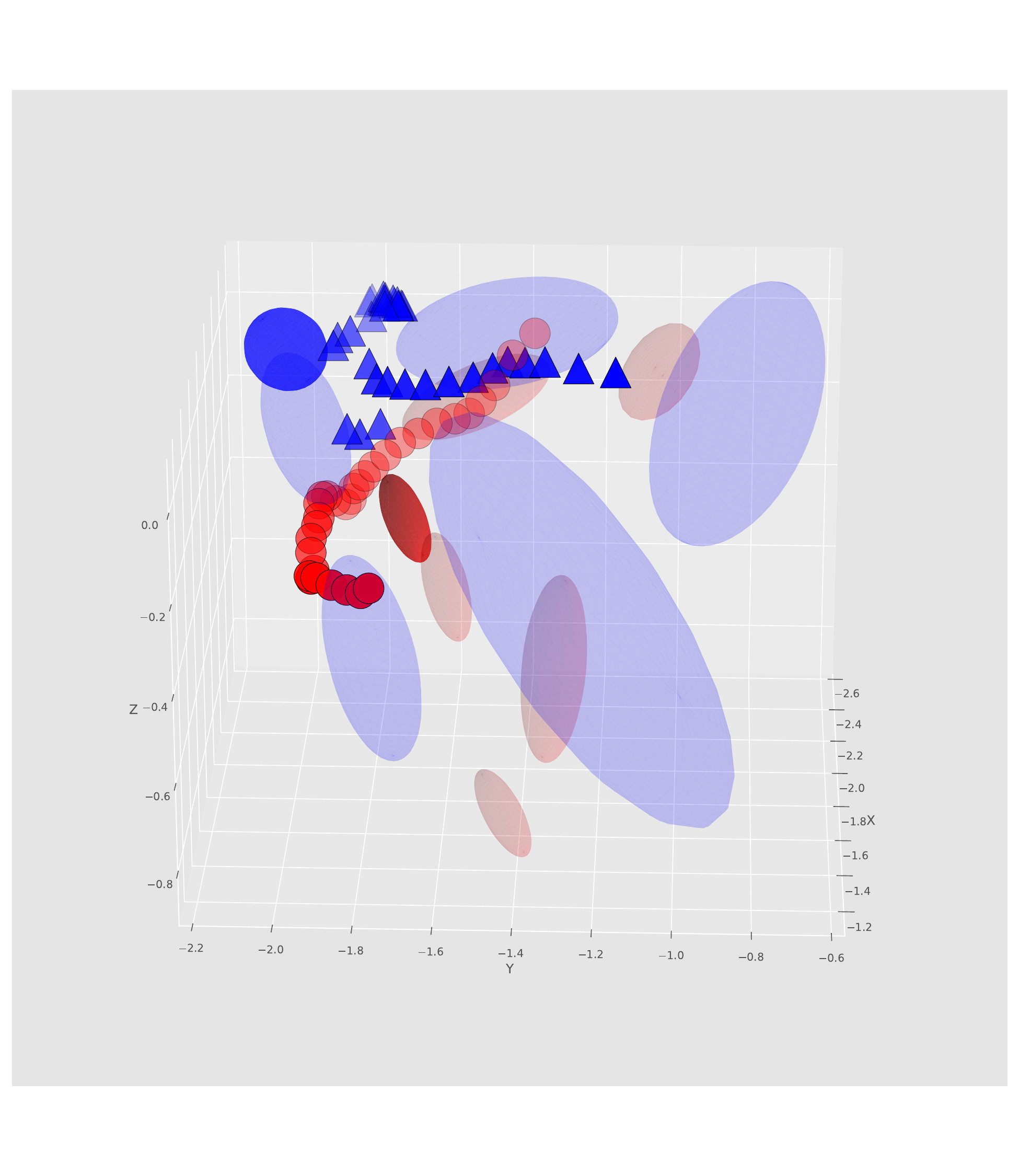} \hfill \includegraphics[width=0.19\textwidth]{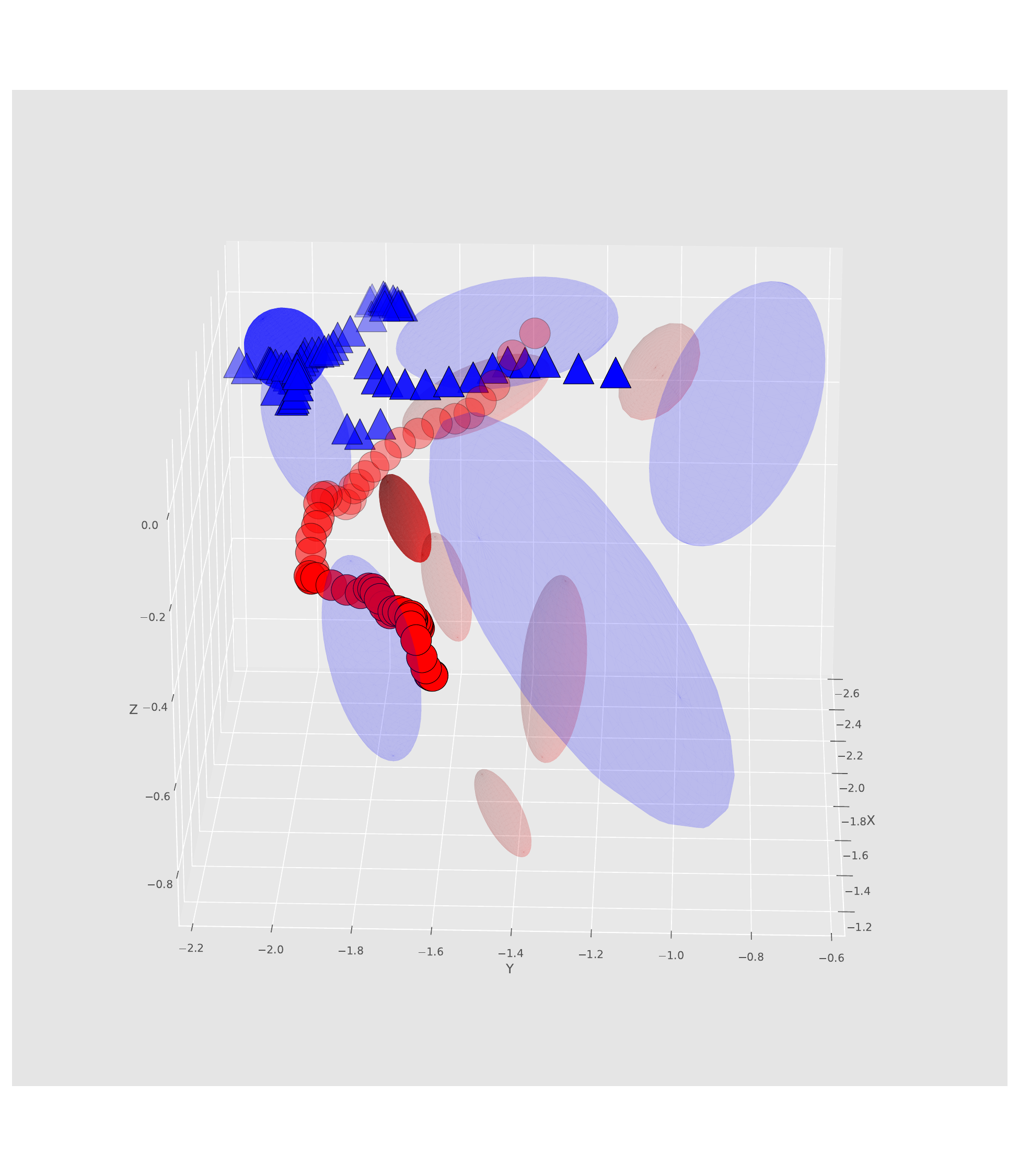} \hfill \includegraphics[width=0.19\textwidth]{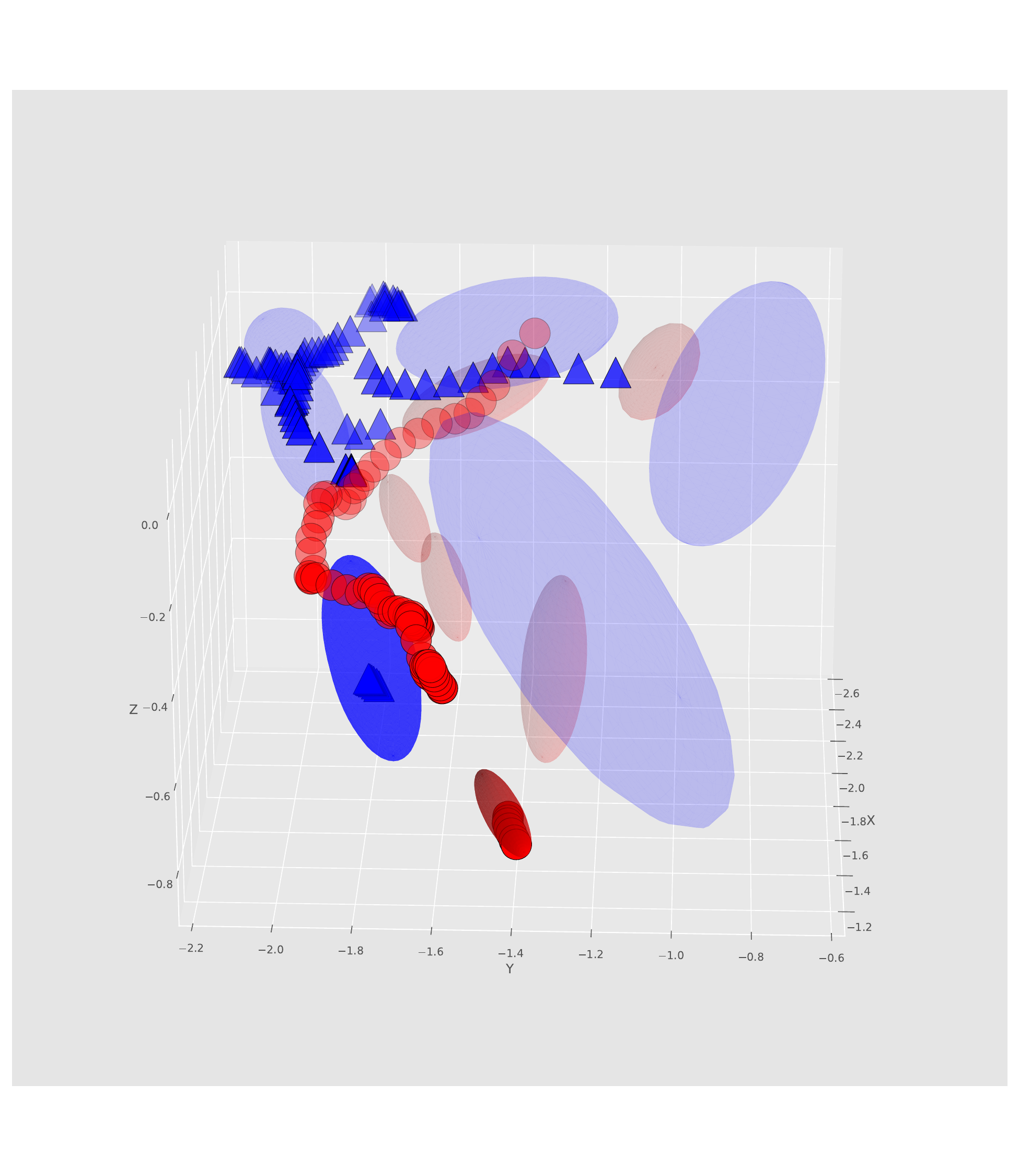} \hfill \includegraphics[width=0.19\textwidth]{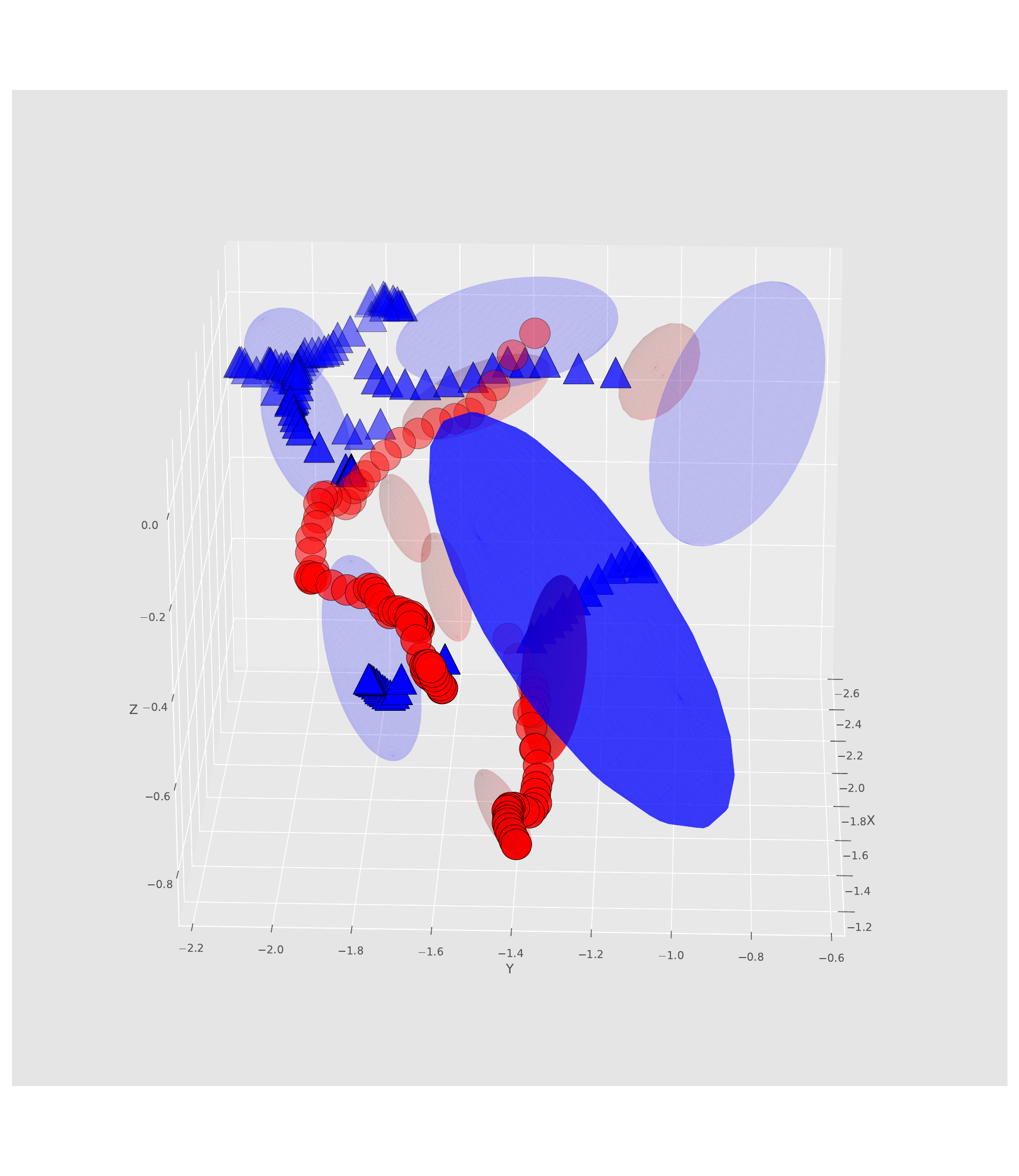} \\ 
         \includegraphics[width=\textwidth]{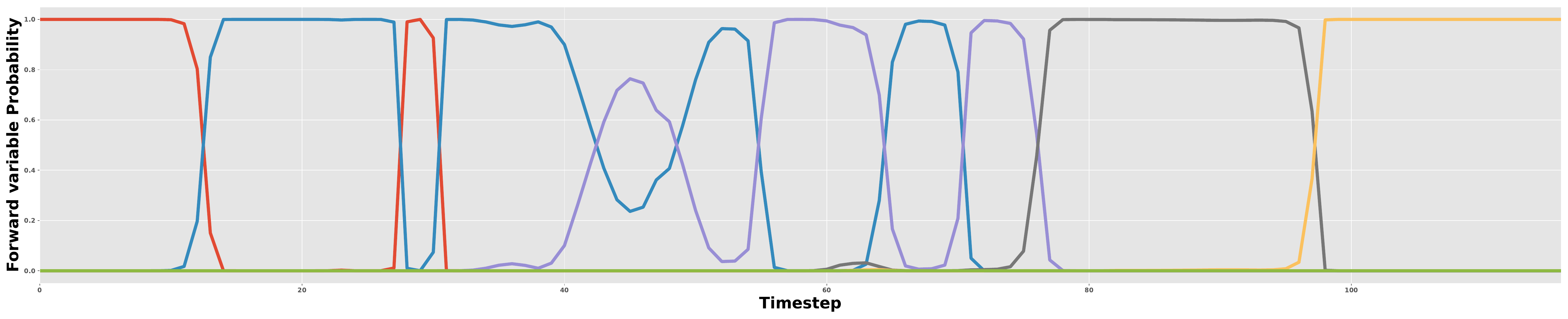}
         \caption{Sample Rocket HRI}
         \label{fig:hri-rocket}
     \end{subfigure}
    \vspace{-0.5em} 
    \caption{Sample sequences for the Handshake and Rocket fistbump interactions on the Pepper robot. The top row of each interaction shows the result of the generated reactive motion after observing the human partner's skeleton. The middle row shows the latent trajectories and the HMM segments over the first 3 dimensions of the latent space (Red - Human, Blue - Robot). The opacity of each cluster is the corresponding cluster probability, given by the HMM forward variable. The progression of the HMM forward variable is shown in the bottom row, with different colors denoting the different segments.}
    \label{fig:hri-samples}
\end{figure*}

%% file: tabs_n_figs/anova_table.tex

\begin{table*}[h!]
\caption{Results of a one-way Repeated Measures ANOVA for each of the survey items. Values less than 0.01 are reported as~0. The last three columns show the mean differences between the different algorithms (1 - Base-IK, 2 - MILD, 3 - MILD-IK) which were analyzed using paired sample t-tests (* - $p<0.05$, ** - $p<0.01$).}
\label{table:anova-results}
\centering
\begin{adjustbox}{max width=\textwidth}
\begin{tabular}{|c|cc|cc|cc|cc|l|l|l|}
\hline
\multicolumn{1}{|c|}{\multirow{2}{*}{Survey Item}} & \multicolumn{2}{c|}{Anova Results} & \multicolumn{2}{c|}{Base-IK} & \multicolumn{2}{c|}{MILD} & \multicolumn{2}{c|}{MILD-IK} & \multicolumn{1}{c|}{\multirow{2}{*}{$\Delta\mu_{1,2}$}} & \multicolumn{1}{c|}{\multirow{2}{*}{$\Delta\mu_{3,2}$}} & \multicolumn{1}{c|}{\multirow{2}{*}{$\Delta\mu_{3,1}$}} \\ \cline{2-9}
 & \multicolumn{1}{c|}{$F_{2,38}$} & $p$-value & \multicolumn{1}{c|}{Mean} & \multicolumn{1}{c|}{SD} & \multicolumn{1}{c|}{Mean} & \multicolumn{1}{c|}{SD} & \multicolumn{1}{c|}{Mean} & \multicolumn{1}{c|}{SD} & \multicolumn{1}{c|}{} & \multicolumn{1}{c|}{} & \multicolumn{1}{c|}{} \\ \hline
Pleasantness & \multicolumn{1}{c|}{0.86} & 0.43 & \multicolumn{1}{c|}{3.50} & \multicolumn{1}{c|}{1.07} & \multicolumn{1}{c|}{3.40} & \multicolumn{1}{c|}{1.07} & \multicolumn{1}{c|}{3.75} & \multicolumn{1}{c|}{0.89} & 0.10 & 0.35 & 0.25 \\ \hline
Excitement & \multicolumn{1}{c|}{1.32} & 0.28 & \multicolumn{1}{c|}{3.85} & \multicolumn{1}{c|}{0.85} & \multicolumn{1}{c|}{3.45} & \multicolumn{1}{c|}{1.02} & \multicolumn{1}{c|}{3.75} & \multicolumn{1}{c|}{0.89} & 0.40 & 0.30 & -0.10 \\ \hline
Human-likeness & \multicolumn{1}{c|}{4.89} & \textbf{0.01} & \multicolumn{1}{c|}{2.60} & \multicolumn{1}{c|}{1.20} & \multicolumn{1}{c|}{2.45} & \multicolumn{1}{c|}{1.16} & \multicolumn{1}{c|}{3.25} & \multicolumn{1}{c|}{1.22} & 0.15 & \textbf{0.80*} & \textbf{0.65*} \\ \hline
Naturalness & \multicolumn{1}{c|}{3.46} & \textbf{0.04} & \multicolumn{1}{c|}{2.95} & \multicolumn{1}{c|}{1.24} & \multicolumn{1}{c|}{2.55} & \multicolumn{1}{c|}{1.20} & \multicolumn{1}{c|}{3.25} & \multicolumn{1}{c|}{1.18} & 0.40 & \textbf{0.70*} & 0.30 \\ \hline
Friendliness & \multicolumn{1}{c|}{0.56} & 0.58 & \multicolumn{1}{c|}{3.75} & \multicolumn{1}{c|}{1.04} & \multicolumn{1}{c|}{3.75} & \multicolumn{1}{c|}{0.83} & \multicolumn{1}{c|}{3.95} & \multicolumn{1}{c|}{0.86} & 0.00 & 0.20 & 0.20 \\ \hline
Comfort & \multicolumn{1}{c|}{2.79} & 0.07 & \multicolumn{1}{c|}{3.35} & \multicolumn{1}{c|}{1.06} & \multicolumn{1}{c|}{3.35} & \multicolumn{1}{c|}{0.96} & \multicolumn{1}{c|}{3.85} & \multicolumn{1}{c|}{0.96} & 0.00 & \textbf{0.50*} & 0.50 \\ \hline
Timing & \multicolumn{1}{c|}{5.77} & \textbf{0.01} & \multicolumn{1}{c|}{3.60} & \multicolumn{1}{c|}{1.07} & \multicolumn{1}{c|}{3.45} & \multicolumn{1}{c|}{1.16} & \multicolumn{1}{c|}{4.15} & \multicolumn{1}{c|}{0.96} & 0.15 & \textbf{0.70**} & \textbf{0.55*} \\ \hline
Accuracy & \multicolumn{1}{c|}{10.15} & \textbf{0.00} & \multicolumn{1}{c|}{3.15} & \multicolumn{1}{c|}{1.28} & \multicolumn{1}{c|}{2.55} & \multicolumn{1}{c|}{1.24} & \multicolumn{1}{c|}{3.95} & \multicolumn{1}{c|}{0.86} & 0.60 & 1.40 & 0.80 \\ \hline
Annoyance & \multicolumn{1}{c|}{2.08} & 0.14 & \multicolumn{1}{c|}{1.60} & \multicolumn{1}{c|}{0.80} &  \multicolumn{1}{c|}{1.95} & \multicolumn{1}{c|}{1.20} & \multicolumn{1}{c|}{1.50} & \multicolumn{1}{c|}{0.81} & -0.35 & \textbf{-0.45*} & -0.10 \\ \hline
Aggression & \multicolumn{1}{c|}{0.77} & 0.47 & \multicolumn{1}{c|}{1.15} & \multicolumn{1}{c|}{0.48} & \multicolumn{1}{c|}{1.30} & \multicolumn{1}{c|}{0.64} & \multicolumn{1}{c|}{1.25} & \multicolumn{1}{c|}{0.62} & -0.15 & -0.05 & 0.10 \\ \hline
Awkwardness & \multicolumn{1}{c|}{2.29} & 0.12 & \multicolumn{1}{c|}{2.30} & \multicolumn{1}{c|}{1.05} & \multicolumn{1}{c|}{2.75} & \multicolumn{1}{c|}{1.34} & \multicolumn{1}{c|}{2.35} & \multicolumn{1}{c|}{1.24} & -0.45 & -0.40 & 0.05 \\ \hline
Scariness & \multicolumn{1}{c|}{0.59} & 0.56 & \multicolumn{1}{c|}{1.15} & \multicolumn{1}{c|}{0.48} & \multicolumn{1}{c|}{1.25} & \multicolumn{1}{c|}{0.54} & \multicolumn{1}{c|}{1.20} & \multicolumn{1}{c|}{0.51} & -0.10 & -0.05 & 0.05 \\ \hline
Satisfaction & \multicolumn{1}{c|}{2.54} & 0.09 & \multicolumn{1}{c|}{3.30} & \multicolumn{1}{c|}{1.14} & \multicolumn{1}{c|}{3.15} & \multicolumn{1}{c|}{0.96} & \multicolumn{1}{c|}{3.80} & \multicolumn{1}{c|}{0.87} & 0.15 & \textbf{0.65*} & 0.50 \\ \hline
\begin{tabular}[c]{@{}c@{}}Effortlessness\\ between\\ the two trials\end{tabular} & \multicolumn{1}{c|}{4.15} & \textbf{0.02} & \multicolumn{1}{c|}{3.35} & \multicolumn{1}{c|}{1.19} & \multicolumn{1}{c|}{3.40} & \multicolumn{1}{c|}{1.16} & \multicolumn{1}{c|}{4.05} & \multicolumn{1}{c|}{1.24} & -0.05 & \textbf{0.65*} & \textbf{0.70*} \\ \hline
\end{tabular}
\end{adjustbox}

\vspace{-1em}
\end{table*}

%% file: tabs_n_figs/robot_mse_pvalues.tex
\begin{table}[h!]
\centering
\caption{$p$-values for pairwise MSE comparisons of the results reported in Table~\ref{tab:pred-hri}. For $p<0.001$, we report zeros.}
\label{tab:hri-significance}
\resizebox{\linewidth}{!}{%
\begin{tabular}{ccc|c|c|c|c|c|}
\cline{4-8}
&&& \multicolumn{5}{|c|}{Version of MILD} \\
\cline{4-8}
\multicolumn{3}{c|}{} & v1 & v2.1 & v2.2 & v3.1 & v3.2 \\ \hline
\multicolumn{1}{|c|}{\parbox[t]{1mm}{\multirow{20}{*}{\rotatebox[origin=c]{90}{HRI-Yumi~\cite{butepage2020imitating}}}}} & \multicolumn{1}{c|}{\parbox[t]{3mm}{\multirow{5}{*}{\rotatebox[origin=c]{90}{Waving}}}} & \cite{butepage2020imitating} & \textbf{0.} & \textbf{0.} & \textbf{0.} & \textbf{0.} & \textbf{0.} \\ \cline{3-8} 
\multicolumn{1}{|c|}{} & \multicolumn{1}{c|}{} & MILD v1 & -- & \textbf{0.} & \textbf{0.} & \textbf{0.} & \textbf{0.} \\ \cline{3-8} 
\multicolumn{1}{|c|}{} & \multicolumn{1}{c|}{} & MILD v2.1 & -- & -- & \textbf{0.} & 0.284 & 0.337 \\ \cline{3-8} 
\multicolumn{1}{|c|}{} & \multicolumn{1}{c|}{} & MILD v2.2 & -- & -- & -- & \textbf{0.} & \textbf{0.} \\ \cline{3-8} 
\multicolumn{1}{|c|}{} & \multicolumn{1}{c|}{} & MILD v3.1 & -- & -- & -- & -- & 0.675 \\ \cline{2-8} 
\multicolumn{1}{|c|}{} & \multicolumn{1}{c|}{\parbox[t]{3mm}{\multirow{5}{*}{\rotatebox[origin=c]{90}{Handshake}}}} & \cite{butepage2020imitating} & \textbf{0.} & \textbf{0.} & 0.397 & \textbf{0.} & \textbf{0.} \\ \cline{3-8} 
\multicolumn{1}{|c|}{} & \multicolumn{1}{c|}{} & MILD v1 & -- & \textbf{0.} & \textbf{0.} & 0. & 0. \\ \cline{3-8} 
\multicolumn{1}{|c|}{} & \multicolumn{1}{c|}{} & MILD v2.1 & -- & -- & \textbf{0.018} & \textbf{0.} & \textbf{0.} \\ \cline{3-8} 
\multicolumn{1}{|c|}{} & \multicolumn{1}{c|}{} & MILD v2.2 & -- & -- & -- & \textbf{0.} & \textbf{0.} \\ \cline{3-8} 
\multicolumn{1}{|c|}{} & \multicolumn{1}{c|}{} & MILD v3.1 & -- & -- & -- & -- & \textbf{0.008} \\ \cline{2-8} 
\multicolumn{1}{|c|}{} & \multicolumn{1}{c|}{\parbox[t]{3mm}{\multirow{5}{*}{\hspace{-0.5em}\rotatebox[origin=c]{90}{\begin{tabular}[c]{@{}c@{}}Rocket\\Fistbump\end{tabular}}}}} & \cite{butepage2020imitating} & \textbf{0.} & \textbf{0.} & \textbf{0.} & 0.976 & 0.999 \\ \cline{3-8} 
\multicolumn{1}{|c|}{} & \multicolumn{1}{c|}{} & MILD v1 & -- & \textbf{0.} & \textbf{0.} & \textbf{0.} & \textbf{0.} \\ \cline{3-8} 
\multicolumn{1}{|c|}{} & \multicolumn{1}{c|}{} & MILD v2.1 & -- & -- & \textbf{0.} & \textbf{0.} & \textbf{0.} \\ \cline{3-8} 
\multicolumn{1}{|c|}{} & \multicolumn{1}{c|}{} & MILD v2.2 & -- & -- & -- & \textbf{0.} & \textbf{0.} \\ \cline{3-8} 
\multicolumn{1}{|c|}{} & \multicolumn{1}{c|}{} & MILD v3.1 & -- & -- & -- & -- & 0.909 \\ \cline{2-8} 
\multicolumn{1}{|c|}{} & \multicolumn{1}{c|}{\parbox[t]{3mm}{\multirow{5}{*}{\hspace{-0.5em}\rotatebox[origin=c]{90}{\begin{tabular}[c]{@{}c@{}}Parachute\\Fistbump\end{tabular}}}}} & \cite{butepage2020imitating} & \textbf{0.} & 0.136 & 0.098 & \textbf{0.} & \textbf{0.} \\ \cline{3-8} 
\multicolumn{1}{|c|}{} & \multicolumn{1}{c|}{} & MILD v1 & -- & \textbf{0.} & \textbf{0.} & 0.055 & \textbf{0.} \\ \cline{3-8} 
\multicolumn{1}{|c|}{} & \multicolumn{1}{c|}{} & MILD v2.1 & -- & -- & 0.785 & \textbf{0.} & \textbf{0.} \\ \cline{3-8} 
\multicolumn{1}{|c|}{} & \multicolumn{1}{c|}{} & MILD v2.2 & -- & -- & -- & \textbf{0.} & \textbf{0.} \\ \cline{3-8} 
\multicolumn{1}{|c|}{} & \multicolumn{1}{c|}{} & MILD v3.1 & -- & -- & -- & -- & \textbf{0.003} \\ \hline
\multicolumn{1}{|c|}{\parbox[t]{1mm}{\multirow{20}{*}{\rotatebox[origin=c]{90}{HRI-Pepper~\cite{butepage2020imitating}}}}} & \multicolumn{1}{c|}{\parbox[t]{3mm}{\multirow{5}{*}{\rotatebox[origin=c]{90}{Waving}}}} & \cite{butepage2020imitating} & \textbf{0.} & \textbf{0.} & \textbf{0.} & \textbf{0.} & \textbf{0.} \\ \cline{3-8} 
\multicolumn{1}{|c|}{} & \multicolumn{1}{c|}{} & MILD v1 & -- & \textbf{0.} & \textbf{0.} & \textbf{0.} & \textbf{0.} \\ \cline{3-8} 
\multicolumn{1}{|c|}{} & \multicolumn{1}{c|}{} & MILD v2.1 & -- & -- & \textbf{0.} & \textbf{0.} & \textbf{0.} \\ \cline{3-8} 
\multicolumn{1}{|c|}{} & \multicolumn{1}{c|}{} & MILD v2.2 & -- & -- & -- & \textbf{0.} & \textbf{0.} \\ \cline{3-8} 
\multicolumn{1}{|c|}{} & \multicolumn{1}{c|}{} & MILD v3.1 & -- & -- & -- & -- & 0.119 \\ \cline{2-8} 
\multicolumn{1}{|c|}{} & \multicolumn{1}{c|}{\parbox[t]{3mm}{\multirow{5}{*}{\rotatebox[origin=c]{90}{Handshake}}}} & \cite{butepage2020imitating} & \textbf{0.} & \textbf{0.} & \textbf{0.} & \textbf{0.} & \textbf{0.} \\ \cline{3-8} 
\multicolumn{1}{|c|}{} & \multicolumn{1}{c|}{} & MILD v1 & -- & \textbf{0.} & \textbf{0.} & \textbf{0.} & \textbf{0.} \\ \cline{3-8} 
\multicolumn{1}{|c|}{} & \multicolumn{1}{c|}{} & MILD v2.1 & -- & -- & \textbf{0.} & \textbf{0.} & \textbf{0.} \\ \cline{3-8} 
\multicolumn{1}{|c|}{} & \multicolumn{1}{c|}{} & MILD v2.2 & -- & -- & -- & \textbf{0.} & \textbf{0.} \\ \cline{3-8} 
\multicolumn{1}{|c|}{} & \multicolumn{1}{c|}{} & MILD v3.1 & -- & -- & -- & -- & \textbf{0.041} \\ \cline{2-8} 
\multicolumn{1}{|c|}{} & \multicolumn{1}{c|}{\parbox[t]{3mm}{\multirow{5}{*}{\hspace{-0.5em}\rotatebox[origin=c]{90}{\begin{tabular}[c]{@{}c@{}}Rocket\\Fistbump\end{tabular}}}}} & \cite{butepage2020imitating} & \textbf{0.} & 0.291 & 0.838 & \textbf{0.} & \textbf{0.} \\ \cline{3-8} 
\multicolumn{1}{|c|}{} & \multicolumn{1}{c|}{} & MILD v1 & -- & \textbf{0.} & \textbf{0.} & \textbf{0.} & \textbf{0.} \\ \cline{3-8} 
\multicolumn{1}{|c|}{} & \multicolumn{1}{c|}{} & MILD v2.1 & -- & -- & \textbf{0.} & \textbf{0.} & \textbf{0.} \\ \cline{3-8} 
\multicolumn{1}{|c|}{} & \multicolumn{1}{c|}{} & MILD v2.2 & -- & -- & -- & \textbf{0.} & \textbf{0.} \\ \cline{3-8} 
\multicolumn{1}{|c|}{} & \multicolumn{1}{c|}{} & MILD v3.1 & -- & -- & -- & -- & \textbf{0.} \\ \cline{2-8} 
\multicolumn{1}{|c|}{} & \multicolumn{1}{c|}{\parbox[t]{3mm}{\multirow{5}{*}{\hspace{-0.5em}\rotatebox[origin=c]{90}{\begin{tabular}[c]{@{}c@{}}Parachute\\Fistbump\end{tabular}}}}} & \cite{butepage2020imitating} & \textbf{0.} & \textbf{0.} & \textbf{0.} & \textbf{0.} & \textbf{0.} \\ \cline{3-8} 
\multicolumn{1}{|c|}{} & \multicolumn{1}{c|}{} & MILD v1 & -- & \textbf{0.} & \textbf{0.} & \textbf{0.} & \textbf{0.} \\ \cline{3-8} 
\multicolumn{1}{|c|}{} & \multicolumn{1}{c|}{} & MILD v2.1 & -- & -- & \textbf{0.016} & \textbf{0.} & \textbf{0.} \\ \cline{3-8} 
\multicolumn{1}{|c|}{} & \multicolumn{1}{c|}{} & MILD v2.2 & -- & -- & -- & \textbf{0.} & \textbf{0.} \\ \cline{3-8} 
\multicolumn{1}{|c|}{} & \multicolumn{1}{c|}{} & MILD v3.1 & -- & -- & -- & -- & \textbf{0.018} \\ \hline
\multicolumn{1}{|c|}{\parbox[t]{1mm}{\multirow{20}{*}{\rotatebox[origin=c]{90}{HRI-Pepper (NuiSI)}}}} & \multicolumn{1}{c|}{\parbox[t]{3mm}{\multirow{5}{*}{\rotatebox[origin=c]{90}{Waving}}}} & \cite{butepage2020imitating} & \textbf{0.} & \textbf{0.} & 0.139 & 0.158 & 0.309 \\ \cline{3-8} 
\multicolumn{1}{|c|}{} & \multicolumn{1}{c|}{} & MILD v1 & -- & \textbf{0.} & \textbf{0.} & \textbf{0.} & \textbf{0.} \\ \cline{3-8} 
\multicolumn{1}{|c|}{} & \multicolumn{1}{c|}{} & MILD v2.1 & -- & -- & \textbf{0.} & \textbf{0.} & \textbf{0.} \\ \cline{3-8} 
\multicolumn{1}{|c|}{} & \multicolumn{1}{c|}{} & MILD v2.2 & -- & -- & -- & \textbf{0.004} & 0.530 \\ \cline{3-8} 
\multicolumn{1}{|c|}{} & \multicolumn{1}{c|}{} & MILD v3.1 & -- & -- & -- & -- & \textbf{0.014} \\ \cline{2-8} 
\multicolumn{1}{|c|}{} & \multicolumn{1}{c|}{\parbox[t]{3mm}{\multirow{5}{*}{\rotatebox[origin=c]{90}{Handshake}}}} & \cite{butepage2020imitating} & \textbf{0.} & \textbf{0.} & \textbf{0.} & \textbf{0.} & \textbf{0.} \\ \cline{3-8} 
\multicolumn{1}{|c|}{} & \multicolumn{1}{c|}{} & MILD v1 & -- & \textbf{0.} & \textbf{0.} & \textbf{0.} & \textbf{0.} \\ \cline{3-8} 
\multicolumn{1}{|c|}{} & \multicolumn{1}{c|}{} & MILD v2.1 & -- & -- & 0.862 & 0.095 & \textbf{0.003} \\ \cline{3-8} 
\multicolumn{1}{|c|}{} & \multicolumn{1}{c|}{} & MILD v2.2 & -- & -- & -- & 0.104 & \textbf{0.007} \\ \cline{3-8} 
\multicolumn{1}{|c|}{} & \multicolumn{1}{c|}{} & MILD v3.1 & -- & -- & -- & -- & 0.065 \\ \cline{2-8} 
\multicolumn{1}{|c|}{} & \multicolumn{1}{c|}{\parbox[t]{3mm}{\multirow{5}{*}{\hspace{-0.5em}\rotatebox[origin=c]{90}{\begin{tabular}[c]{@{}c@{}}Rocket\\Fistbump\end{tabular}}}}} & \cite{butepage2020imitating} & \textbf{0.} & \textbf{0.} & \textbf{0.} & \textbf{0.} & \textbf{0.} \\ \cline{3-8} 
\multicolumn{1}{|c|}{} & \multicolumn{1}{c|}{} & MILD v1 & -- & \textbf{0.} & \textbf{0.} & \textbf{0.} & \textbf{0.} \\ \cline{3-8} 
\multicolumn{1}{|c|}{} & \multicolumn{1}{c|}{} & MILD v2.1 & -- & -- & \textbf{0.036} & \textbf{0.} & \textbf{0.} \\ \cline{3-8} 
\multicolumn{1}{|c|}{} & \multicolumn{1}{c|}{} & MILD v2.2 & -- & -- & -- & 0.050 & \textbf{0.023} \\ \cline{3-8} 
\multicolumn{1}{|c|}{} & \multicolumn{1}{c|}{} & MILD v3.1 & -- & -- & -- & -- & 0.685 \\ \cline{2-8} 
\multicolumn{1}{|c|}{} & \multicolumn{1}{c|}{\parbox[t]{3mm}{\multirow{5}{*}{\hspace{-0.5em}\rotatebox[origin=c]{90}{\begin{tabular}[c]{@{}c@{}}Parachute\\Fistbump\end{tabular}}}}} & \cite{butepage2020imitating} & \textbf{0.} & \textbf{0.} & \textbf{0.} & \textbf{0.} & \textbf{0.} \\ \cline{3-8} 
\multicolumn{1}{|c|}{} & \multicolumn{1}{c|}{} & MILD v1 & -- & \textbf{0.} & \textbf{0.} & \textbf{0.} & \textbf{0.} \\ \cline{3-8} 
\multicolumn{1}{|c|}{} & \multicolumn{1}{c|}{} & MILD v2.1 & -- & -- & \textbf{0.026} & \textbf{0.017} & \textbf{0.} \\ \cline{3-8} 
\multicolumn{1}{|c|}{} & \multicolumn{1}{c|}{} & MILD v2.2 & -- & -- & -- & 0.808 & \textbf{0.018} \\ \cline{3-8} 
\multicolumn{1}{|c|}{} & \multicolumn{1}{c|}{} & MILD v3.1 & -- & -- & -- & -- & \textbf{0.023} \\ \hline
\end{tabular}
}
\end{table}